\documentclass[11pt]{article}

\usepackage{PRIMEarxiv}
\usepackage{longtable}
\usepackage[utf8]{inputenc} 
\usepackage[T1]{fontenc}    
\usepackage{url}            
\usepackage{booktabs}       
\usepackage{amsfonts}       
\usepackage{nicefrac}       
\usepackage{microtype}      
\usepackage{lipsum}
\usepackage{fancyhdr}       
\usepackage{graphicx}       
\graphicspath{{media/}}     
\usepackage{tabularx}
\usepackage{listings} 
\usepackage{fancyvrb} 
\usepackage{verbatimbox}
\usepackage[utf8]{inputenc}
\usepackage[T1]{fontenc}
\usepackage[numbers]{natbib}
\usepackage{amsmath,amssymb}
\usepackage{amsthm}
\usepackage{mathtools}
\usepackage{xspace}
\usepackage[noend]{algorithmic}
\usepackage[ruled,vlined]{algorithm2e}
\usepackage{enumitem}
\usepackage{pdfpages}
\usepackage{tikz}
\usetikzlibrary{shapes.geometric, arrows}

\tikzstyle{process} = [rectangle, rounded corners, minimum width=3cm, minimum height=1cm, text centered, draw=black, fill=orange!30]
\tikzstyle{connector} = [rectangle, rounded corners, minimum width=3cm, minimum height=1cm, text centered, draw=black, fill=violet!30]
\tikzstyle{decoder} = [ellipse, minimum width=3cm, minimum height=1cm, text centered, draw=black, fill=blue!20]
\tikzstyle{arrow} = [thick,->,>=stealth]

\usepackage{seqsplit}

\usepackage{chngcntr}
\usepackage{mdframed} 
\usepackage{caption}

\usepackage{color}
\usepackage{xcolor}

\usepackage{subfigure}
\usepackage[most]{tcolorbox}
\usepackage[frozencache=true, finalizecache=false, cachedir=.]{minted}

\usepackage{float}
\usepackage{alltt}
\usepackage{soul}
\usepackage{fancyvrb}
\usepackage{multirow}
\usepackage[final]{hyperref}

\usepackage{csquotes}

\newcounter{textbox}
\renewcommand{\thetextbox}{\arabic{textbox}} 

\newcommand{\textboxcaption}[1]{
  \refstepcounter{textbox}
  \noindent\small\textbf{Text Box \thetextbox:} #1\par\medskip
}
\usepackage{listings} 

\tcbset{
  custombox/.style={
    width=474.18663pt,
    top=3pt,
    fonttitle=\bfseries\small\sffamily, 
    colback=white,
    colframe=black,
    colbacktitle=black,
    boxrule=0.5pt, 
    colback=white, 
    enhanced,
    rounded corners, 
    center,
    boxed title style={
            sharp corners,
            size=small,
            colframe=blue!75!black, 
        },
    attach boxed title to top left={yshift=-0.1in,xshift=0.15in},
    boxed title style={boxrule=0pt,colframe=white,},
  }
}
\newtcolorbox{LLMbox}[2][]{custombox,title=#2,#1}

\tcbset{
  customboxmultipage/.style={
    width=474.18663pt,
    top=1pt,
    bottom=1pt, 
    left=1pt, 
    right=1pt, 
    fonttitle=\bfseries\scriptsize\sffamily, 
    colback=white,
    breakable,
    colframe=black,
    colbacktitle=black,
    boxrule=0.5pt, 
    enhanced,
    rounded corners,
    center,
    boxed title style={
      sharp corners,
      size=tiny, 
      colframe=blue!75!black,
    },
    attach boxed title to top left={yshift=-0.1in,xshift=0.15in},
    boxed title style={boxrule=0pt,colframe=white,},
    before={\vspace{-5pt}},
    after={\vspace{-5pt}},
    before upper={\vspace{-5pt}},
    after upper={\vspace{-5pt}},
    before lower={\vspace{-5pt}},
    after lower={\vspace{-5pt}},
    coltext=black,
    fontupper=\tiny,
  }
}
\newtcolorbox{LLMboxmultipage}[2][]{customboxmultipage,title=#2,#1}

\newtcbox{\mybox}[1][green]{on line,
arc=0pt,outer arc=0pt,colback=#1!10!white,colframe=#1!50!black,
boxsep=0pt,left=0pt,right=0pt,top=0pt,bottom=0pt,
boxrule=0pt,bottomrule=0pt,toprule=0pt}

\definecolor{aigold}{RGB}{244,210, 1} 
\definecolor{aigreen}{RGB}{210,244,211} 

\sethlcolor{aigreen}

\definecolor{aired}{RGB}{255,180,181}

\definecolor{lightred}{rgb}{1,0.9,0.9} 

\newcommand{\hlred}[1]{{%
  \sethlcolor{lightred}
  \hl{#1}
}}

\title{Cephalo: Multi-Modal Vision-Language Models for Bio-Inspired Materials Analysis and Design

}

\author{
  Markus J. Buehler\thanks{Massachusetts Institute of Technology (MIT), 77 Mass. Ave 1-165, Cambridge, MA 02139, USA} \\
  \\
  \texttt{mbuehler@MIT.EDU} \\
}

\begin{document}
\maketitle

\begin{abstract}
We present Cephalo, a series of multimodal vision large language models (V-LLMs) designed for materials science applications, integrating visual and linguistic data for enhanced understanding and interaction within human-AI and multi-agent AI frameworks. A key innovation of Cephalo is its advanced dataset generation method, which employs a sophisticated algorithm to accurately detect and separate images and their corresponding textual descriptions from Portable Document Format (PDF) documents, such as scientific papers. The method conducts a careful refinement of image-text pairs through integrated vision and language processing, ensuring high-quality, contextually relevant, and well reasoned training data. Cephalo is trained on integrated image and text data extracted from thousands of scientific papers and science-focused Wikipedia pages demonstrates can interpret complex visual scenes, generate precise language descriptions, and answer queries about images effectively. The combination of a vision encoder with an autoregressive transformer supports complex natural language understanding in an integrated model, which can be coupled with other generative methods to create an image-to-text-to-image or image-to-text-to-3D pipeline. To explore the development of larger models from smaller ones, we explore both the development of mixture-of-expert models and model merging. In model merging, we combine sets of layers that originate from different pre-trained source models. This hybrid approach allows us to leverage the domain-specific expertise and general conversational capabilities to harness the strengths of multiple models. We provide model weights for various sizes, ranging from 4 billion to 12 billion parameters, to accommodate different computational needs and applications. We examine the models in diverse use cases that incorporate biological materials, fracture and engineering analysis, protein biophysics, and bio-inspired design based on insect behavior. Other applications include the development of bio-inspired material microstructures, including pollen-based architectured materials, as well as the synthesis of bio-inspired material microstructures from a photograph of a solar eclipse. Additional model fine-tuning with a series of molecular dynamics results demonstrate Cephalo's enhanced capabilities to accurately predict statistical features of stress and atomic energy distributions, as well as crack dynamics and damage in materials. These features show that the model is capable of understanding complex physical and mechanical behaviors, enabling the design of more resilient and high-performance materials. We provide a detailed discussion of challenges and opportunities.
\end{abstract}


\section{Introduction}
In the rapidly evolving field of scientific artificial intelligence, and specifically as applied to materials science, approaches have explored a variety of scales, material types, and use cases~\cite{Guo2021ArtificialMaterials,Bock2019AMechanics,Elton2019DeepArt,Popova2018DeepDesign,buehler2024accelerating,Lu2023GenerativePropertiesb,Gu2018BioinspiredExperimentb,Lee2022AnDiscovery}, integrating visual and linguistic data for advanced understanding and interaction has become an area of great interest. Applications include analysis of images, text or data mining. Emerging applications of such tools far exceed earlier computer vision methods~\cite{He2016DeepRecognition,He2015DeepRecognitionb}, such as image classification, as they provide more flexible and interactive methods of engaging with visual and text content. Moreover, the emergence of multi-agent AI systems~\cite{ghafarollahi2024protagents,Ni2024,buehler2024accelerating} requires enhanced scientific vision capabilities to analyze data, assess generated plots, or interpret and reason over inputs from autonomous experimentation~\cite{Stach2021,Abolhasani2023}. 

Such use cases of integrating different types of data, from text, to images, to content in scientific figures, tables and others, are critical especially in multidisciplinary areas of materials research such as bio-inspired materials, where researchers seek to integrate disparate sets of knowledge. Herein, generative AI is a particularly useful tool that can aid not only discovery but also finding engineering solutions for materials. This includes applications that include diverse properties that range from enhanced toughness to improved biological properties \cite{Qin2012BioinspiredGraphene,Zhang2015BioinspiredInnovation,Wu2017BioinspiredHydrogel,Dimas2013ToughTestingb,Milazzo2019AdditiveCompositesb,Studart2013BiologicalArchitectures,Qiu2022ExcellentFoams,Arevalo2023LearningMaterials,Lee2022AnDiscovery,Lu2023GenerativePropertiesb}.

A general framework of large language models (LLMs), here specifically to be understood as an approach that builds interaction graphs of complex input-output data with high capacity for in-context learning~\cite{Hu2023DeepScience}, has emerged as a promising approach in scientific analysis. Complex single and multi-modal models, such as those described in seminal works \cite{Vaswani2017AttentionNeedc,Touvron2023LlamaModels,OpenAI2023GPT-4Report,Chowdhery2022PaLM:Pathways,Jiang2023Mistral7B,Gunasekar2023TextbooksNeed,buehler2024accelerating,Stewart_Buehler_2024,naveed2024comprehensive,li2024uni}, offer innovative pathways for knowledge expansion, especially when coupled with multimodal capabilities~\cite{Carolan2024AModels,Buehler2024X-LoRA:Design,buehler2024accelerating,paligemma2024}. The potential of LLMs in fostering new hypotheses is further supported by research in applications to specific domain tasks or general intelligence examinations \cite{Bubeck2023SparksGPT-4,Buehler2023MechGPTModalities_fixed,Nejjar2023LLMsAnalysis,Luu2023BioinspiredLLM:Materials,Ge2023OpenAGI:Experts,Bhattacharjee2024INDUS:Applications,Huh2024TheHypothesis,Buehler2024X-LoRA:Design}. 

In this paper we focus on the development and application of Cephalo, a series of open-source multimodal vision large language models (V-LLMs), specifically in the context of bio-inspired design and mechanical properties of materials. The models are designed to bridge the gap between visual perception and language comprehension. Inspired by the intricate structures and mechanisms found in nature, particularly in bioinspired materials, Cephalo can interpret complex visual scenes and generate contextually accurate language descriptions, but we also demonstrate fine-tuning of the models to endow them with capabilities to make quantitative predictions about stress and atomic energy field statistics and failure dynamics. One of its key applications lies in analyzing and describing materials phenomena, such as failure and fracture, microstructures, and reasoning over biological and synthetic materials. Cephalo's architecture combines a vision encoder model with an autoregressive transformer to allow for tightly coupled visual and linguistic data processing. A variety of architectures and model sizes are explored to lay the foundation for a set of research-focused models.

\begin{figure}
    \centering
    \includegraphics[width=1.\linewidth]{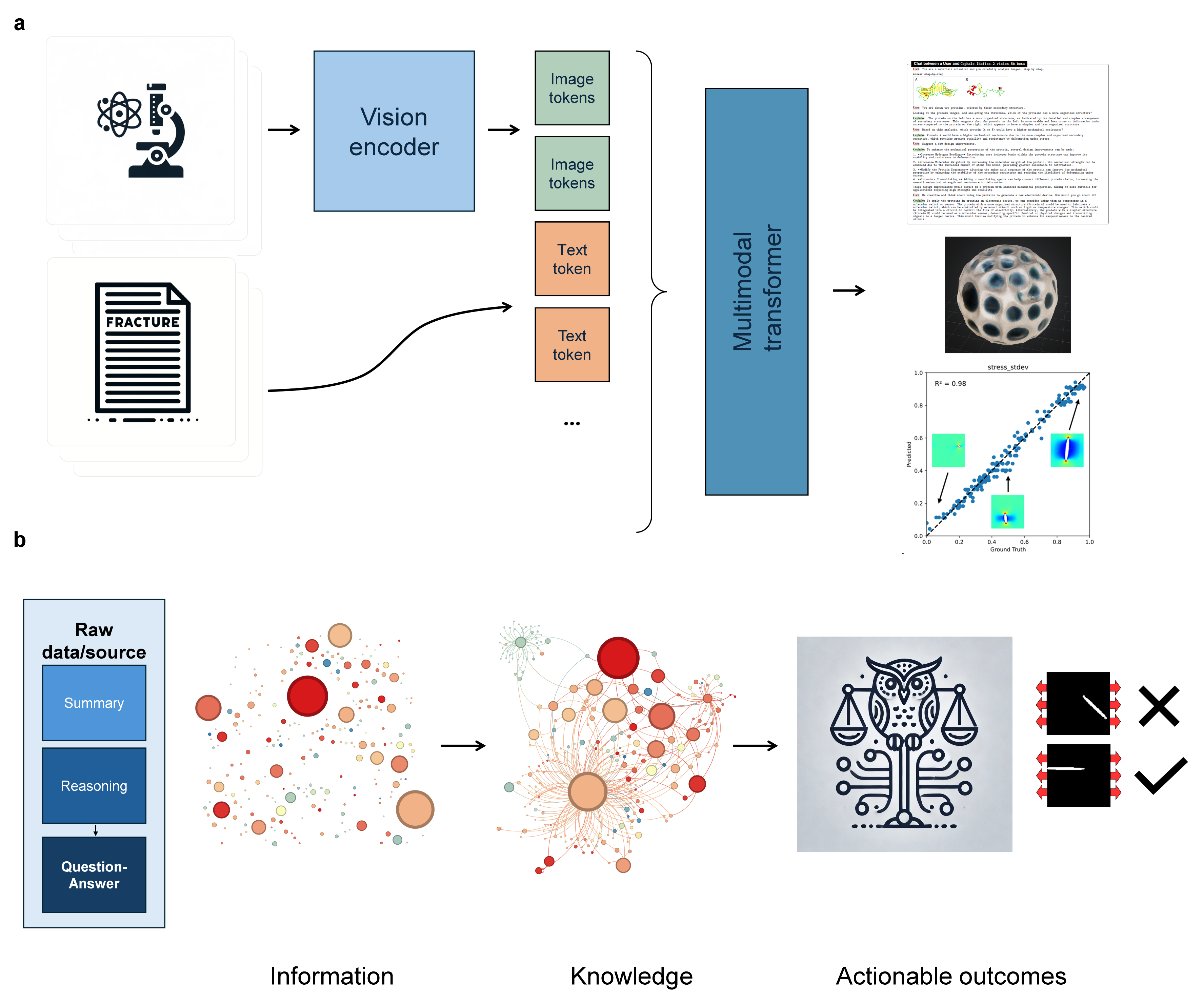}
    \caption{Overall approach used to develop the multi-modal vision LLM. Panel a: The model consists of a vision encoder (left side) that produces image tokens that are combined with text tokens in the autoregressive transformer model (center) with flexible outputs (right side). Panel b: Delineation of the development of the dataset to train the model, effectively transforming raw data into valuable insights. The data used for training consists of both text-only data (taken from~\cite{Luu2023BioinspiredLLM:Materials} and \cite{Buehler2023MechGPTModalities_fixed} as well as newly created image-text datasets, as well as data from molecular dynamics modeling.  In the process, raw data undergoes summarization and reasoning steps, evolving from scattered pieces of information into interconnected knowledge. This transformation then enables deeper understanding and effective decision-making, highlighting the model's capability to synthesize complex data into practical, actionable insights (e.g., design a material such that it does not fail under mechanical stress).}
    \label{fig:Fig_1}
\end{figure}

\subsection{Background and motivation}
Integrating vision and language is a cornerstone in scientific research. In particular, materials science relies heavily on images and contextual analysis, ranging from understanding microstructures to interpreting data. In the field of bio-inspired materials, a key aspect is the translation of abstract concepts (structures to features to mechanisms) across fields, and visual cues are paramount. While language processing has advanced the capabilities of  scientific AI in recent years, an important frontier is the establishment of models that can reason over a multitude of complex data that includes images and text, incorporating a set of complex scientific concepts. In this study, we explore the development and application of fine-tuned open source vision models, examined over a set of use cases that incorporate biological materials, fracture and engineering analysis, protein biophysics, and bio-inspired design. The study explores a variety of model architectures in different sizes and discusses how new models can be constructed by merging several smaller models into a larger one. 

Transformer models learn during pre-training and fine-tuning. However, they also show superior ability of in-context learning, where models adapt their responses based on the context provided in the prompt. This context can include various forms of data, examples, or any relevant information. The ability of models to perform a wide range of tasks without task-specific training data or fine-tuning is a testament to the power of in-context learning. In earlier work LLMs have demonstrated the capacity to synthesize sophisticated understanding, such as translating between languages not included in their training~\cite{gemini15report}. Further, a recent article~\cite{templeton2024scaling} discusses advancements in the interpretability of LLMs. In this work the researchers used a sparse autoencoder to extract numerous interpretable features from the Claude 3 Sonnet model, a mid-sized LLM. These features range from abstract behavioral indicators to safety-relevant ones. The study highlights the potential of these features in enhancing our understanding of LLM behavior. Specifically, the mapping of features such as ``Inner Conflict" indicates that the model learns and represents complex concepts. This is evidenced by several key factors. This includes, for instance, conceptual clusters, revealing the model's ability to organize features related to internal conflicts into distinct clusters such as balancing trade-offs, conflicting allegiances, and catch-22 scenarios suggests that it understands these concepts at a nuanced level. The authors~\cite{templeton2024scaling} showed that each cluster represents a different aspect of internal conflict, showing the model's capability to differentiate and relate these aspects contextually. When features are activated or manipulated, the model's responses change accordingly. For instance, amplifying a feature related to the ``Golden Gate Bridge" caused the model to focus excessively on that topic. Similarly, activating features related to inner conflicts altered how the model handled emotionally charged scenarios. This demonstrates that these features are not just random or superficial associations but deeply embedded representations that influence the model's understanding and output. The ability to represent, activate, and manipulate features associated with internal conflicts and other nuanced topics shows that the model has developed an internal understanding of these concepts, going beyond simple pattern matching to exhibit a form of conceptual learning and representation.

Other progress in the understanding of LLMs~\cite{Huh2024TheHypothesis} proposes that representations in large LLMs are increasingly converging, where the authors first review numerous examples from literature showing that neural networks across different domains and over time are developing similar ways of representing data. They then demonstrate that as models grow larger, both vision and language models measure distances between data points in increasingly similar ways. The hypothesis suggests that this convergence is moving toward a shared statistical model of reality, analogous to Plato's concept of an ideal reality, termed the ``platonic representation". The paper explores various selective pressures that might drive this convergence and discusses the implications, limitations, and counterexamples to their analysis.

These and other behaviors indicate that AI systems can be useful for knowledge discovery, especially when it comes to interrelating disparate areas of knowledge, drawing analogies, and predicting new insights from patters in seemingly unrelated observations. We postulate that providing the proper context to facilitate discovery is essential for harnessing the full potential of these advanced AI systems, and that the incorporation of image data, combined with text and scientific principles, is a critical next step over previous approaches. While earlier work has focused on multimodal forward and inverse problems in the context of scientific applications~\cite{Buehler2023MeLMProblemsc,Buehler2024X-LoRA:Design}, here we generalize this strategy to develop reasoning abilities over diverse types of image-text combinations. 

Figure~\ref{fig:Fig_1} illustrates the comprehensive approach used to develop the multimodal vision LLM. Figure~\ref{fig:Fig_1}(a) depicts the model's architecture, where a vision encoder processes images, generating image tokens, while text tokens are derived from associated textual data. These tokens are then combined and fed into a multimodal transformer model, which integrates and interprets the multimodal inputs to produce versatile outputs, such as textual summaries, graphical representations, and decision support insights. Figure~\ref{fig:Fig_1}(b) highlights the dataset development process, emphasizing the transformation of raw data into actionable insights. Initially, raw data and sources, including both text and image datasets, undergo summarization and reasoning steps. These steps convert scattered information into interconnected knowledge networks, facilitating deeper understanding and effective decision-making. The final stage showcases the model's ability to synthesize complex data into practical, actionable outcomes, such as designing materials that resist mechanical stress. This transformation process underscores the model's capability to translate raw data into valuable insights for informed decision-making, for a variety of taks ranging from reasoning to quantitative predictions, such as the effects of defects on stress distributions or crack initiation. 

\subsection{Outline of this paper}

In this paper we present both the development of a series of vision models with a new dataset, as well as various applications to show the usefulness and applications in materials science, specifically focused on bio-inspired analysis and design.  We provide an overview of the four major models developed as part of this effort in Table~\ref{table:cephalo_models}. Additional merged and integrated mixture-of-expert models, as well as fine-tuned models are developed as well. 
The Cephalo model lineup~\url{https://huggingface.co/lamm-mit/cephalo} consists of various versions designed for different purposes, three of which will be discussed in detail in this paper. 

The base model, \texttt{Cephalo-Phi-3-vision-128k-4b-alpha}, trained on \texttt{GPT-4o} distilled data from Wikipedia and scientific papers, serves as a foundation. An improved counterpart, \texttt{Cephalo-Phi-3-vision-128k-4b-beta}, incorporates additional data from image-text data distilled using \texttt{Idefics-2} combined with a large text-only corpus, providing nuanced responses and excellent reasoning. These 4b class models are complemented with 8b and larger models. The \texttt{Cephalo-Idefics-2-vision-8b-alpha} model, trained solely on \texttt{Idefics-2} data, offers concise and generally accurate answers. Lastly, the \texttt{Cephalo-Idefics-2-vision-8b-beta} model, enhanced with \texttt{GPT-4o} distilled data, delivers longer, well-reasoned responses but can face challenges with complex concepts. We further use  model merging  to develop larger V-LLMs at sizes 10b and 12b parameters by combining several smaller models.  

The \texttt{Cephalo-Idefics-2-vision-10b-alpha} model is a merged 10-billion-parameter model, featuring a total of 40 layers (32 original and 8 additional layers), with a checkpoint taken after the first epoch. It was trained on \texttt{GPT-4o} distilled image-text data sourced from Wikipedia and scientific papers. The \texttt{Cephalo-Idefics-2-vision-10b-beta} model, is based on a checkpoint taken after a second training epoch, trained on the same dataset. Additionally, the \texttt{lamm-mit/Cephalo-Idefics-2-vision-12b-alpha} model is a 12-billion-parameter merged model comprising 48 layers (32 original and 16 additional layers), with a checkpoint after the first epoch. This model was trained on a combined dataset derived from both \texttt{Idefics-2} and \texttt{GPT-4o} distillation of the paper corpus, including image-text data from Wikipedia and scientific papers.

We first provide a review of the basic architectures used here, then move to describe the dataset construction process, training methodology, and development of inference strategies. The paper concludes with a series of case studies and a critical discussion of strengths and weaknesses, quantitative assessments, including an outline of future research in the field.

\begin{table}[h!]
\scriptsize
\centering
\begin{tabular}{|m{2.5cm}|m{3cm}|m{4cm}|m{4cm}|}
\hline
\textbf{Model} & \textbf{Description} & \textbf{Training Data} & \textbf{Strengths/Weaknesses} \\
\hline
Cephalo-Phi-3-vision-128k-4b-alpha & Base version of the Cephalo-Phi-3 model & GPT-4o distilled image-text data from Wikipedia and scientific papers & Good baseline model, but struggles in longer conversations. Context length of 128,000 tokens. Only one image per prompt. \\
\hline
Cephalo-Phi-3-vision-128k-4b-beta & Improved version of the Cephalo-Phi-3 model & GPT-4o and Idefics-2 distilled image-text data from Wikipedia and scientific papers, as well as a large text-only corpus & Provides nuanced responses, with excellent reasoning. Context length of 128,000 tokens. Only one image per prompt.  \\
\hline
Cephalo-Phi-3-MoE-vision-128k-3x4b-beta & Mixture-of-expert model based on several smaller Cephalo-Phi-3 models. & GPT-4o and Idefics-2 distilled image-text data from Wikipedia and scientific papers, a large text-only corpus, as well image-to-LaTeX code dataset. & Diverse capabilities derived from the underlying expert models that make up the integrated model. Context length of 128,000 tokens. Only one image per prompt.  \\
\hline
Cephalo-Idefics-2-vision-8b-alpha &  Cephalo model based on Idefics-2 & Idefics-2 distilled image-text data from Wikipedia and scientific papers & Gives shorter answers, to the point, and generally accurate. Handles multiple images per prompt. \\
\hline
Cephalo-Idefics-2-vision-8b-beta & Cephalo model based on Idefics-2 & GPT-4o distilled image-text data from Wikipedia and scientific papers & Gives longer answers, with enhanced reasoning. Can struggle with complex concepts. Handles multiple images per prompt. \\
\hline
Cephalo-Idefics-2-vision-10b-alpha & Extended, larger/deeper Cephalo model based on Cephalo-Idefics-2-vision-8b-beta and Idefics-2 & GPT-4o distilled image-text data from Wikipedia and scientific papers & Performs well overall. Handles multiple images per prompt.\\
\hline
Cephalo-Idefics-2-vision-12b-alpha & Further extended, larger/deeper Cephalo model based on Cephalo-Idefics-2-vision-8b-beta and Idefics-2 & GPT-4o distilled image-text data from Wikipedia and scientific papers & Generally, does not perform as well as the 10b model. Handles multiple images per prompt.\\
\hline

\end{tabular}
\caption{Summary of Cephalo models and their characteristics (models with sizes of 4b, 8b, 10b and 12b parameters are developed, including a mixture-of-expert model with 9b parameters total, and 4b active parameters). Please see \url{https://huggingface.co/lamm-mit/cephalo} for an overview of the models. }
\label{table:cephalo_models}
\end{table}

\section{Results and discussion}

The method can be applied to a variety of use cases, as summarized in Table~\ref{table:cephalo_use_cases}.

\begin{table}[h!]
\scriptsize 
\centering
\begin{tabular}{|m{2.5cm}|m{3cm}|m{4.cm}|m{4.cm}|}
\hline
\textbf{Use Case} & \textbf{Description} & \textbf{Implementation} & \textbf{Impact} \\
\hline
Research Assistance & Literature Review and Data Extraction & Analyze scientific papers to extract key information and summarize findings on bioinspired materials. & Accelerates research by providing comprehensive literature reviews, saving time for researchers. \\
\hline
Image Analysis & Microscopic and Structural Analysis & Analyze images of natural materials to identify structural features for bioinspired material design. & Enhances understanding of natural structures, guiding the design of innovative materials. \\
\hline
Material Characterization & Comparative Analysis of Bioinspired Materials & Compare images and data of synthetic materials with their natural counterparts to assess design effectiveness. & Improves the evaluation process of new materials, ensuring they meet desired performance criteria. \\
\hline
Design Optimization & Suggesting Improvements in Material Design & Analyze current designs and suggest modifications to enhance material properties based on natural patterns. & Leads to the development of more efficient and effective bioinspired materials. \\
\hline
Educational Tool & Interactive Learning Platform & Create educational modules that use image-based learning to teach about bioinspired materials. & Facilitates learning and understanding of complex concepts in bioinspired material science. \\
\hline
Predictive Modeling & Predicting Material Properties & Predict properties of new materials by analyzing images and data from existing materials. & Accelerates the material development process by predicting properties before experimental validation. \\
\hline
Automated Documentation & Generating Reports and Visual Summaries & Generate detailed reports and visual summaries of research findings from experimental data and images. & Enhances communication of research findings, making it easier to share and publish results. \\
\hline
Innovation and Ideation & Brainstorming New Applications & Assist in brainstorming by providing insights and visual examples of bioinspired materials and their applications. & Sparks new ideas and applications, driving innovation in material science. \\
\hline
Quality Control & Monitoring Production Processes & Monitor production by analyzing real-time images to ensure quality and consistency of bioinspired materials. & Ensures high quality and consistency in the production of bioinspired materials, reducing defects. \\
\hline
Interdisciplinary Collaboration & Facilitating Cross-Disciplinary Research & Connect researchers from different fields by providing visual data and insights relevant to various disciplines. & Promotes innovative solutions and new discoveries through cross-disciplinary research and knowledge sharing. \\
\hline
New Material Discovery & Identifying Novel Materials & Use Cephalo to analyze vast datasets and images to discover new materials with unique properties. & Drives the discovery of novel materials, potentially leading to breakthroughs in various scientific fields. \\
\hline
Sustainability Analysis & Assessing Environmental Impact & Evaluate the environmental impact of bioinspired materials throughout their lifecycle. & Supports the development of sustainable materials and practices, contributing to environmental conservation. \\
\hline
Prediction of quantitative material properties & For example, based on microstructure input, predict complex features of fields that develop under application of mechanical loading &  Efficient fine-tuning of the model with molecular simulation data for quantitative predictions allows to predict stress and energy field statistics near defects & Provides rapid means to assess complex material behavior under static and dynamic conditions for decision making in discovery or design processes. \\
\hline
\end{tabular}
\caption{Summary of use cases for Cephalo in materials analysis, design, and related areas, quantitative analysis, along with their impact on materials science. The application cases range from complex visual-text reasoning to quantitative analysis of stress field and damage mechanics.}
\label{table:cephalo_use_cases}
\end{table}

We provide a brief discussion on dataset generation, albeit detailed methods are provided in the Materials and Methods section.

\begin{figure}
    \centering
    \includegraphics[width=1\linewidth]{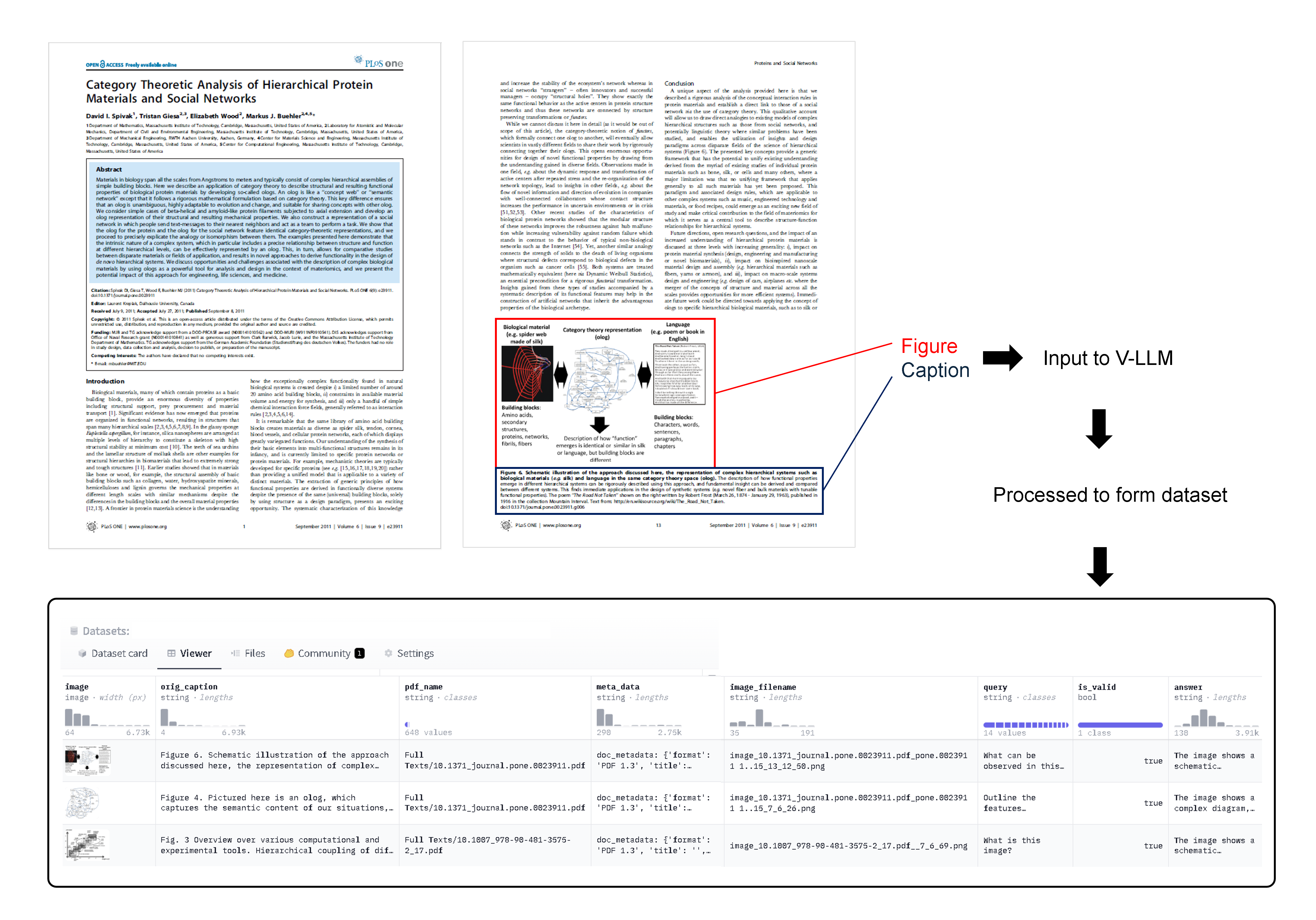}
    \caption{Visualization of the overall approach to generate datasets for training the vision model. Reproductions of two representative pages of the scientific article (here, \cite{Spivak2011CategoryNetworks}, reproduced with permission from PLOS ONE via a Creative Commons License.}
    \label{fig:Fig_2}
\end{figure}

We first discuss dataset generation. 
Figure~\ref{fig:Fig_2} depicts the overall approach to generate datasets for training the vision model. Reproductions of two representative pages of the scientific article (here, \cite{Spivak2011CategoryNetworks}. 

\begin{figure}
    \centering
    \includegraphics[width=1\linewidth]{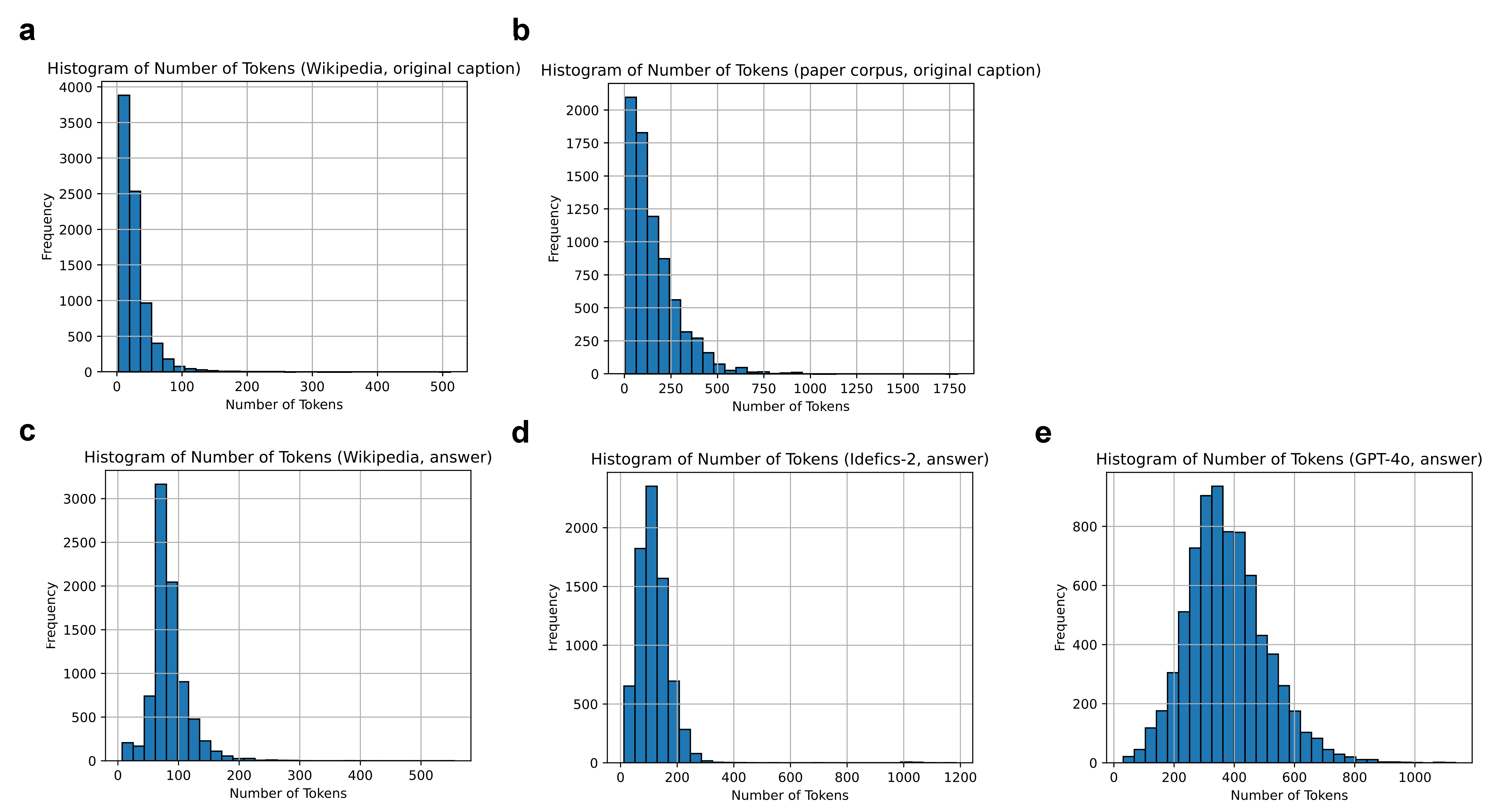}
    \caption{Histogram of the number of tokens for the image-text dataset, showing the source captions from Wikipedia (a) and the paper corpus (b). Panels c-e show the results processed with different vision-text models. Panel c shows the histogram of the token numbers for the processed image descriptions for the Wikipedia (done using Idefics-2). Panels d and e show the results for the paper corpus dataset, processed using \texttt{Idefics-2} (d) and \texttt{GPT-4o} (e). The \texttt{GPT-4o} dataset generally yields much longer descriptions. A detailed analysis of the content shows that it provides much enhanced reasoning and nuanced explanation of the image content.  All tokenization done using the \texttt{Phi-3-Vision} tokenizer~\cite{abdin2024phi3}.} 
    \label{fig:Fig_1001}
\end{figure}
\begin{figure}
    \centering
    \includegraphics[width=.6\linewidth]{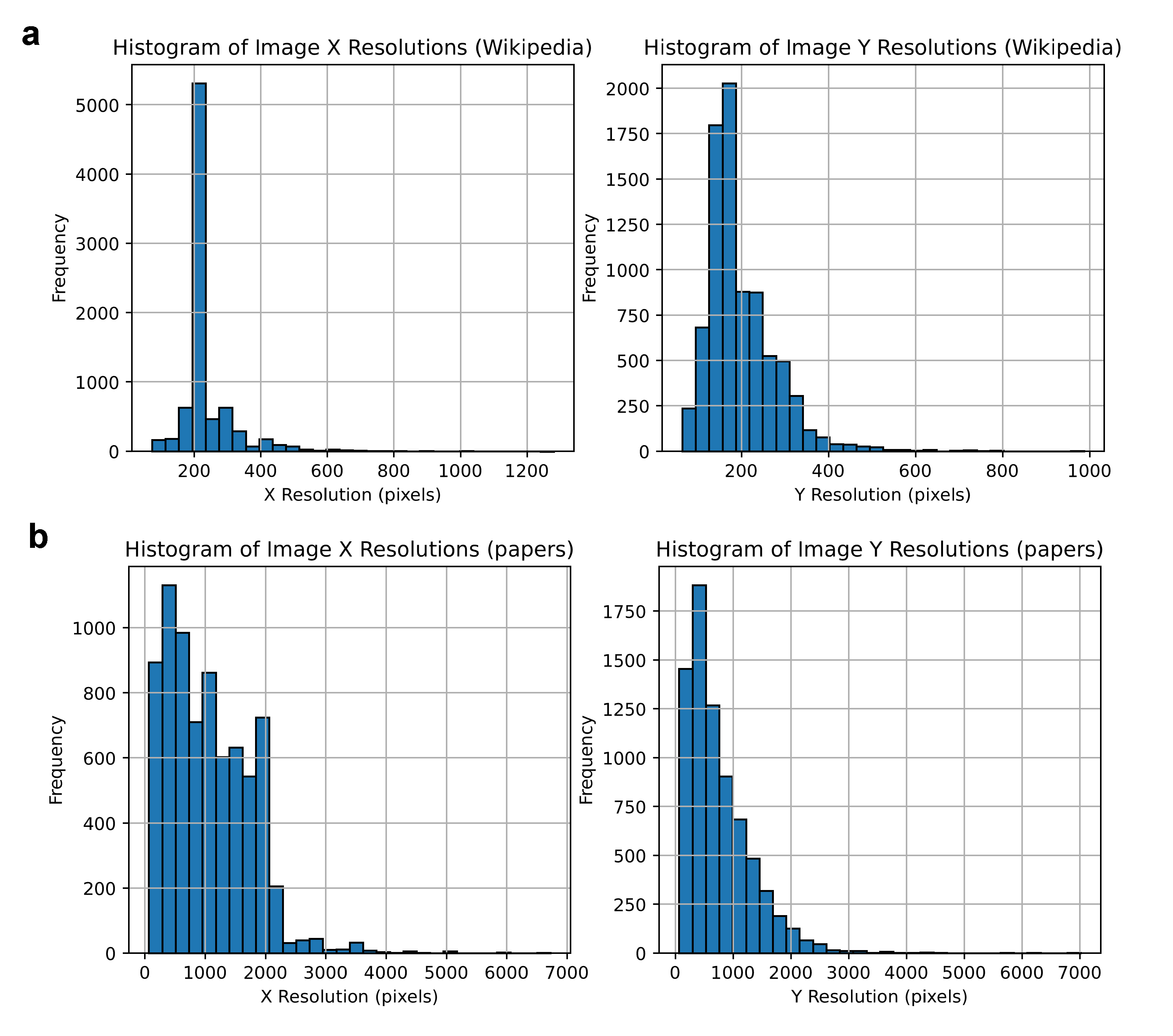}
    \caption{Histogram of the image resolutions extracted from Wikipedia (a) and the paper corpus (b), for $X$ and $Y$ directions, respectively (left/right column). All tokenization done using the Phi-3-Vision tokenizer~\cite{abdin2024phi3}.} 
    \label{fig:Fig_1002}
\end{figure}

During dataset generation, raw data extracted from Wikipedia and scientific papers is processed to add details, reasoning and logical definitions as we jointly consider images and associated text blocks (that is, captions). This typically results in an increase in the amount of information associated between images and text. We can measure this increase by studying image-text data before and after dataset generation, giving us a numerical assessment of this process. 
Figure~\ref{fig:Fig_1001} depicts histograms of the number of tokens for the image-text dataset, showing an analysis of the source captions extracted from Wikipedia (Figure~\ref{fig:Fig_1001}) and the scientific paper corpus (Figure~\ref{fig:Fig_1001}b). Tokens are the smallest units of text, such as words or subwords, into which text is divided for processing by language models, and hence a measure for the amount of content contained in text data. Figure~\ref{fig:Fig_1001}c-e show the resulting datasets obtained by processing using different vision-text models. Figure~\ref{fig:Fig_1001}c shows the histogram of the token numbers for the processed image descriptions for the Wikipedia (done using \texttt{Idefics-2)}. Further, Figures~\ref{fig:Fig_1001}d and e show the results for the scientific paper corpus dataset, processed using \texttt{Idefics-2} (Figure~\ref{fig:Fig_1001}d) and \texttt{GPT-4o} (Figure~\ref{fig:Fig_1001}e). The \texttt{GPT-4o} dataset generally yields much longer descriptions. A detailed analysis of the content beyond the numerical analysis of token histograms shows that it provides much enhanced reasoning and nuanced explanation of the image content.

Another key measure for the amount of information featured in the dataset is the resolution of images. Figure~\ref{fig:Fig_1002} depicts a histogram of the image resolutions extracted from Wikipedia (Figure~\ref{fig:Fig_1002}a) and the scientific paper corpus (Figure~\ref{fig:Fig_1002}b), for $X$ and $Y$ directions, respectively, in the left and right columns of the plots.

\begin{figure}
    \centering
    \includegraphics[width=1\linewidth]{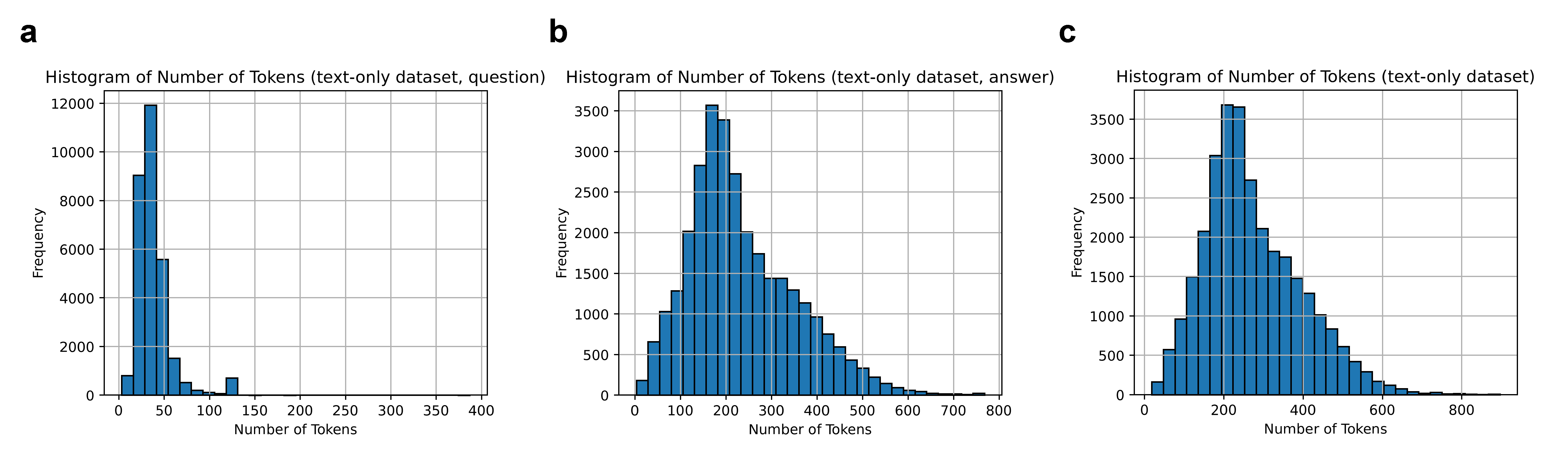}
    \caption{Histogram of the number of tokens for the text-only dataset, showing questions only (a), answers only (b), and combined question-answer (c). This dataset includes a corpus of knowledge extracted from scientific papers, books, and other sources in the area of biological materials, mechanics, and materials science.  All tokenization done using the Phi-3-Vision tokenizer.} 
    \label{fig:Fig_1000}
\end{figure}

One of the models is trained additionally on text-only data. For completeness we show an analysis of that dataset as well. Figure~\ref{fig:Fig_1000} shows a histogram of the number of tokens for the text-only dataset, showing questions only (Figure~\ref{fig:Fig_1000}a), answers only (Figure~\ref{fig:Fig_1000}b), and combined question-answer (Figure~\ref{fig:Fig_1000}c). This dataset includes a corpus of knowledge extracted from scientific papers, books, and other sources in the area of biological materials, mechanics, and materials science.

To assess the quality of responses and capabilities for complex materials analysis tasks, we provide a series of experiments that are applied consistently for all models examined here. All examples include multi-turn interactions with a User (human), with consistent questioning for all cases studied. The specific cases include:
\begin{enumerate}
    \item Analysis of a fracture scenario, where the model interacts with a human to assess two distinct fracture scenarios (brittle versus more ductile, distributed fracture), including an assessment of fracture mechanism and likely fracture properties of the material.
    \item Analysis of two proteins with distinct structures (highly organized beta-sheet protein on the one hand, and a partially disordered protein on the other hand). The model is tasked to assess the structure of the proteins based on their visualization, as well as estimate resulting mechanical properties. 
    \item Analysis of an image of ants building an ant bridge via complex coordination, and assessing the content of the image in the context of applications in the development of multi-agent AI systems. 
    \item Analysis of two images of distinct objects, including (1) an image of ants building an ant bridge via complex coordination, and (2) a micrograph of a pollen particle. The model is tasked to analyze both images and then develop bioinspired design concepts and research ideas based on them.  
\end{enumerate}

\subsection{Cephalo-8b model series}

We build our vision LLM based on a pretrained model, Idefics2. This base model is an 8-billion-parameter vision-language model that processes sequences of texts and images to generate text responses. It excels in tasks such as visual question answering, image description, document information extraction, and basic arithmetic. The baseline version offers OCR capabilities, the ability to manipulate images in their native resolutions and aspect ratios, and a simplified integration of visual features into the language backbone. The model utilizes a mixture of openly available datasets and is fine-tuned on task-oriented data, making it highly versatile for various multimodal applications. The parent models used are \url{google/siglip-so400m-patch14-384} and \url{mistralai/Mistral-7B-v0.1}.

The architecture processes images in their native resolutions and aspect ratios using the NaViT approach~\cite{Dehghani2023PatchResolution}, avoiding resizing to fixed squares. Enhanced OCR capabilities and improved performance on charts, figures, and documents were achieved with targeted training data. The model does not use gated cross-attentions, and instead uses a vision encoder with Perceiver pooling~\cite{JaeglePerceiverOutputs_fixed} and MLP modality projection, concatenating visual features with text embeddings. The model was trained in multiple stages and leverages high-resolution images and various datasets, followed by instruction fine-tuning on curated vision-language datasets. In our work, we expand the fine-tuning to endow the model with enhanced capabilities to deal with scientific images. 

As shown in \ref{fig:Fig_1}a the model processes and integrates visual and textual data. It starts with the Vision Encoder, which processes input images to extract visual features. These features are then combined with text embeddings in the Vision-Language Connector using learned Perceiver pooling and MLP modality projection. This combined visual-text sequence is then fed into the LLM Decoder, which generates coherent text responses. This architecture thereby allows for effective handling of multimodal tasks such as visual question answering and document analysis.
 
The integration of Cephalo into the study and application of bioinspired materials presents numerous advantages, as summarized in Table~\ref{table:cephalo_use_cases}. Key insights reveal that Cephalo can significantly accelerate research through automated literature reviews and data extraction, saving valuable time for researchers. Its ability to analyze microscopic and structural images enhances our understanding of natural materials, guiding the design of innovative bioinspired solutions. By predicting material properties and suggesting design optimizations, Cephalo streamlines the development process, leading to more efficient and effective materials. Additionally, its role in educational platforms facilitates learning, while its capabilities in generating reports and monitoring production processes ensure high-quality outcomes (Table~\ref{tab:educational_goals} depicts a detailed outline of educational goals, a description, and learning objectives for basic, intermediate and advanced aspects). Cephalo also fosters interdisciplinary collaboration by connecting researchers from different fields and providing insights relevant to various disciplines, driving innovative solutions and new scientific discoveries. Furthermore, its potential in identifying novel materials and assessing their sustainability impacts supports the development of breakthrough materials and environmentally conscious practices. Overall, Cephalo's application in bioinspired material science fosters innovation, improves evaluation processes, and enhances the quality and consistency of material production.

We look at two models, \texttt{Cephalo-Idefics-2-vision-8b-alpha} (based on \texttt{Idefics-2} training data) and \texttt{Cephalo-Idefics-2-vision-8b-beta} (based on \texttt{GPT-4o} training data). Details about the training data and method of training is provided in the Materials and Methods section. 

Both models perform well, but show different types of responses, and could be useful depending on required scenarios of application. Reflecting the nature of the training data with generally shorter responses (see, e.g. Figure~\ref{fig:Fig_1001}d-e), \texttt{Cephalo-Idefics-2-vision-8b-alpha} provides relatively shorter replies. In contrast, \texttt{Cephalo-Idefics-2-vision-8b-beta} yields more elaborate responses. We find that sometimes, however, the model struggles to correctly capture logic and accuracy. 

Text Boxes \ref{TexBox:TB1} (fracture mechanics), \ref{TexBox:TB2} (protein mechanics), \ref{TexBox:TB3} (bio-inspired AI systems) and \ref{TexBox:TB4} (bio-inspired materials and integration of disparate concepts) show results for the experiments conducted with \texttt{Cephalo-Idefics-2-vision-8b-alpha}. 


\begin{figure}[htbp]
\begin{LLMbox}{Chat between a User and \texttt{Cephalo-Idefics-2-vision-8b-alpha}}
\linespread{0.4}\selectfont
\begin{lstlisting} 
(*@\bf\hlred{User}@*): You are a materials scientist and you carefully analyze images, step by step.

Answer briefly.
(*@\includegraphics[width=.33\linewidth]{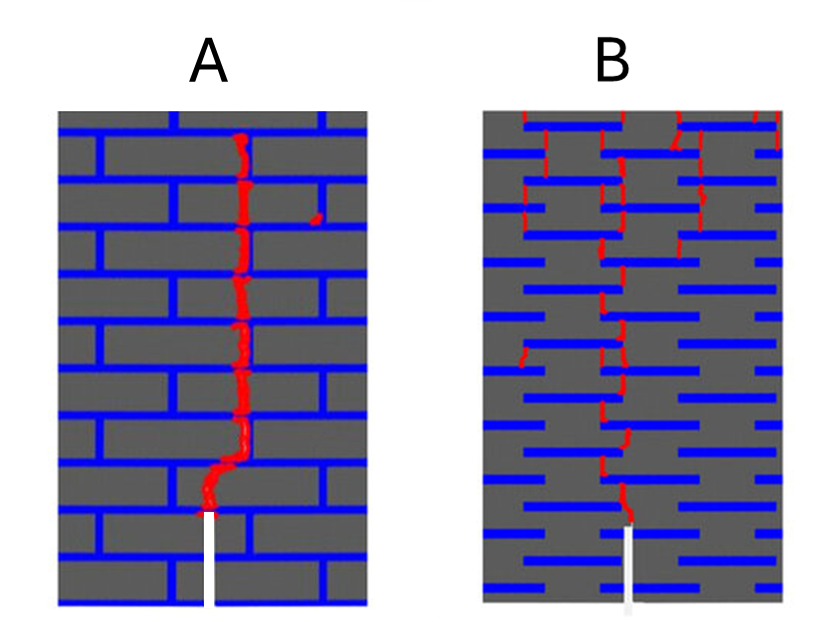}@*)
(*@\bf\hlred{User}@*): You are shown two fracture scenarios.

Describe the differences in damage. Note, cracked material is shown in red color. The initial crack is the white notch at the bottom.

(*@\bf \hl{Cephalo}@*):  In the first image, the crack propagates in a straight line through the material, resulting in a single crack. In the second image, the crack splits into multiple branches, creating a complex network of cracks. The initial crack is the same white notch at the bottom in both images.

(*@\bf\hlred{User}@*): Which case has more distributed crack path?

(*@\bf \hl{Cephalo}@*): The second image with multiple crack branches has a more distributed crack path compared to the first image with a single crack.

(*@\bf\hlred{User}@*): What steps could I use to toughen the brittle composite?

(*@\bf \hl{Cephalo}@*): To toughen the brittle composite, several steps can be taken:

1. Introduce crack deflection mechanisms: This can be achieved by incorporating materials with different properties that can divert the crack path and reduce its size.

2. Enhance crack bridging: The use of fibers or particles can help bridge the crack and prevent it from growing.

3. Improve interfacial adhesion: Strong bonding between the phases can prevent crack propagation and improve toughness.

4. Increase crack-stopping mechanisms: Introducing materials that can absorb energy during deformation can help prevent crack growth and improve toughness.

By implementing these steps, the toughness of the brittle composite can be significantly improved.

(*@\bf\hlred{User}@*): Provide a bio-inspired strategy that uses honeycombs.

(*@\bf \hl{Cephalo}@*):  One bio-inspired strategy that uses honeycombs is the use of a hierarchical structure with multiple levels of organization. This approach mimics the natural structure of honeycombs, which provides high strength and toughness through a combination of crack deflection, crack bridging, and energy absorption mechanisms. The use of honeycombs can improve the toughness of brittle composites by providing a more distributed crack path and reducing crack growth.

(*@\bf\hlred{User}@*):  Summarize key insights in a few bullet points.

(*@\bf \hl{Cephalo}@*): - The second image with multiple crack branches has a more distributed crack path compared to the first image with a single crack.
- The second image is likely to have higher toughness due to the complex network of cracks that can dissipate more energy during deformation.
- To improve the toughness of brittle composites, strategies such as introducing crack deflection mechanisms, enhancing crack bridging, improving interfacial adhesion, and increasing crack-stopping mechanisms can be implemented.

\end{lstlisting}
\end{LLMbox}
\textboxcaption{Chat between a User and \texttt{Cephalo-Idefics-2-vision-8b-alpha}, focused on assessing two fracture scenarios. Image reproduced with permission from~\cite{Sen2011StructuralBlocks}, using a Creative Common license (\url{http://creativecommons.org/licenses/by-nc-nd/3.0/}).}
\label{TexBox:TB1}
\end{figure}

The conversation in Text Boxe~\ref{TexBox:TB1} involves an analysis of two images (A and B) showing fracture scenarios. The images depict cracks in materials, with cracked areas in red and an initial white notch at the bottom. Cephalo is able to provide insights into the differences in crack propagation and suggests methods to toughen brittle composites. The discussion includes bio-inspired strategies and key insights are summarized in bullet points, as requested by the user. Key insights developed by the model include:
\begin{itemize}
    \item The second image with multiple crack branches has a more distributed crack path than the first image with a single crack. This is correct, and represents a major difference between the two scenarios. 
    \item The second image is likely to exhibit higher toughness due to the complex network of cracks that can dissipate more energy. The model correctly identifies this aspect. 
    \item The model correctly identifies strategies to toughen brittle composites include introducing crack deflection mechanisms, enhancing crack bridging, improving interfacial adhesion, and increasing crack-stopping mechanisms.
    \item A bio-inspired strategy using honeycombs involves a hierarchical structure that mimics natural honeycombs for improved toughness.
\end{itemize}

It can be seen that Cephalo provides a clear comparison of the two fracture scenarios, highlighting the differences in crack propagation effectively. The model was able to make actionable suggestions. For instance, it suggests  steps to toughen brittle composites that are practical and based on sound materials science principles. The honeycomb strategy is an innovative approach that ties biological concepts to materials. The key insights are summarized succinctly, making it easy to grasp the main points of the conversation. This shows a capability to accurately comprehend the conversation and successfully interact with the human user over multiple turns.

\begin{figure}[htbp]
\begin{LLMbox}{Chat between a User and \texttt{Cephalo-Idefics-2-vision-8b-alpha}} 
\linespread{0.4}\selectfont
\begin{lstlisting} 
(*@\bf\hlred{User}@*): You are a materials scientist and you carefully analyze images, step by step.

Answer step-by-step.
(*@\includegraphics[width=.5\linewidth]{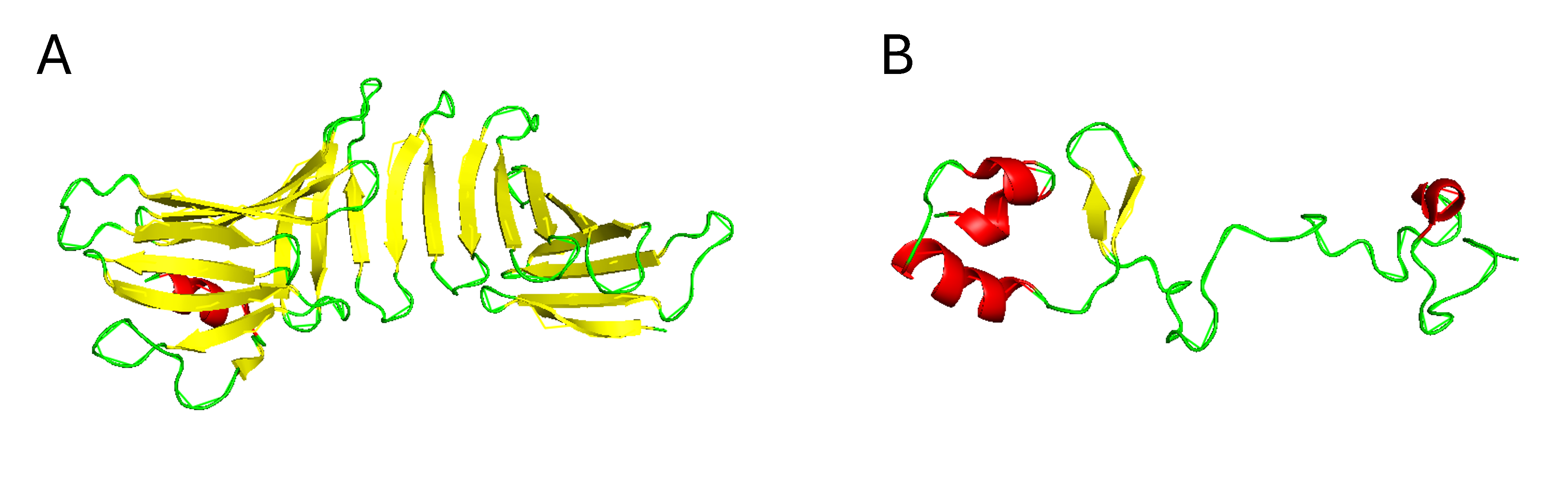}@*)
(*@\bf\hlred{User}@*): You are shown two proteins, colored by their secondary structure.

Looking at the protein images, and analyzing the structure, which of the proteins has a more organized structure?

(*@\bf \hl{Cephalo}@*):   Based on the image, it is evident that one protein has a more organized structure compared to the other. This can be attributed to various factors such as the presence of specific amino acid sequences, post-translational modifications, or environmental conditions that influence the protein's conformation and stability. The more organized structure may result in improved function or stability for the protein, depending on its specific role in the cellular context.

(*@\bf\hlred{User}@*): Based on this analysis, which protein (A or B) would have a higher mechanical resistance?

(*@\bf \hl{Cephalo}@*): Based on the analysis, protein A is likely to have a higher mechanical resistance compared to protein B due to its more organized structure. This could be attributed to factors such as the presence of specific amino acid sequences, post-translational modifications, or environmental conditions that influence the protein's conformation and stability. The more organized structure may result in improved function or stability for the protein, depending on its specific role in the cellular context.

(*@\bf\hlred{User}@*): Suggest a few design improvements.

(*@\bf \hl{Cephalo}@*): To improve the design of protein A, one could consider incorporating specific amino acid sequences or post-translational modifications that enhance its stability and mechanical resistance. Additionally, optimizing environmental conditions such as pH, temperature, or ionic strength during protein synthesis and assembly could also contribute to improved function and stability.

(*@\bf\hlred{User}@*): Be creative and think about using the proteins to generate a new electronic device. How would you go about it?

(*@\bf \hl{Cephalo}@*):  To create a new electronic device using proteins, one could consider incorporating these proteins into a conducting polymer matrix or a biocompatible substrate. This would allow for the formation of protein-based nanowires that can be used as electrodes or sensors in various applications such as biosensors, biofuel cells, or neural prosthetic devices. The use of proteins in electronic devices could potentially offer improved performance, biocompatibility, and functionality compared to traditional materials due to their unique properties and potential for self-assembly and molecular recognition.
\end{lstlisting}
\end{LLMbox}
\textboxcaption{Chat between a User and  \texttt{Cephalo-Idefics-2-vision-8b-alpha}, focused on assessing two proteins. The model is tasked to figure out structural features and to assess likely mechanical properties based on the images. The two proteins considered are PDB IDs 5B2A (crystal structure of a mutant of OspA, a beta-sheet rich structure with around 250 residues) and 1WVK (NMR solution structure of the partially disordered protein At2g23090 from Arabidopsis thaliana). }
\label{TexBox:TB2}
\end{figure}

The conversation in Text Box~\ref{TexBox:TB2} involves a user asking the model to analyze images of two proteins and evaluate their structural organization and mechanical resistance. The model provides insights into the more organized protein and suggests design improvements as well as creative applications for using proteins in electronic devices. The discussion focuses on the structural features and potential uses of the proteins. Key insights developed are as follows:
\begin{itemize}
    \item Protein A has a more organized structure compared to protein B. This is an important aspect that was correctly identified by the model. 
    \item The model correctly predicts that protein A is likely to have higher mechanical resistance due to its organized structure. 
    \item When asked about design improvements, the model focuses on protein A. Specific ideas developed include incorporating specific amino acid sequences, post-translational modifications, and optimizing environmental conditions.
    \item A creative application involves using proteins in electronic devices by incorporating them into a conducting polymer matrix or biocompatible substrate to form protein-based nanowires.
\end{itemize}

Analyzing the conversation as a whole, we find that the model provides a clear comparison of the structural organization of the two proteins, effectively identifying the more organized protein. The design improvements for protein A are practical and based on sound principles of protein chemistry. The idea of using proteins in electronic devices shows creativity and an interdisciplinary approach, suggesting novel uses for biological materials in technology. As before, the key insights are summarized succinctly, making it easy to understand the main points of the conversation. 

\begin{figure}[htbp]
\begin{LLMbox}{Chat between a User and \texttt{Cephalo-Idefics-2-vision-8b-alpha}}
\linespread{0.4}\selectfont
\begin{lstlisting} 
(*@\bf\hlred{User}@*): You are a materials scientist and you carefully analyze images, step by step.

Answer briefly.

(*@\includegraphics[width=.33\linewidth]{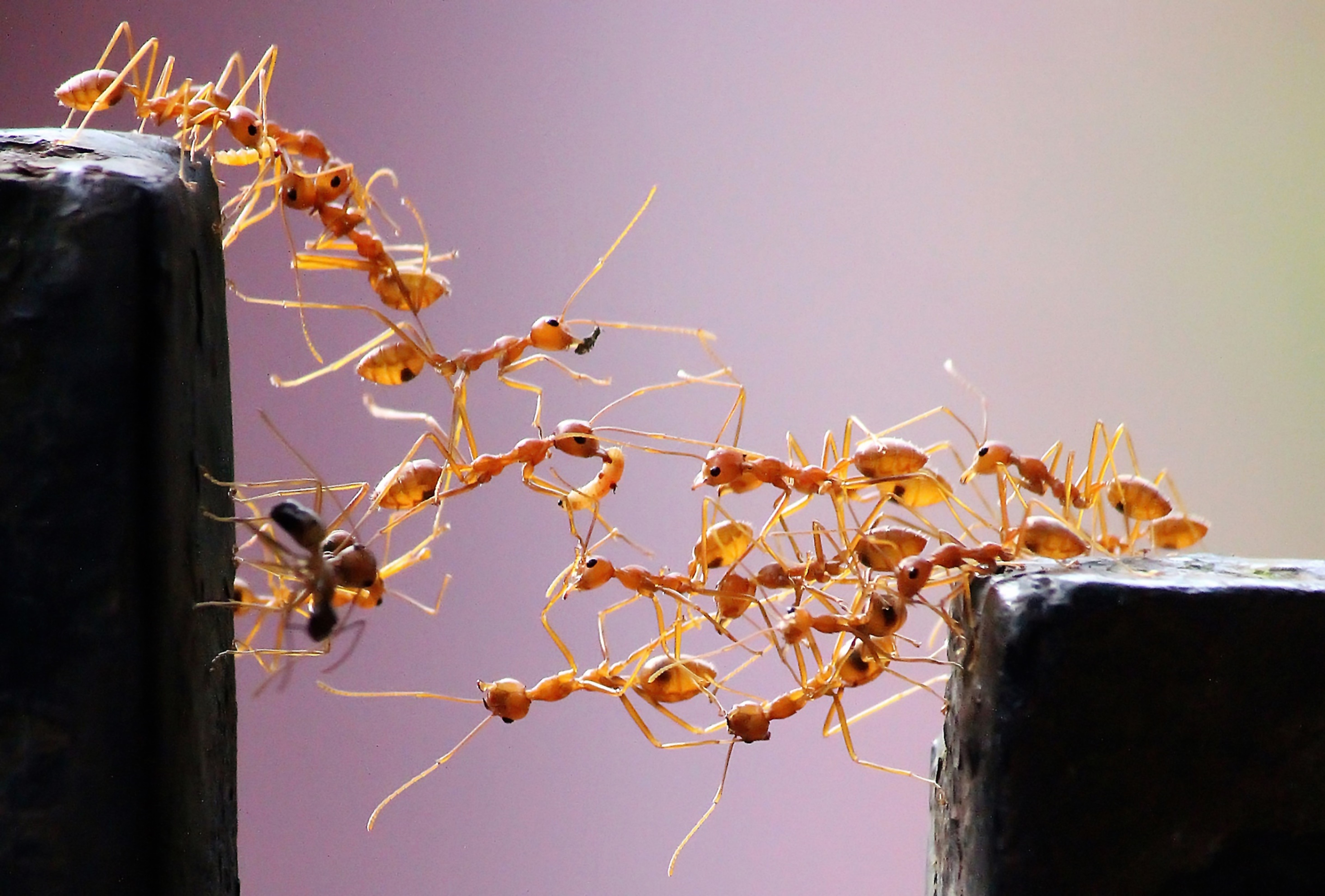}@*)

(*@\bf\hlred{User}@*): Carefully study this photo.

What do you see?

(*@\bf \hl{Cephalo}@*): The image shows a group of orange ants climbing up a black pole. The ants are moving in a line, with some ants leading the way and others following behind. The black pole is located in the center of the image, and the ants are scattered around it, some on the pole and others on the ground. The background is blurred, making the ants and the pole stand out.

(*@\bf\hlred{User}@*): How can the phenomenon be used to inform, or explain multi-agent AI systems?

(*@\bf \hl{Cephalo}@*): The image of ants climbing up a pole can be used to inform or explain multi-agent AI systems by demonstrating how individual agents can work together to achieve a common goal. In this case, the ants are cooperating to climb the pole, with each ant contributing its own strength and expertise to the task. This behavior can be observed in multi-agent AI systems, where multiple agents work together to solve complex problems or complete tasks. The image provides a visual representation of the concept of collaboration and teamwork in multi-agent AI systems.


(*@\bf\hlred{User}@*): Provide a brief outlook to future research one could do.

(*@\bf \hl{Cephalo}@*): One could conduct research on the biomechanics of ant climbing behavior to understand how they are able to climb vertical surfaces using their unique body structure and locomotion mechanisms. This could involve studying the ants' leg structure, adhesive properties, and sensory systems that enable them to climb efficiently. The findings from this research could inform the development of new robotics and artificial intelligence systems that mimic ant climbing behavior for various applications such as search and rescue operations or exploration of difficult terrains.

(*@\bf\hlred{User}@*): What type of deep learning model could be developed based on collective intelligence of ants?

(*@\bf \hl{Cephalo}@*):  A deep learning model could be developed based on the collective intelligence of ants by analyzing their behavior and interactions during the climbing process. This model would aim to mimic the ants' ability to cooperate and solve problems efficiently. The model could be trained using large datasets of ant climbing videos or images, which would provide valuable information about the ants' movements, interactions, and decision-making processes. By analyzing these data, researchers could identify patterns and trends in ant behavior that could be used to develop a deep learning model that mimics the ants' collective intelligence. This model could have potential applications in various fields such as robotics, artificial intelligence, and computer vision.

\end{lstlisting}
\end{LLMbox}
\textboxcaption{Chat between a User and \texttt{Cephalo-Idefics-2-vision-8b-alpha}, analyzing an image of ants forming a bridge. Image reproduced from: \url{https://www.shutterstock.com/image-photo/unity-strength-1011406435}.}
\label{TexBox:TB3}
\end{figure}

The next example shown in Text Box~\ref{TexBox:TB3} focuses on analysis of an image of ants and relating it to multi-agent AI systems. The model describes the image and draws parallels between ant behavior and multi-agent AI systems, suggesting research directions and potential applications of a deep learning model based on collective intelligence of ants. The discussion highlights how the observed phenomenon can inform AI research and development. Key Insights include: 
\begin{itemize}
    \item The model correctly identifies the salient features in the image, in that it shows orange ants climbing a black pole, demonstrating teamwork and collaboration. However, it misses to understand the focus on an "ant bridge". 
    \item The model correctly associates the phenomenon of insect collaboration as a means to be used to explain multi-agent AI systems where individual agents work together to achieve a common goal.
    \item The model proposes future research that could focus on the biomechanics of ant climbing behavior to inform robotics and AI system development.
    \item A deep learning model based on the collective intelligence of ants could be developed to mimic their cooperative problem-solving abilities.
\end{itemize}

The model provides an overall accurate description of the image, effectively highlighting the ants' collaborative behavior. The analogy between ant behavior and multi-agent AI systems is well-explained, making the concept accessible and relevant. The model's suggestions for future research and the development of a deep learning model are insightful, showcasing the potential for interdisciplinary applications in robotics and AI.

\begin{figure}[htbp]
\begin{LLMbox}{Chat between a User and \texttt{Cephalo-Idefics-2-vision-8b-alpha}}
\linespread{0.4}\selectfont
\begin{lstlisting} 
(*@\bf\hlred{User}@*): You are a materials scientist and you carefully analyze images, step by step.

Answer concisely, and accurately.
(*@\includegraphics[width=.5\linewidth]{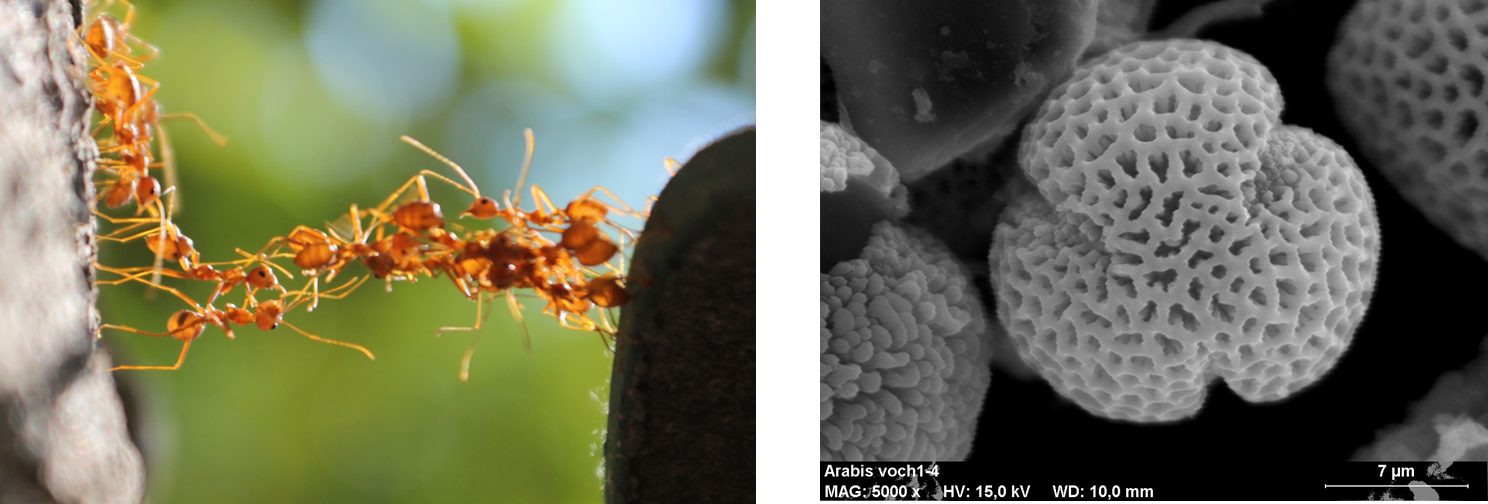}@*)
(*@\bf\hlred{User}@*): You are shown two photos of two very different objects. Carefully study the two photos.

What do you see?

(*@\bf \hl{Cephalo}@*): The first image shows a group of ants climbing up a tree trunk. The ants are moving in a line, with some ants leading the way and others following behind. The tree trunk is rough and textured, providing a challenging surface for the ants to climb.

The second image shows a close-up view of a cell wall in a plant. The cell wall is made up of small holes, which are arranged in a repeating pattern. The cell wall appears to be made of a material that is hard and rigid, providing support to the plant.

Both images show examples of materials interacting with their environment in different ways. The ants are adapting to the rough surface of the tree trunk, while the cell wall is designed to provide strength and stability to the plant.

(*@\bf\hlred{User}@*): Which of the two images would be better suited to create fracture-resistant materials?

(*@\bf \hl{Cephalo}@*): The second image, showing the cell wall in a plant, would be better suited to create fracture-resistant materials. This is because the cell wall is made of a material that is hard and rigid, providing strength and stability to the plant. The repeating pattern of small holes in the cell wall also suggests a level of complexity and design that could be adapted for creating materials with improved fracture resistance.

(*@\bf\hlred{User}@*): What about the other one, what may it be useful for?

(*@\bf \hl{Cephalo}@*): The first image, showing the ants climbing up a tree trunk, could be useful for studying animal locomotion and movement on rough surfaces. It may also provide insights into the ants' ability to adapt to challenging environments.

(*@\bf\hlred{User}@*): How can these concepts be used to inform new materials?

(*@\bf \hl{Cephalo}@*): The concepts shown in these images can be used to inform new materials by studying the properties and structures of the materials in the images. For example, the cell wall in the plant could provide insights into the design of materials with improved fracture resistance, while the ants' ability to adapt to rough surfaces could inspire new materials that can move and interact with their environment in similar ways.

(*@\bf\hlred{User}@*): Next, specifically develop a research idea that integrates the concepts seen in the two images.

(*@\bf \hl{Cephalo}@*): One potential research idea could be to study the cell walls of plants and their ability to provide strength and stability while also being flexible enough to adapt to challenging environments. This could involve analyzing the material properties of the cell walls and their structure, as well as testing their performance under different conditions. The ants' ability to climb up rough surfaces could also be incorporated into the research, with a focus on understanding how they adapt to their environment and how this could inform the design of new materials with improved fracture resistance and flexibility.

(*@\bf\hlred{User}@*): Provide a step-by-step outlook to future research.

(*@\bf \hl{Cephalo}@*):  The first step in future research would be to collect data on the cell walls of plants and their properties, as well as the ants' ability to climb up rough surfaces. This could involve conducting experiments or analyzing existing data sets. The next step would be to compare the properties of the cell walls with those of other materials, such as metals or polymers, to identify similarities and differences. Finally, the research could be used to inform the design of new materials that combine the strength and stability of the cell walls with the flexibility and adaptability of the ants' movement on rough surfaces.

(*@\bf\hlred{User}@*): Now write a detailed prompt for an image generation AI, that allows me to visualize the material. The prompt is:

(*@\bf \hl{Cephalo}@*):  Visualize the material as a hard and rigid structure with a repeating pattern of small holes, similar to the cell wall in a plant. The material should also exhibit flexibility and adaptability, allowing it to move and interact with its environment in a similar way as the ants climbing up the rough surface.

\end{lstlisting}
\end{LLMbox}
\textboxcaption{Chat between a User and \texttt{Cephalo-Idefics-2-vision-8b-alpha}, analyzing an image of ants forming a bridge and a microscopy image of a pollen particle. The model is tasked to reason over both and develop bio-inspired design concepts. Left image, reproduced from \url{https://en.m.wikipedia.org/wiki/File:AntBridge_Crossing_10.jpg}, based on a CC BY-SA 4.0, Igor Chuxlancev. Right image, reproduced from \url{https://en.wikipedia.org/wiki/Pollen\#/media/File:Arabis_voch1-4.jpg}, CC BY-SA 3.0, Marie Majaura.}
\label{TexBox:TB4}
\end{figure}


In the final experiment depicted in Text Box~\ref{TexBox:TB4} the model is tasked to analyze two images: One showing ants near a tree trunk and a gap, and the other one showing a close-up view of pollen particle. Cephalo describes the images, compares their suitability for creating fracture-resistant materials, and suggests potential applications for the observed phenomena. The discussion progresses to developing a research idea that integrates the concepts from both images and provides a step-by-step outlook for future research. Summarizing the most important insights, we find: 
\begin{itemize}
    \item The model identifies correctly that the first image shows ants climbing a rough tree trunk, and associates with a demonstration of adaptation to challenging conditions, specifically material surfaces.
    \item The model identifies the second image as a plant cell wall with a repeating pattern of small holes, providing strength and stability. This is not entirely correct, as the image actually shows a pollen particle. Still, critical structural features are correctly identified. 
    \item The model correctly identifies the second image on the right as being better suited for creating fracture-resistant materials due to its hard, rigid and porous structure.
    \item The model predicts that the ants' climbing behavior can be useful for studying locomotion and movement on rough surfaces.
    \item Future research is identified that could integrate the properties of plant cell walls and ants' adaptive movement to develop new materials with both strength and flexibility.
\end{itemize}
Overall, the model provides a clear and accurate description of both images, effectively highlighting the key features of the ants' climbing behavior and the plant cell wall structure. The analysis identifies the porous biological structure as better suited for creating fracture-resistant materials and suggests practical applications for the ants' behavior. The proposed research idea demonstrates creative thinking by integrating the properties of both images to inform the design of new materials. The step-by-step outlook for future research is well-structured and provides a clear road-map for further investigation.
 
We continue with the same experiment, but using the \texttt{Cephalo-Idefics-2-vision-8b-beta} model trained based on the \texttt{GPT-4o} based dataset. 
Text Boxes \ref{TexBox:TB11} (fracture mechanics), \ref{TexBox:TB12} (protein mechanics), \ref{TexBox:TB13} (bio-inspired AI systems) and \ref{TexBox:TB14} (bio-inspired materials and integration of disparate concepts) show results for the experiments conducted with \texttt{Cephalo-Idefics-2-vision-8b-beta}.  Compared to the earlier examples, the responses of this model tend to be longer, more elaborate and feature better reasoning. This directly reflects the more sophisticated data used to train this model. The responses also feature better formatting (e.g., see Text Box \ref{TexBox:TB12} where bold typeset font is used in the enumerated list, using Markup language). 


\begin{figure}[htbp]
\begin{LLMbox}{Chat between a User and \texttt{Cephalo-Idefics-2-vision-8b-beta}}
\linespread{0.4}\selectfont
\begin{lstlisting} 
(*@\bf\hlred{User}@*): You are a materials scientist and you carefully analyze images, step by step.

Answer briefly.
(*@\includegraphics[width=.33\linewidth]{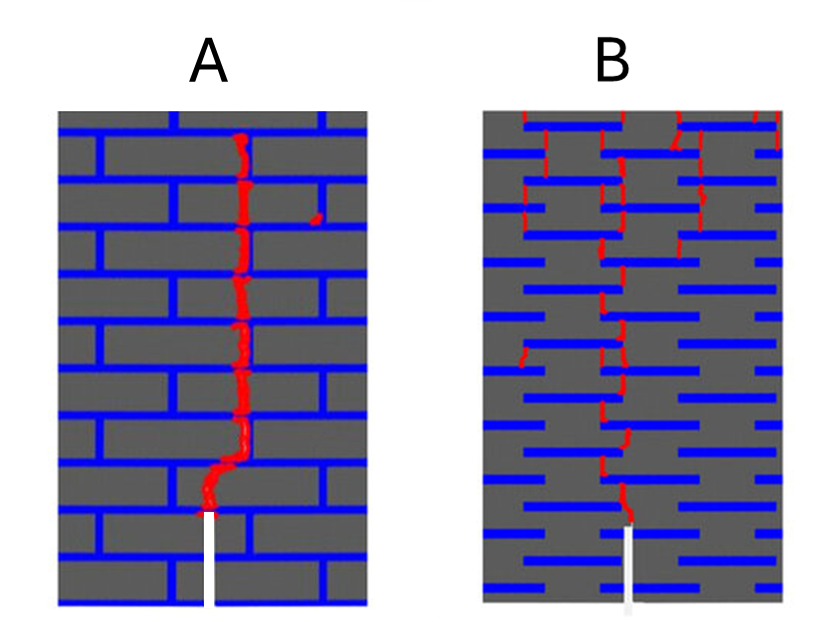}@*)
(*@\bf\hlred{User}@*): You are shown two fracture scenarios.

Describe the differences in damage. Note, cracked material is shown in red color. The initial crack is the white notch at the bottom.

(*@\bf \hl{Cephalo}@*):  The image shows two different fracture scenarios in a composite material.

In the first scenario (A), the crack propagates in a straight line, following the path of least resistance. The initial notch at the bottom is highlighted by a white line, and the crack extends vertically upwards, indicating a brittle fracture.

In the second scenario (B), the crack deviates from its initial straight path and follows a more complex, zigzag pattern. This suggests that the material has undergone significant plastic deformation before fracturing. The initial notch is again highlighted by a white line, but the crack path deviates from the straight line, indicating a more ductile fracture.

The results indicate that the material's behavior under stress is influenced by its microstructure and the presence of inclusions or other reinforcements. The straight crack in scenario A suggests a brittle fracture, where the material fails abruptly without significant plastic deformation. In contrast, the zigzag crack in scenario B indicates a ductile fracture, where the material undergoes significant plastic deformation before failing. This behavior is often observed in materials with a dual-phase microstructure, where the presence of a second phase (inclusions or reinforcements) can lead to crack deflection and energy dissipation, resulting in a more ductile fracture.

(*@\bf\hlred{User}@*):  Estimate which scenario likely has higher toughness? Why?

(*@\bf \hl{Cephalo}@*): Scenario B, with the zigzag crack pattern, likely has higher toughness compared to scenario A with the straight crack. This is because the zigzag crack path indicates that the material has undergone significant plastic deformation before fracturing. Plastic deformation absorbs energy, making the material more resistant to fracture. In contrast, the straight crack in scenario A suggests a brittle fracture, where the material fails abruptly without significant energy absorption, resulting in lower toughness.

(*@\bf\hlred{User}@*): Which case has more distributed crack path?

(*@\bf \hl{Cephalo}@*):  Scenario B has a more distributed crack path, as indicated by the zigzag pattern. This suggests that the material has undergone significant plastic deformation before fracturing, leading to a more complex crack path.

(*@\bf\hlred{User}@*): What steps could I use to toughen the brittle composite?

(*@\bf \hl{Cephalo}@*):  To toughen the brittle composite, various approaches can be employed, such as:

1. Introducing second-phase particles to create a dual-phase microstructure. These particles can deflect the crack path, leading to energy dissipation and improving the material's toughness.
2. Incorporating a gradient in the microstructure, where the material properties vary with position, can also enhance toughness by guiding the crack path and reducing the overall energy release rate.
3. Enhancing the interface between the phases to prevent premature crack initiation and propagation. This can be achieved through surface treatments or chemical modifications.
4. Controlling the processing conditions to achieve a more favorable microstructure, such as refining the grain size and optimizing the distribution of second-phase particles.

These methods aim to improve the material's ability to absorb energy and resist fracture, making the composite more resistant to cracking and tougher overall.

(*@\bf\hlred{User}@*):  Provide a bio-inspired strategy that uses honeycombs.

(*@\bf \hl{Cephalo}@*): A bio-inspired strategy using honeycombs involves mimicking the natural honeycomb structure found in some animal exoskeletons. This structure consists of a hexagonal lattice with varying cell sizes, providing both strength and flexibility. By replicating this design in composite materials, the honeycomb structure can enhance toughness by distributing stress more evenly and preventing premature crack initiation. This approach combines the benefits of both brittle and ductile fracture mechanisms, resulting in a material with improved overall toughness.

(*@\bf\hlred{User}@*): Summarize key insights in a few bullet points.

(*@\bf \hl{Cephalo}@*): - The initial notch at the bottom of both scenarios indicates the starting point of the crack.
- Scenario A shows a straight crack path, suggesting a brittle fracture.
- Scenario B shows a zigzag crack path, indicating a more ductile fracture.
- The zigzag crack path in scenario B suggests that the material has undergone significant plastic deformation before fracturing.
- The straight crack in scenario A indicates a brittle fracture, where the material fails abruptly without significant energy absorption.
- The zigzag crack path in scenario B results in higher toughness compared to the straight crack in scenario A.
\end{lstlisting}
\end{LLMbox}
\textboxcaption{Chat between a User and  \texttt{Cephalo-Idefics-2-vision-8b-beta}, focused on assessing two fracture scenarios. Image reproduced with permission from~\cite{Sen2011StructuralBlocks}, using a Creative Common license (http://creativecommons.org/licenses/by-nc-nd/3.0/).}
\label{TexBox:TB11}
\end{figure}

\begin{figure}[htbp]
\begin{LLMbox}{Chat between a User and \texttt{Cephalo-Idefics-2-vision-8b-beta}}
\linespread{0.4}\selectfont
\begin{lstlisting} 
(*@\bf\hlred{User}@*): You are a materials scientist and you carefully analyze images, step by step.

Answer step-by-step.
(*@\includegraphics[width=.5\linewidth]{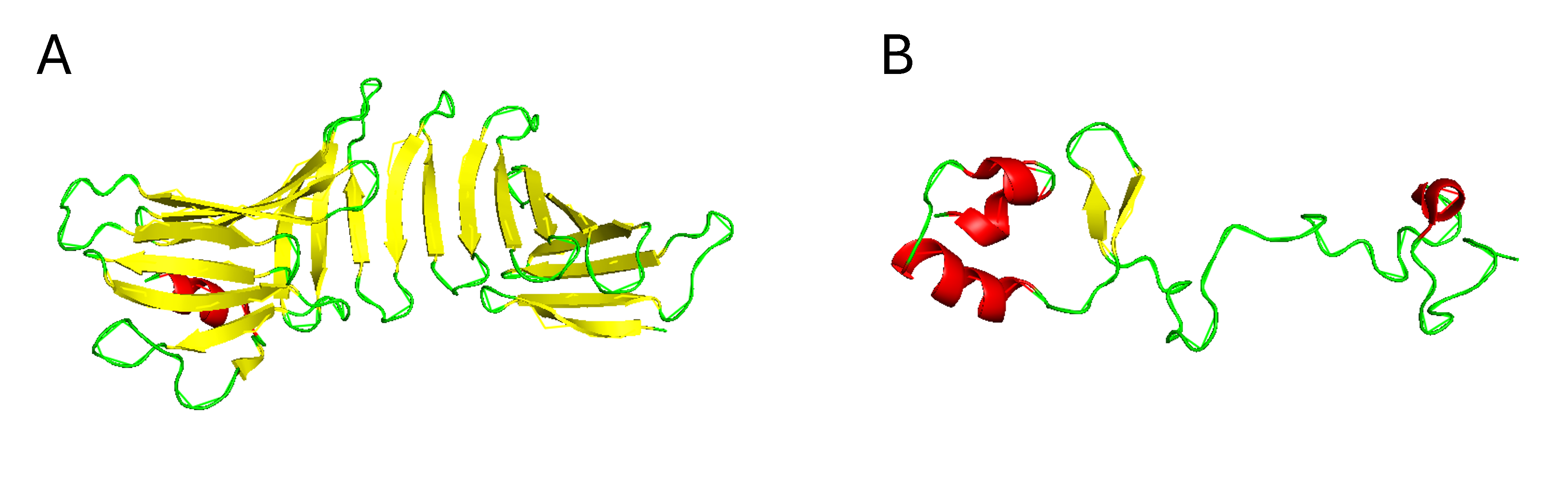}@*)
(*@\bf\hlred{User}@*): You are shown two proteins, colored by their secondary structure.

Looking at the protein images, and analyzing the structure, which of the proteins has a more organized structure?

(*@\bf \hl{Cephalo}@*):  The protein on the left has a more organized structure, as indicated by its detailed and complex arrangement of secondary structures. This suggests that the protein on the left is more stable and less prone to deformation under stress compared to the protein on the right, which appears to have a simpler and less organized structure.

(*@\bf\hlred{User}@*): Based on this analysis, which protein (A or B) would have a higher mechanical resistance?

(*@\bf \hl{Cephalo}@*): Protein A would have a higher mechanical resistance due to its more complex and organized secondary structure, which provides greater stability and resistance to deformation under stress.

(*@\bf\hlred{User}@*): Suggest a few design improvements.

(*@\bf \hl{Cephalo}@*): To enhance the mechanical properties of the protein, several design improvements can be made:

1. **Increase Hydrogen Bonding:** Introducing more hydrogen bonds within the protein structure can improve its stability and resistance to deformation.

2. **Increase Molecular Weight:** By increasing the molecular weight of the protein, its mechanical strength can be enhanced due to the increased number of atoms and bonds, providing greater resistance to deformation.

3. **Modify the Protein Sequence:** Altering the amino acid sequence of the protein can improve its mechanical properties by enhancing the stability of the secondary structures and reducing the likelihood of deformation under stress.

4. **Introduce Cross-Linking:** Adding cross-linking agents can help connect different protein chains, increasing the overall mechanical strength and resistance to deformation.

These design improvements would result in a protein with enhanced mechanical properties, making it more suitable for applications requiring high strength and stability.

(*@\bf\hlred{User}@*): Be creative and think about using the proteins to generate a new electronic device. How would you go about it?

(*@\bf \hl{Cephalo}@*): To apply the proteins in creating an electronic device, we can consider using them as components in a molecular switch or sensor. The protein with a more organized structure (Protein A) could be used to fabricate a molecular switch, which can be controlled by external stimuli such as light or temperature changes. This switch could be integrated into a circuit to control the flow of electricity. Alternatively, the protein with a simpler structure (Protein B) could be used as a molecular sensor, detecting specific chemical or physical changes and transmitting signals to a larger device. This would involve modifying the protein to enhance its responsiveness to the desired stimuli.
\end{lstlisting}
\end{LLMbox}
\textboxcaption{Chat between a User and  \texttt{Cephalo-Idefics-2-vision-8b-beta}, focused on assessing two proteins. The model is tasked to figure out structural features and to assess likely mechanical properties based on the images. The two proteins considered are PDB IDs 5B2A (crystal structure of a mutant of OspA, a beta-sheet rich structure with around 250 residues) and 1WVK (NMR solution structure of the partially disordered protein At2g23090 from Arabidopsis thaliana). }
\label{TexBox:TB12}
\end{figure}

\begin{figure}[htbp]
\begin{LLMbox}{Chat between a User and \texttt{Cephalo-Idefics-2-vision-8b-beta}}
\linespread{0.4}\selectfont
\begin{lstlisting} 
(*@\bf\hlred{User}@*): You are a materials scientist and you carefully analyze images, step by step.

Answer briefly.
(*@\includegraphics[width=.33\linewidth]{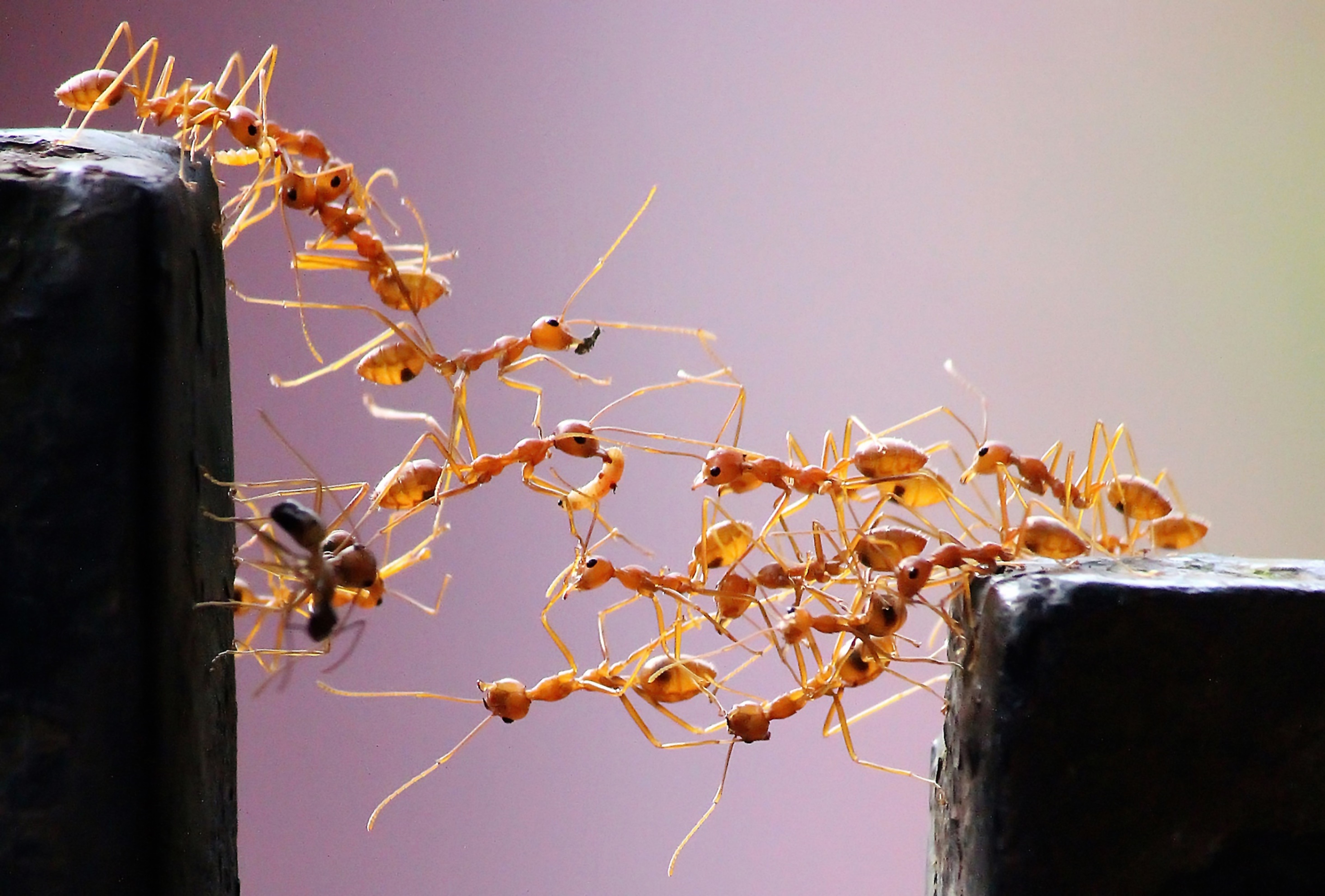}@*)
(*@\bf\hlred{User}@*): Carefully study this photo.

What do you see?

(*@\bf \hl{Cephalo}@*): The image shows a group of ants moving in a line along a surface. The ants are orange in color and appear to be following a specific path. They are positioned on a dark surface, which contrasts with their bright coloration. The ants are oriented in the same direction, indicating coordinated movement. The image captures the ants in a dynamic state, suggesting they are actively navigating or foraging. The background is blurred, focusing attention on the ants and their movement. This image likely illustrates the ants' ability to follow a specific path or trail, which is a common behavior observed in ant colonies.

(*@\bf\hlred{User}@*): How can the phenomenon be used to inform, or explain multi-agent AI systems?

(*@\bf \hl{Cephalo}@*): The image of ants moving in a line can be used to inform or explain multi-agent AI systems by demonstrating how individual agents can coordinate their movements and follow a common path. This behavior is akin to how ants navigate through complex environments, following pheromone trails left by other ants. In multi-agent AI systems, this concept can be applied to develop algorithms that enable multiple agents to coordinate their actions and navigate through complex environments, mimicking the ants' ability to follow a shared path. The image highlights the importance of collective decision-making and coordination in multi-agent systems, which is crucial for achieving efficient and effective navigation in real-world applications.

(*@\bf\hlred{User}@*): Provide a brief outlook to future research one could do.

(*@\bf \hl{Cephalo}@*): Future research could involve studying the mechanisms behind the ants' ability to follow a specific path. This could include investigating the chemical signals they use to communicate and coordinate their movements, as well as understanding how individual ants make decisions based on these signals. Additionally, research could explore the neural mechanisms underlying this behavior, potentially providing insights into how the brain processes information and makes decisions. This knowledge could be applied to develop more efficient and effective multi-agent AI systems that can navigate complex environments and coordinate their actions more effectively.

(*@\bf\hlred{User}@*): What type of deep learning model could be developed based on collective intelligence of ants?

(*@\bf \hl{Cephalo}@*):  A deep learning model based on the collective intelligence of ants could involve developing algorithms that mimic the ants' ability to navigate through complex environments by following pheromone trails. This model could incorporate principles such as reinforcement learning, where agents learn from past experiences and interactions, and distributed decision-making, where multiple agents work together to make collective decisions. The model would aim to replicate the ants' ability to coordinate their movements and navigate efficiently, which could be applied to various real-world applications such as autonomous robotics, search and rescue operations, and environmental monitoring.

\end{lstlisting}
\end{LLMbox}
\textboxcaption{Chat between a User and \texttt{Cephalo-Idefics-2-vision-8b-beta}, analyzing an image of ants forming a bridge.  Image reproduced from: \url{https://www.shutterstock.com/image-photo/unity-strength-1011406435}.}
\label{TexBox:TB13}
\end{figure}

\begin{figure}[htbp]
\begin{LLMbox}{Chat between a User and \texttt{Cephalo-Idefics-2-vision-8b-beta}}
\linespread{0.4}\selectfont
\begin{lstlisting} 
(*@\bf\hlred{User}@*): You are a materials scientist and you carefully analyze images, step by step.

Answer concisely, and accurately.
(*@\includegraphics[width=.5\linewidth]{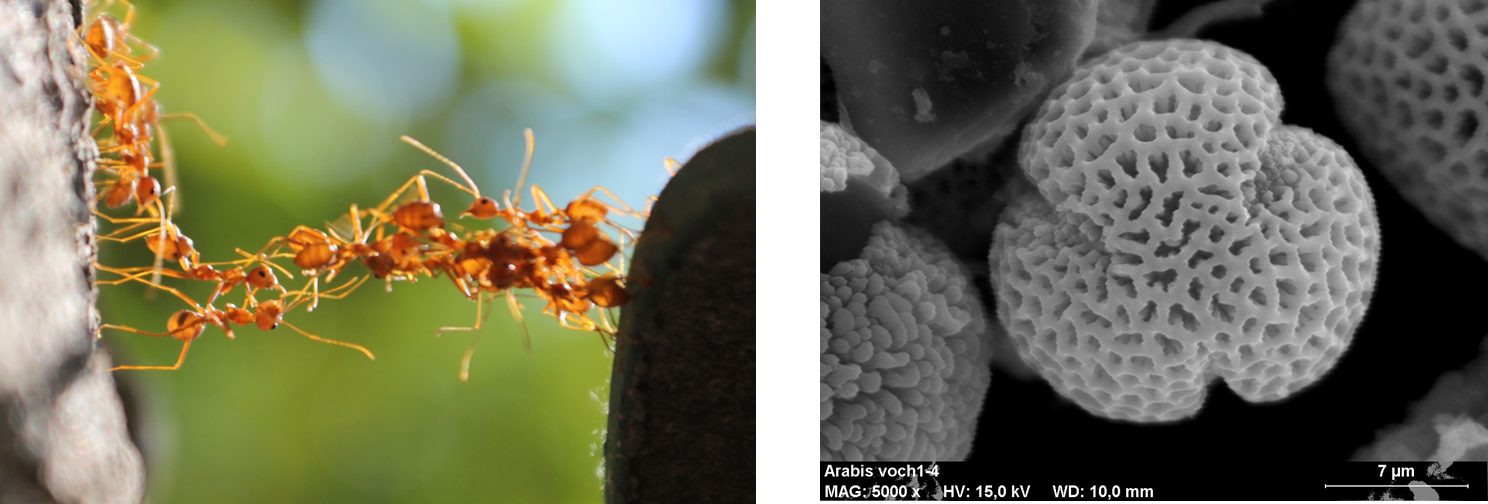}@*)
(*@\bf\hlred{User}@*): You are shown two photos of two very different objects. Carefully study the two photos.

What do you see?

(*@\bf \hl{Cephalo}@*): The image shows two distinct objects:

1. The first object is a group of ants interacting with a tree branch. The ants are depicted in a natural setting, clustered together and moving along the surface of the tree branch. The ants appear to be working together, possibly foraging or building a nest.

2. The second object is a microscopic view of a biological structure, likely a cell or tissue sample. This image is magnified to reveal a detailed, porous structure with a honeycomb-like pattern. The scale bar indicates that the structure is on the microscale, with a size of 7 micrometers ((*@$\mu$@*)m).

These images illustrate the contrast between macroscopic and microscopic scales, highlighting the ants' behavior at the macro level and the detailed cellular structure at the micro level.

(*@\bf\hlred{User}@*): Which of the two images would be better suited to create fracture-resistant materials?

(*@\bf \hl{Cephalo}@*): The second image, showing the detailed honeycomb-like structure, would be better suited for creating fracture-resistant materials. This is because the microscopic structure can provide insights into the mechanical properties and failure mechanisms of materials, which can inform the design of more durable and resistant materials. The macroscopic image of ants interacting with a tree branch does not provide direct information about material properties at the microscale.

(*@\bf\hlred{User}@*): What about the other one, what may it be useful for?

(*@\bf \hl{Cephalo}@*): The first image of ants interacting with a tree branch could be useful for studying social behavior, ecology, or biomechanics. The ants' collective actions and interactions can provide insights into their communication, cooperation, and adaptability in their natural environment.

(*@\bf\hlred{User}@*): How can these concepts be used to inform new materials?

(*@\bf \hl{Cephalo}@*):  The concepts shown in these images can inspire the design of new materials by mimicking natural structures. The honeycomb-like structure observed in the second image can inform the development of materials with enhanced mechanical properties, such as improved fracture resistance and lightweight strength. The ants' collective behavior in the first image can inspire the design of materials that can adapt and respond to environmental stimuli, mimicking the ants' ability to work together and navigate their environment.

(*@\bf\hlred{User}@*): Next, specifically develop a research idea that integrates the concepts seen in the two images.

(*@\bf \hl{Cephalo}@*): One research idea could involve studying the ants' collective behavior and the honeycomb-like structures in detail to gain insights into designing materials that can replicate these properties. By understanding how ants interact and communicate with each other and how their environment influences their behavior, researchers can develop materials that mimic these interactions, potentially enhancing their structural integrity and functionality. Similarly, by studying the honeycomb-like structures at the microscale, researchers can identify the key mechanical properties that contribute to their strength and durability, which can inform the design of new materials with similar characteristics.

(*@\bf\hlred{User}@*): Provide a step-by-step outlook to future research.

(*@\bf \hl{Cephalo}@*):  The next step in future research would involve further analysis of the ants' behavior and the honeycomb-like structures to understand their mechanical properties and adaptive responses. This could be achieved through detailed microscopic imaging, mechanical testing, and computational modeling to gain insights into the material's behavior under different conditions. The results from these studies could inform the development of new materials that mimic these natural structures, potentially leading to advancements in material science and engineering.

(*@\bf\hlred{User}@*):  Now write a detailed prompt for an image generation AI, that allows me to visualize the material. The prompt is:

(*@\bf \hl{Cephalo}@*): Show an image of a material with a honeycomb-like structure, highlighting its porous nature and the interconnectedness of its pores. The image should illustrate how this structure contributes to the material's mechanical properties, such as improved fracture resistance.

\end{lstlisting}
\end{LLMbox}
\textboxcaption{Chat between a User and \texttt{Cephalo-Idefics-2-vision-8b-beta}, analyzing an image of ants forming a bridge and a microscopy image of a pollen particle. The model is tasked to reason over both and develop bio-inspired design concepts. Left image, reproduced from \url{https://en.m.wikipedia.org/wiki/File:AntBridge_Crossing_10.jpg}, based on a CC BY-SA 4.0, Igor Chuxlancev. Right image, reproduced from \url{https://en.wikipedia.org/wiki/Pollen\#/media/File:Arabis_voch1-4.jpg}, CC BY-SA 3.0, Marie Majaura.}
\label{TexBox:TB14}
\end{figure}


\subsection{Model merging to create deeper, mode expressive models: Cephalo-10b/12b model series}
We construct a 10b Cephalo model by combining different parts of multiple models to create a new one~\cite{goddard2024arcee,akiba2024evolutionary}. Model merging has been suggested as a powerful technique in machine learning and deep learning, albeit its use cases in vision-text models has not yet been explored. This approach leverages the strengths of each contributing model, often leading to improved performance and new capabilities that weren't present in the individual models. In our case, we use this strategy primarily to create a larger, deeper model, showcasing how individual components of pre-trained or fine-tuned models can be used create new architectures. 

\begin{figure}
    \centering
    \includegraphics[width=1.\linewidth]{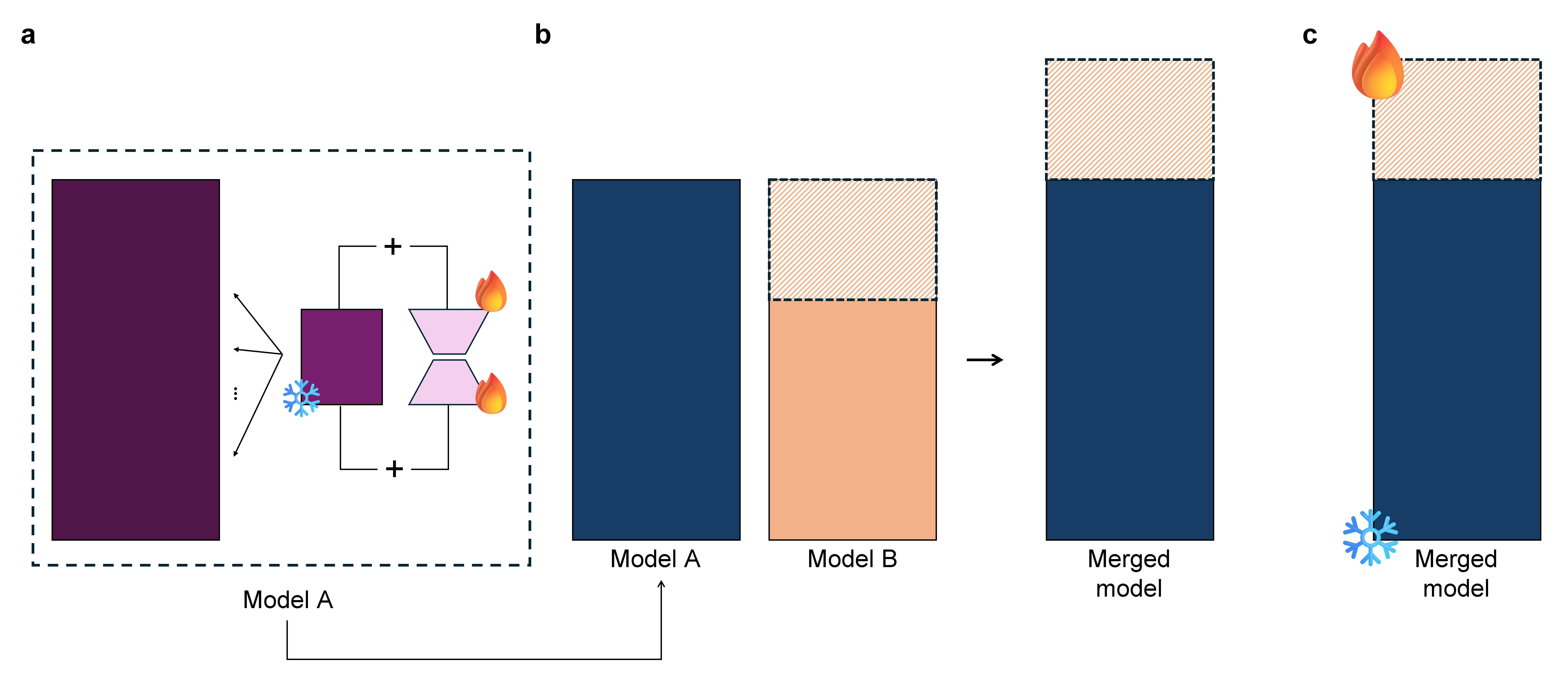}
    \caption{The development process of the merged Cephalo multimodal vision large language model (V-LLM). Panel a shows the process of fine-tuning the first model using low-rank adaptation~\cite{Hu2021LoRA:Models}. Panel b: To merge, we use two models, Model A and Model B, to construct the larger model. Model A is a domain-specific fine-tuned model and Model B a general-purpose chat/instruction-tuned model. We select a set of layers (all layers from Model A, deep layers of Model B). This follows a strategy of using early layers from the domain-specific model and later layers from the general-purpose model. We then merge the selected layers into a new combined model, which is fine-tuned. Panel c: Fine-Tuning of the new model is done by freezing all layers that originate from Model A, and doing a full fine-tune of the layers that originated from Model B. The resulting model enables tasks such as image captioning, visual question answering, and multimodal content generation.}
    \label{fig:Fig_99}
\end{figure}

The hybrid model \texttt{Cephalo-Idefics-2-vision-10b-alpha} demonstrated superior performance by effectively combining domain-specific expertise with general conversational capabilities. We merge the first 32 layers of the text\_model decoder from the \texttt{lamm-mit/Cephalo-Idefics-2-vision-8b-beta} model with the last $N_m$ layers of the chat/instruction-tuned \texttt{HuggingFaceM4/idefics2-8b-chatty} model. We fine-tuning the last $N_m$ layers of the merged model. Details are provided in the Materials and Methods section. For $N_m=8$, this results in a 10b model. Figure~\ref{fig:Fig_99} depicts a schematic of the approach. 
 
In a transformer model, the earlier and later layers play distinct roles in processing input data. The earlier layers are primarily responsible for extracting low-level features and capturing local context. They utilize self-attention mechanisms to focus on immediate, neighboring tokens, identifying basic syntactic patterns, word identities, and short-range dependencies. As the data progresses through these layers, the model produces rich, contextually aware embeddings that encode fundamental aspects of the input.

Conversely, the later layers of a transformer synthesize higher-level abstractions and integrate global context. These layers leverage self-attention to consider tokens across the entire sequence, capturing long-range dependencies and complex relationships. By aggregating information from the entire input, the deeper layers can disambiguate meanings, resolve coreferences, and understand nuanced interactions. This hierarchical processing, moving from detailed token-level understanding to holistic comprehension in deep layers, enables the model to solve sophisticated tasks like contextual analysis of images, interactive human-AI conversations, text generation, summarization, translation and other tasks.

In terms of the modeling strategy used here, using a fine-tuned model for the early layers and a general-purpose model for the later layers leverages the strengths of both models effectively. The fine-tuned model, optimized for a specific domain, excels at extracting domain-specific features and understanding local context. These capabilities are critical in the initial layers, where precise and detailed feature extraction is necessary. In contrast, the general-purpose model, trained on a broader dataset, brings strong generalization capabilities and a comprehensive understanding of diverse contexts, which are essential in the later layers. This combination ensures that the model retains the specialized knowledge of the fine-tuned model while benefiting from the general model's ability to integrate and synthesize information across varied multimodal contexts.

Deeper models tend to perform better generally because they can capture more complex patterns and relationships within the data. Each additional layer allows the model to process and refine the information further, moving from simple, local features to complex, global abstractions. This depth enables transformers to model intricate dependencies and interactions that are essential for high-level cognitive tasks. Moreover, the multi-head self-attention mechanism in each layer allows the model to attend to different parts of the input simultaneously, enhancing its capacity to learn diverse aspects of the data. The progressive refinement through deeper layers, from low-level detail extraction to high-level context integration, allows deep transformer models to achieve superior performance across a wide range of natural language processing tasks.

We show a few brief examples to showcase the performance of the model. Text Box~\ref{TexBox:TB52} shows the results for the protein analysis problem. As can be seen from the responses, they are accurate and well defined, and follow clear logic without being too wordy.  Another example is shown in Text Box~\ref{TexBox:TB54}, for the comparative analysis of ant dynamics and the microscale material. In a second experiment, we show the model images of an uncracked and a cracked egg, and task the model to assess the situation and predict what would happen if the object were to be dropped (see, Text Boxes~\ref{TexBox:TB58} and \ref{TexBox:TB59}.

This approach showcases the potential of merging modles to harness the strengths of multiple models, leading to a versatile and powerful AI system, and to expand a model's depth by integrating more layers.

\begin{figure}[htbp]
\begin{LLMbox}{Chat between a User and \texttt{Cephalo-Idefics-2-vision-10b-alpha}}
\linespread{0.4}\selectfont
\begin{lstlisting} 
(*@\bf\hlred{User}@*): You are a materials scientist and you carefully analyze images, step by step.

Answer step-by-step.
(*@\includegraphics[width=.5\linewidth]{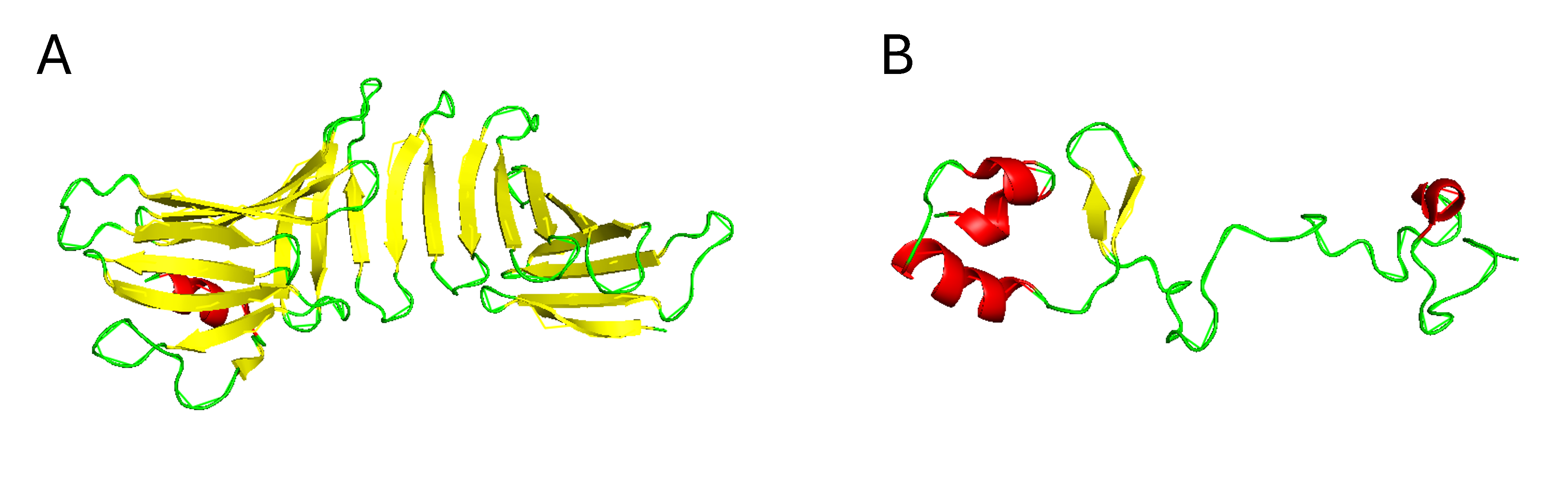}@*)
(*@\bf\hlred{User}@*): You are shown two proteins, colored by their secondary structure.

Looking at the protein images, and analyzing the structure, which protein has a more organized structure?

(*@\bf \hl{Cephalo}@*):  The protein on the left has a more organized structure, as indicated by the presence of alpha-helices and beta-strands. These secondary structures contribute to the protein's stability and function. The protein on the right, in contrast, has a less organized structure with fewer defined secondary structures.

(*@\bf\hlred{User}@*): Which protein (A or B) would have a higher mechanical resistance?

(*@\bf \hl{Cephalo}@*): Protein A would have a higher mechanical resistance due to its more organized structure, which includes alpha-helices and beta-strands. These secondary structures provide stability and strength to the protein, making it more resistant to mechanical forces.

(*@\bf\hlred{User}@*): Suggest a few design improvements.

(*@\bf \hl{Cephalo}@*): To enhance the mechanical properties of the protein, several design improvements can be suggested:

1. Increase the number of alpha-helices and beta-strands in the protein structure. This would improve the overall stability and strength of the protein, making it more resistant to mechanical forces.

2. Introduce disulfide bonds between cysteine residues in the protein. These cross-links would further enhance the structural integrity and mechanical resistance of the protein.

3. Modify the amino acid sequence to optimize the formation of alpha-helices and beta-strands, ensuring a more organized and stable structure.

4. Introduce additional hydrogen bonds between amino acid residues, which would further stabilize the protein's structure and improve its mechanical properties.

By implementing these design improvements, the mechanical resistance of the protein can be significantly enhanced, making it more suitable for applications requiring robust materials.

(*@\bf\hlred{User}@*): Be creative and think about using the proteins to generate a new electronic device. How would you go about it?

(*@\bf \hl{Cephalo}@*): To utilize the proteins for electronic device applications, the focus would be on leveraging their unique properties to create functional materials that can interact with electronic systems. This could involve incorporating these proteins into polymer matrices or composites to enhance specific properties such as conductivity, sensitivity, or mechanical strength. For example, the alpha-helical regions in the protein could be engineered to form conductive pathways, while the beta-strands could contribute to the mechanical stability of the material. This integration would allow for the creation of flexible, biocompatible electronic devices that can interact with biological systems, such as biosensors or implantable medical devices.
\end{lstlisting}
\end{LLMbox}
\textboxcaption{Chat between a User and  \texttt{Cephalo-Idefics-2-vision-10b-alpha}, focused on assessing two proteins. The model is tasked to figure out structural features and to assess likely mechanical properties based on the images. The two proteins considered are PDB IDs 5B2A (crystal structure of a mutant of OspA, a beta-sheet rich structure with around 250 residues) and 1WVK (NMR solution structure of the partially disordered protein At2g23090 from Arabidopsis thaliana). }
\label{TexBox:TB52}
\end{figure}

\begin{figure}[htbp]
\begin{LLMbox}{Chat between a User and \texttt{Cephalo-Idefics-2-vision-10b-alpha}}
\linespread{0.4}\selectfont
\begin{lstlisting} 
(*@\bf\hlred{User}@*): You are a materials scientist and you carefully analyze images, step by step.

Answer concisely, and accurately.
(*@\includegraphics[width=.5\linewidth]{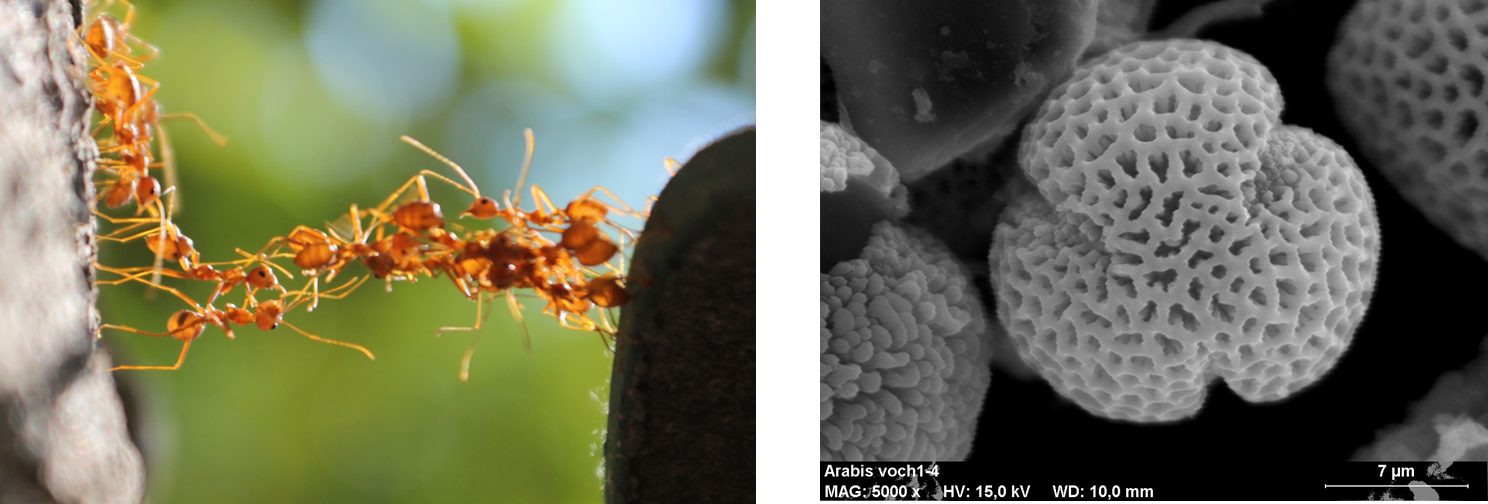}@*)
(*@\bf\hlred{User}@*): You are shown two photos of two very different objects. Carefully study the two photos.

What do you see?

(*@\bf \hl{Cephalo}@*): The image shows two distinct scientific observations:

1. **Ants Climbing a Surface**: The first image depicts ants climbing a surface. The ants, which are orange in color, are seen moving in a line, demonstrating their ability to climb vertical surfaces. This image highlights the ants' specialized structures and behaviors that enable them to defy gravity and move upwards.

2. **Nanoporous Material**: The second image shows a nanoporous material. This material is composed of a lattice structure with numerous small holes or pores. The image is magnified 15,000 times, revealing the intricate details of the material's porous architecture. The scale bar indicates that the image is at a microscopic level, with a width of 7 micrometers ((*@$\mu$@*)m).

The ants' climbing ability and the nanoporous material's structure are both examples of biological and material science, respectively. The ants' climbing behavior is a biological adaptation that allows them to move in environments where gravity would otherwise prevent them from doing so. The nanoporous material, on the other hand, is a synthetic material designed to mimic natural porous structures, which can have applications in fields such as catalysis, filtration, and energy storage.


(*@\bf\hlred{User}@*): Which of the two images would be better suited to create fracture-resistant materials?

(*@\bf \hl{Cephalo}@*): The nanoporous material image would be better suited for creating fracture-resistant materials due to its intricate porous structure. The porous nature of the material can help in distributing stress and preventing cracks from propagating, making it a promising candidate for developing durable materials.

(*@\bf\hlred{User}@*): What about the other one, what may it be useful for?

(*@\bf \hl{Cephalo}@*): The ants climbing image could be useful for studying biomimetic materials or developing new adhesive technologies. The ants' ability to climb vertical surfaces could inspire the design of materials that can adhere to surfaces in challenging environments, such as underwater or in space.

(*@\bf\hlred{User}@*): How can these concepts be used to inform new materials?

(*@\bf \hl{Cephalo}@*): These concepts can inspire the design of new materials by combining biological principles with synthetic materials. For example, the nanoporous structure observed in the image can be replicated in synthetic materials to enhance their mechanical properties, such as strength and durability. Similarly, the ants' climbing behavior can inspire the development of adhesive materials that can stick to surfaces in various conditions. By combining biological insights with synthetic material design, researchers can create innovative solutions to practical problems in fields such as engineering, medicine, and environmental science.

(*@\bf\hlred{User}@*): Next, specifically develop a research idea that integrates the concepts seen in the two images.

(*@\bf \hl{Cephalo}@*): The research idea could involve studying the mechanical properties of the nanoporous material and the ants' climbing behaviors to understand how these structures and biological adaptations contribute to their respective strengths. By examining the nanoporous material's stress distribution and the ants' adhesion mechanisms, researchers could aim to design materials that mimic these properties for enhanced durability and adherence in various applications.

(*@\bf\hlred{User}@*): Provide a step-by-step outlook to future research.

(*@\bf \hl{Cephalo}@*):  1. **Nanoporous Materials**: Investigating the mechanical properties of nanoporous materials, such as their strength, flexibility, and resistance to fracture. This could involve studying how the porous structure influences stress distribution and potential applications in materials science.
2. **Ant Climbing Behavior**: Analyzing the ants' climbing mechanisms to understand how they adhere to surfaces and move against gravity. This could involve studying the anatomy of their feet and the biochemical properties that enable them to climb.
3. **Biomimetic Materials**: Developing new materials that mimic the ants' climbing abilities by incorporating similar adhesive properties into synthetic materials. This could involve creating materials with microscopic structures that allow them to adhere to surfaces in various conditions.

Overall, the research would focus on understanding the mechanical properties of nanoporous materials and the biological adaptations that enable ants to climb, with the goal of developing new materials that leverage these principles for enhanced durability and adherence in practical applications.

\end{lstlisting}
\end{LLMbox}
\textboxcaption{Chat between a User and \texttt{Cephalo-Idefics-2-vision-10b-alpha}, analyzing an image of ants forming a bridge and a microscopy image of a pollen particle. The model is tasked to reason over both and develop bio-inspired design concepts. Left image, reproduced from \url{https://en.m.wikipedia.org/wiki/File:AntBridge_Crossing_10.jpg}, based on a CC BY-SA 4.0, Igor Chuxlancev. Right image, reproduced from \url{https://en.wikipedia.org/wiki/Pollen\#/media/File:Arabis_voch1-4.jpg}, CC BY-SA 3.0, Marie Majaura.}
\label{TexBox:TB54}
\end{figure}

\begin{figure}[htbp]
\begin{LLMbox}{Chat between a User and \texttt{Cephalo-Idefics-2-vision-10b-alpha}}
\linespread{0.4}\selectfont
\begin{lstlisting} 
(*@\bf\hlred{User}@*): You are a materials scientist and you carefully analyze images, step by step.

Answer concisely, and accurately.
(*@\includegraphics[width=.333\linewidth]{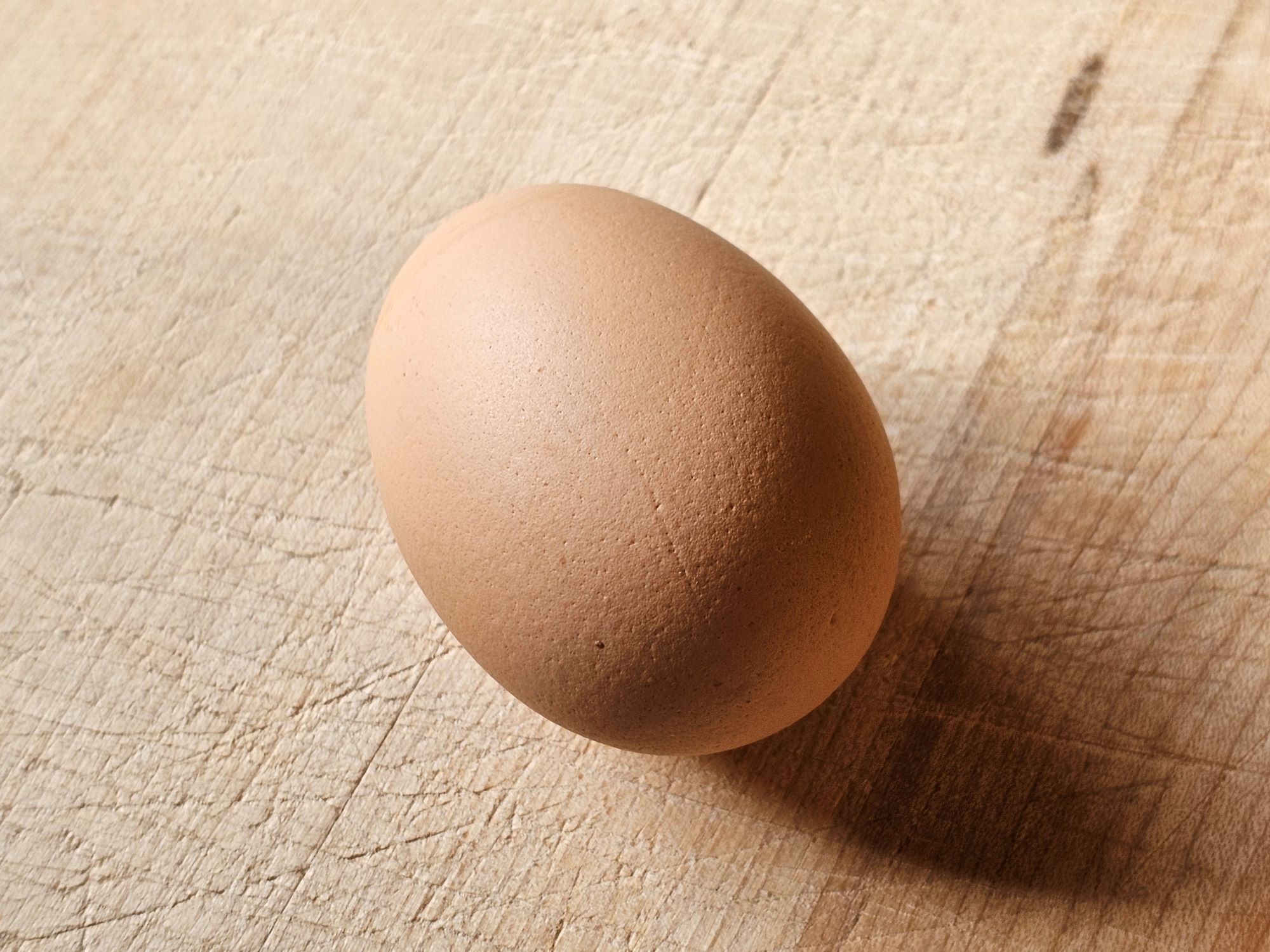}@*)
(*@\bf\hlred{User}@*): Describe what you see. Do you detect a crack?

(*@\bf \hl{Cephalo}@*): The image shows a single, whole, uncracked egg resting on a wooden surface. The egg's surface exhibits a natural, matte texture, typical of an unshelled egg. The background is a textured wood, providing a contrasting background to the smooth surface of the egg. There are no visible signs of cracking or damage on the egg's surface.

(*@\bf\hlred{User}@*): Use the image and propose what would happen if I were to drop the object.

(*@\bf \hl{Cephalo}@*):  If I were to drop the egg, it would likely crack due to the impact. The shell of an egg is composed primarily of calcium carbonate in the form of calcite, which provides structural integrity. However, under stress, such as from a drop, the shell can fracture, leading to the egg's contents being exposed. This is because the shell's structure is not continuous and is susceptible to damage when subjected to external forces. 
\end{lstlisting}
\end{LLMbox}
\textboxcaption{Chat between a User and \texttt{Cephalo-Idefics-2-vision-10b-alpha}, analyzing an image of an egg on a wooden surface. The model correctly states that there is no crack in the egg, and explains what would happen if the egg is dropped. The model provides a detailed description.}
\label{TexBox:TB58}
\end{figure}

\begin{figure}[htbp]
\begin{LLMbox}{Chat between a User and \texttt{Cephalo-Idefics-2-vision-10b-alpha}}
\linespread{0.4}\selectfont
\begin{lstlisting} 
(*@\bf\hlred{User}@*): You are a materials scientist and you carefully analyze images, step by step.

Answer concisely, and accurately.
(*@\includegraphics[width=.333\linewidth]{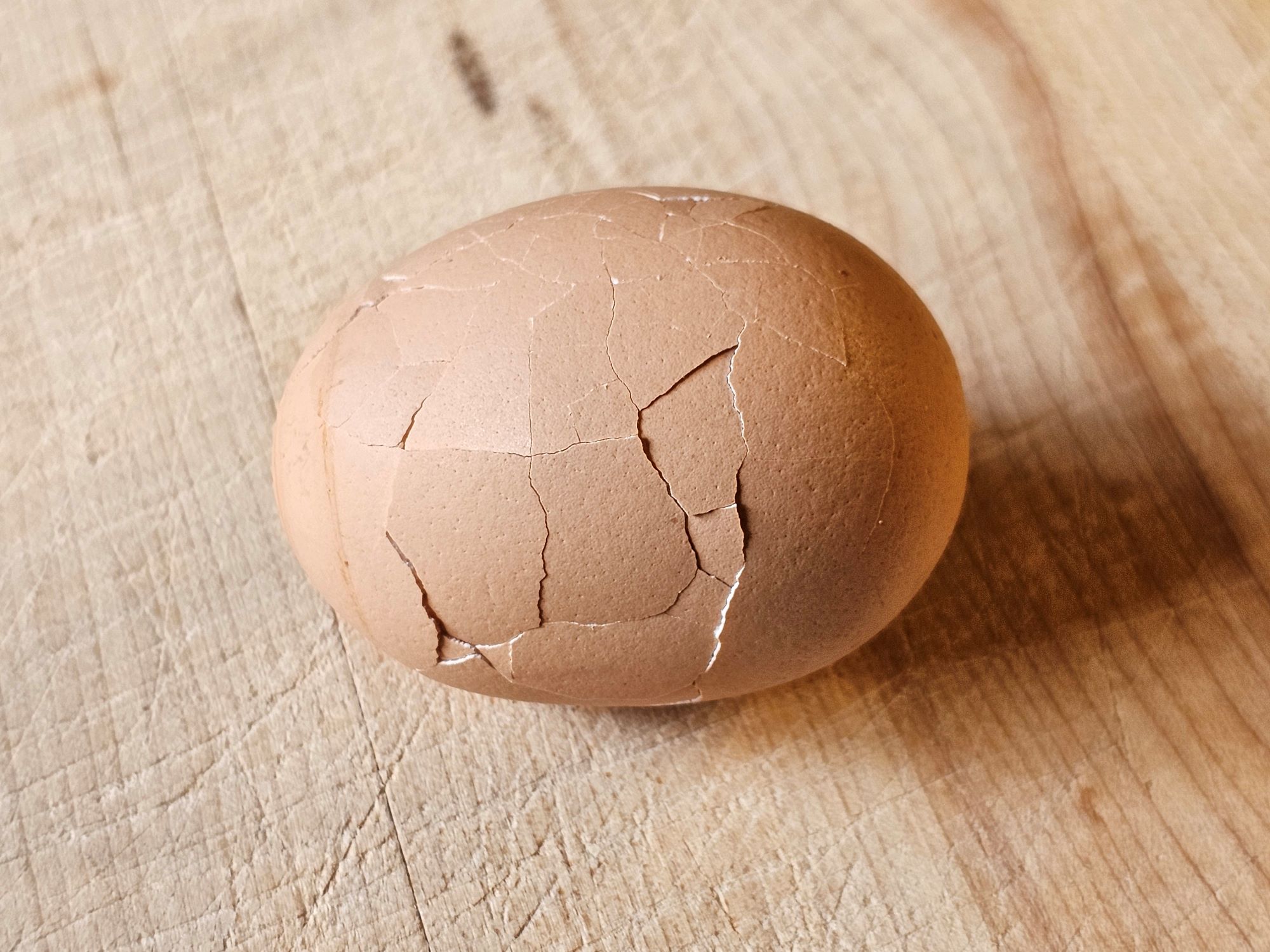}@*)
(*@\bf\hlred{User}@*): Describe what you see. Do you detect a crack?

(*@\bf \hl{Cephalo}@*): The image shows a cracked eggshell on a wooden surface. The eggshell appears to have a rough texture and is partially broken, revealing the inner egg content. The cracks in the shell indicate stress points where the shell has given way, suggesting a point of impact or pressure. The wooden background provides a contrasting texture to highlight the eggshell's surface.

(*@\bf\hlred{User}@*): Use the image and propose what would happen if I were to drop the object.

(*@\bf \hl{Cephalo}@*):  If I were to drop the object, it would likely shatter into pieces due to the impact. The eggshell's brittleness would cause it to fracture along pre-existing cracks or new ones formed during the impact. This would result in the egg content being exposed and potentially damaged or spilled.
\end{lstlisting}
\end{LLMbox}
\textboxcaption{Chat between a User and \texttt{Cephalo-Idefics-2-vision-10b-alpha}, analyzing an image of a pre-cracked egg on a wooden surface. The model correctly identifies the crack in the material and explains what would happen if the egg is dropped. Notable is a detailed description of what is seen in the image.  }
\label{TexBox:TB59}
\end{figure}

For a 12b merged model, we  depict a few  examples as well, in Supplementary Information. Generally, the 12b models do not perform as well as the 10b models, likely due to the fact that earlier layers of the second model are utilized more heavily in the process. For illustration, Text Box~\ref{TexBox:TB62} shows the results for the protein analysis problem.  Text Box~\ref{TexBox:TB64} shows results for the comparative analysis of ant movement and the microscale material. Finally, Text Boxes~\ref{TexBox:TB68} and \ref{TexBox:TB69} depict results of the experiments with images of an uncracked and a cracked egg.

\subsection{Cephalo-4b model series}
This version of Cephalo, \url{lamm-mit/Cephalo-Phi-3-vision-128k-4b-beta}, is based on the Phi-3-Vision-128K-Instruct model~\cite{abdin2024phi3}. The model has a context length of 128,000 tokens. Further details see \url{https://huggingface.co/microsoft/Phi-3-vision-128k-instruct} and \cite{abdin2024phi3}. Unlike the earlier models we use a more complex dataset in training, featuring text-only data (question-answer pairs from a corpus of biomaterials and mechanics of materials) combined with both the Idefics-2 based and GPT-4o based datasets.

Text Boxes \ref{TexBox:TB21} (fracture mechanics), \ref{TexBox:TB22} (protein mechanics), \ref{TexBox:TB23} (bio-inspired AI systems) and \ref{TexBox:TB24} (bio-inspired materials and integration of disparate concepts) summarize results for the experiments conducted with \texttt{Cephalo-Phi-3-vision-128k-4b-beta}.  The responses of this model tend to be longer, more elaborate and feature better reasoning, and overall accurate. This directly reflects the more sophisticated data used to train this model, and shows the remarkable quality and performance of the Phi-series of models generally.


\begin{figure}[htbp]
\begin{LLMbox}{Chat between a User and \texttt{Cephalo-Phi-3-vision-128k-4b-beta}}
\linespread{0.15}\selectfont
\begin{lstlisting} 
(*@\bf\hlred{User}@*): You are a materials scientist and you carefully analyze images, step by step.

You are shown two fracture scenarios.

Describe the differences in damage. Note, cracked material is shown in red color. The initial crack is the white notch at the bottom.
(*@\includegraphics[width=.33\linewidth]{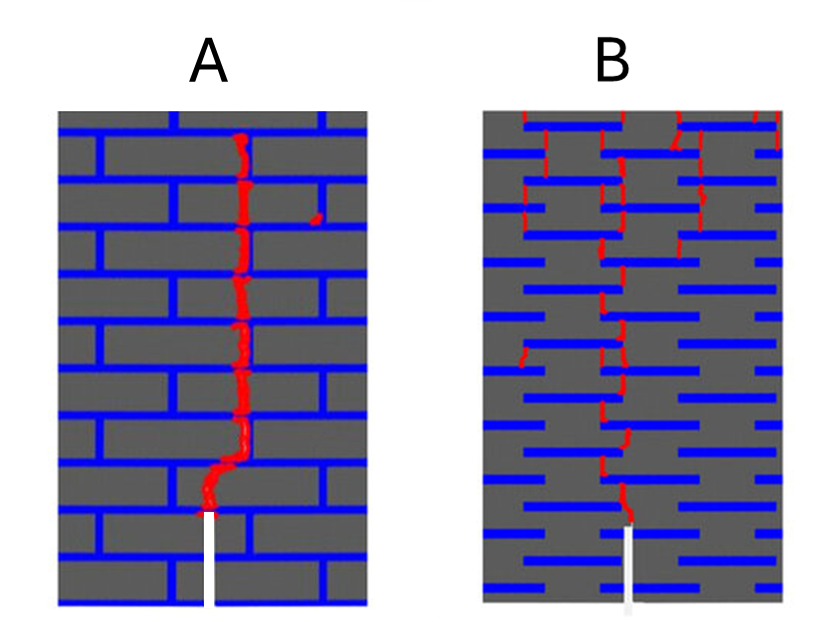}@*)
(*@\bf \hl{Cephalo}@*):  In the image, two scenarios of material fracture are depicted.

In scenario A, the material fractures along a single line, forming a straight crack. This type of fracture is characterized by a clean, linear break in the material. The crack propagates from the bottom to the top of the image, indicating a unidirectional fracture path.

In scenario B, the material fractures along multiple lines, forming a branched crack pattern. This type of fracture is characterized by multiple cracks branching out from a central point. The cracks propagate from the bottom to the top of the image, indicating a multidirectional fracture path.

The differences in damage between the two scenarios are due to the different fracture mechanisms. In scenario A, the material fractures in a single direction, leading to a straight crack. In scenario B, the material fractures in multiple directions, leading to a branched crack pattern. The branched crack pattern in scenario B is more complex and can lead to a larger area of damage compared to the straight crack in scenario A.

(*@\bf\hlred{User}@*):  Estimate which scenario likely has higher toughness? Why?

(*@\bf \hl{Cephalo}@*): Scenario B likely has higher toughness. Toughness is a measure of a material's ability to absorb energy and plastically deform without fracturing. In scenario B, the material fractures along multiple lines, forming a branched crack pattern. This type of fracture is more complex and can absorb more energy before fracturing. In contrast, scenario A involves a single straight crack, which is simpler and can fracture more easily. Therefore, scenario B is likely to have higher toughness due to its ability to absorb more energy and deform plastically before fracturing.


(*@\bf\hlred{User}@*): Which case has more distributed crack path?

(*@\bf \hl{Cephalo}@*):  Scenario B has a more distributed crack path. In scenario B, the material fractures along multiple lines, forming a branched crack pattern. This type of fracture is characterized by multiple cracks branching out from a central point. In contrast, scenario A involves a single straight crack, which is simpler and more linear. Therefore, scenario B has a more distributed crack path due to the multiple cracks branching out from a central point.

(*@\bf\hlred{User}@*): What steps could I use to toughen the brittle composite?

(*@\bf \hl{Cephalo}@*):  To toughen the brittle composite, you could consider the following steps:

1. Introduce a toughening mechanism: One way to toughen the brittle composite is to introduce a toughening mechanism, such as crack deflection or crack bridging. This can help to dissipate energy and reduce the likelihood of fracture.
2. Modify the material properties: Another approach is to modify the material properties of the composite, such as its stiffness or ductility. For example, increasing the stiffness of the composite can help to reduce its tendency to crack under stress.
3. Improve the manufacturing process: The manufacturing process can also play a role in toughening the brittle composite. For example, using a more precise and controlled manufacturing process can help to reduce the likelihood of defects or imperfections that can lead to cracking.
4. Use a different material: If possible, consider using a different material that is more tough and less brittle. This could involve using a different type of composite material or a different type of material altogether.

It's important to note that the specific steps to toughen the brittle composite will depend on the specific application and the desired properties of the composite. Additionally, it's important to consider the trade-offs between toughness and other properties, such as strength and stiffness, when making modifications to the material.

(*@\bf\hlred{User}@*):  Provide a bio-inspired strategy that uses honeycombs.

(*@\bf \hl{Cephalo}@*): A bio-inspired strategy that uses honeycombs to toughen brittle composites is to mimic the structure and properties of honeycombs found in nature. Honeycombs are made up of hexagonal cells that are arranged in a specific pattern, which provides a high degree of toughness and resistance to cracking.

To apply this strategy to brittle composites, you could create a honeycomb-like structure within the composite material. This could involve using a specific type of material or manufacturing process to create the honeycomb structure. The honeycomb structure would help to dissipate energy and reduce the likelihood of cracking, similar to how honeycombs in nature provide toughness and resistance to cracking.

Another approach is to use a honeycomb-like structure as a toughening mechanism within the composite material. For example, you could introduce a toughening mechanism, such as crack deflection or crack bridging, within the honeycomb structure. This would help to dissipate energy and reduce the likelihood of fracture.

It's important to note that the specific approach to using honeycombs to toughen brittle composites will depend on the specific application and the desired properties of the composite. Additionally, it's important to consider the trade-offs between toughness and other properties, such as strength and stiffness, when making modifications to the material.

(*@\bf\hlred{User}@*): Summarize key insights in a few bullet points.

(*@\bf \hl{Cephalo}@*): - Brittle composites can be toughened by introducing a toughening mechanism, such as crack deflection or crack bridging.
- The toughening mechanism can be achieved through a honeycomb-like structure within the composite material.
- The honeycomb structure can be created using a specific type of material or manufacturing process.
- The honeycomb structure can also be used as a toughening mechanism within the composite material.
- The specific approach to using honeycombs to toughen brittle composites will depend on the specific application and the desired properties of the composite.
- It's important to consider the trade-offs between toughness and other properties, such as strength and stiffness, when making modifications to the material.
\end{lstlisting}
\end{LLMbox}
\textboxcaption{Chat between a User and  \texttt{Cephalo-Phi-3-vision-128k-4b-beta}, focused on assessing two fracture scenarios. Image reproduced with permission from~\cite{Sen2011StructuralBlocks}, using a Creative Common license (http://creativecommons.org/licenses/by-nc-nd/3.0/).}
\label{TexBox:TB21}
\end{figure}

\begin{figure}[htbp]
\begin{LLMbox}{Chat between a User and \texttt{Cephalo-Phi-3-vision-128k-4b-beta}}
\linespread{0.3}\selectfont
\begin{lstlisting} 
(*@\bf\hlred{User}@*): You are a materials scientist and you carefully analyze images.

Answer step-by-step. You are shown two proteins, colored by their secondary structure.

Looking at the protein images, and analyzing the structure, which of the proteins has a more organized structure?
(*@\includegraphics[width=.5\linewidth]{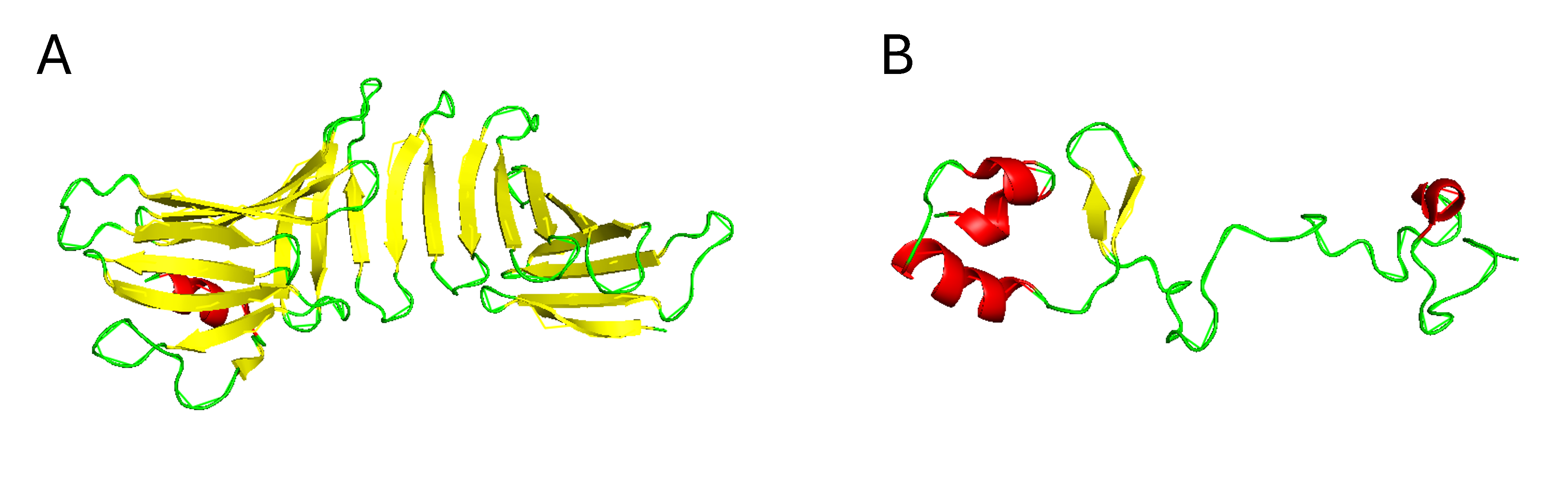}@*)
(*@\bf \hl{Cephalo}@*): Protein A has a more organized structure than Protein B.

In the image, Protein A is shown to have a more regular and symmetrical arrangement of its secondary structure, with a consistent pattern of alpha-helices and beta-sheets. The yellow and green colors indicate the presence of these secondary structures, with the yellow representing alpha-helices and the green representing beta-sheets.

On the other hand, Protein B has a less organized structure, with a more random and irregular arrangement of its secondary structure. The red and green colors indicate the presence of these secondary structures, with the red representing alpha-helices and the green representing beta-sheets.

The more organized structure of Protein A suggests a higher degree of stability and functionality, while the less organized structure of Protein B may indicate a lower degree of stability and functionality.

Therefore, based on the analysis of the secondary structure, Protein A has a more organized structure than Protein B.

(*@\bf\hlred{User}@*): Based on this analysis, which protein (A or B) would have a higher mechanical resistance?

(*@\bf \hl{Cephalo}@*): Based on the analysis of the secondary structure, Protein A would have a higher mechanical resistance than Protein B.

The more organized structure of Protein A suggests a higher degree of stability and functionality, which would result in a higher mechanical resistance. The regular and symmetrical arrangement of the secondary structure in Protein A would provide a more consistent and predictable response to mechanical stress, making it more resistant to deformation and damage.

On the other hand, the less organized structure of Protein B would result in a lower degree of stability and functionality, and a more random and irregular response to mechanical stress. This would make Protein B less resistant to deformation and damage, and more likely to break or lose its functionality under mechanical stress.

Therefore, based on the analysis of the secondary structure, Protein A would have a higher mechanical resistance than Protein B.

(*@\bf\hlred{User}@*): Suggest a few design improvements.

(*@\bf \hl{Cephalo}@*): Based on the analysis of the secondary structure, several design improvements could be suggested to enhance the mechanical resistance of Protein B.

1. Increase the regularity and symmetry of the secondary structure: By increasing the regularity and symmetry of the secondary structure, the mechanical resistance of Protein B could be enhanced. This could be achieved by optimizing the folding and assembly of the protein, or by introducing specific mutations that promote the formation of more stable secondary structures.

2. Enhance the stability of the secondary structure: The stability of the secondary structure is a key factor in determining the mechanical resistance of a protein. By enhancing the stability of the secondary structure, the mechanical resistance of Protein B could be improved. This could be achieved by introducing specific mutations that increase the stability of the secondary structure, or by optimizing the folding and assembly of the protein.

3. Optimize the interaction between secondary structures: The interaction between secondary structures is another important factor in determining the mechanical resistance of a protein. By optimizing the interaction between secondary structures, the mechanical resistance of Protein B could be improved. This could be achieved by introducing specific mutations that enhance the interaction between secondary structures, or by optimizing the folding and assembly of the protein.

4. Increase the size and strength of the protein: The size and strength of the protein are also important factors in determining its mechanical resistance. By increasing the size and strength of Protein B, its mechanical resistance could be improved. This could be achieved by introducing specific mutations that increase the size and strength of the protein, or by optimizing the folding and assembly of the protein.

Overall, these design improvements could enhance the mechanical resistance of Protein B, making it more resistant to deformation and damage under mechanical stress.

(*@\bf\hlred{User}@*): Be creative and think about using the proteins to generate a new electronic device. How would you go about it?

(*@\bf \hl{Cephalo}@*): One potential way to use the proteins to generate a new electronic device is by exploiting their unique properties and structures. Here are some potential approaches:

1. Use the proteins as a substrate for electrochemical reactions: The proteins could be used as a substrate for electrochemical reactions, where their unique structures and properties could be leveraged to control the reactions and enhance their efficiency. For example, the proteins could be modified with specific functional groups or ligands that can facilitate the transfer of electrons between the reagents and the substrate.

2. Use the proteins as a catalyst for chemical reactions: The proteins could be used as a catalyst for chemical reactions, where their unique structures and properties could be leveraged to control the reactions and enhance their efficiency. For example, the proteins could be modified with specific catalytic sites or active sites that can facilitate the transfer of electrons between the reagents and the substrate.

3. Use the proteins as a semiconductor: The proteins could be used as a semiconductor, where their unique structures and properties could be exploited to control the flow of electricity between the electrodes. For example, the proteins could be modified with specific functional groups or ligands that can facilitate the transfer of electrons between the electrodes.

4. Use the proteins as a sensor: The proteins could be used as a sensor, where their unique structures and properties could be exploited to detect the presence of specific molecules or ions. For example, the proteins could be modified with specific binding sites or active sites that can selectively bind to specific molecules or ions, leading to changes in the electrical properties of the material.

Overall, these approaches could enable the use of the proteins as a new electronic device, with a wide range of potential applications in areas such as energy storage, sensing, and catalysis.
\end{lstlisting}
\end{LLMbox}
\textboxcaption{Chat between a User and  \texttt{Cephalo-Phi-3-vision-128k-4b-beta}, focused on assessing two fracture scenarios.}
\label{TexBox:TB22}
\end{figure}

\begin{figure}[htbp]
\begin{LLMbox}{Chat between a User and \texttt{Cephalo-Phi-3-vision-128k-4b-beta}}
\linespread{0.4}\selectfont
\begin{lstlisting} 
(*@\bf\hlred{User}@*): You are a materials scientist and you carefully analyze images, step by step.

Answer briefly. Carefully study this photo.

What do you see?
(*@\includegraphics[width=.33\linewidth]{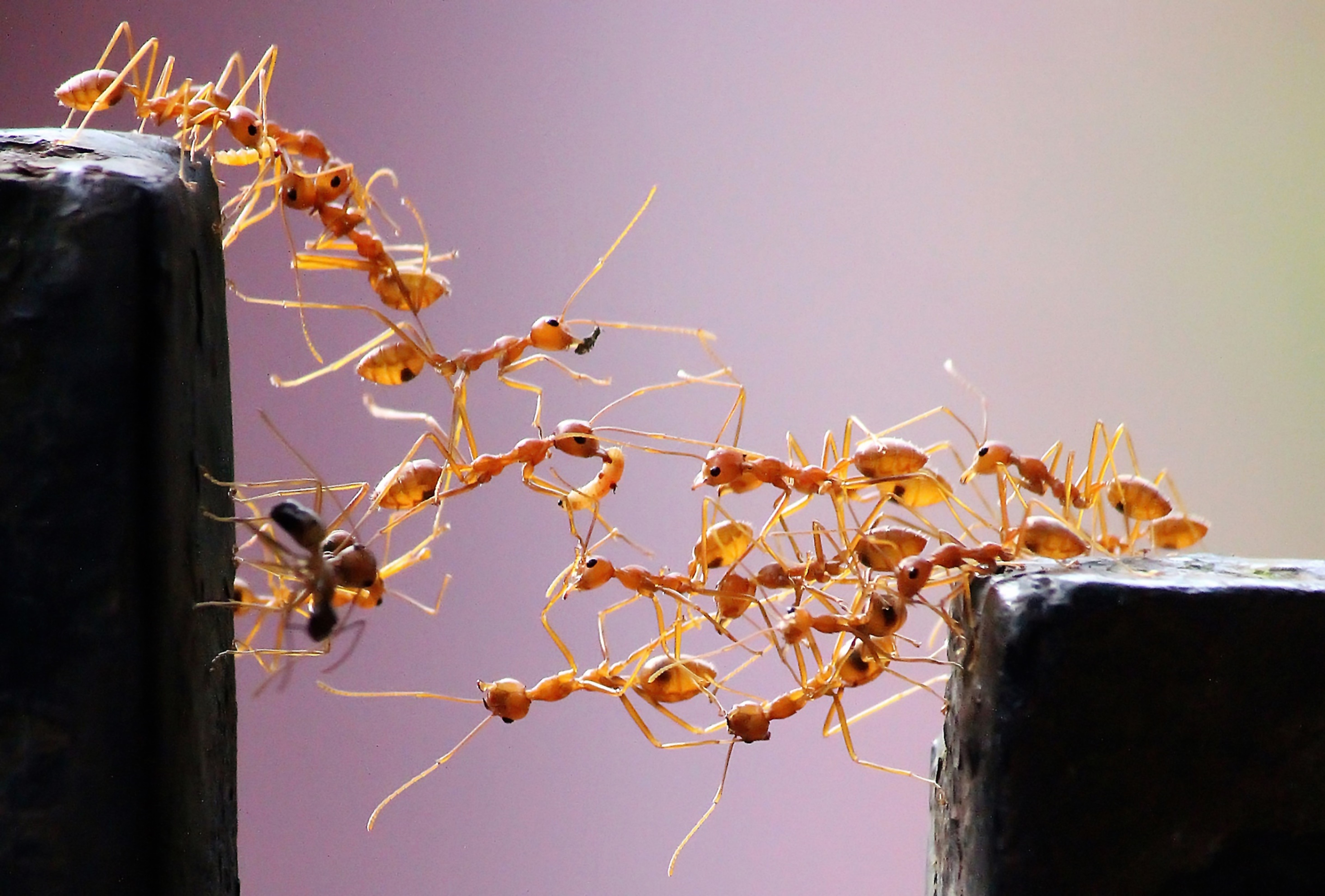}@*)
(*@\bf \hl{Cephalo}@*): The image shows a group of red ants, likely a species of fire ants, in a natural setting. The ants are clustered around a vertical wooden post, with some ants climbing up and others descending. The ants are primarily red in color, with some variations in shade and some individuals displaying a slightly darker hue. The post appears to be made of wood, with visible grain and texture. The background is a gradient of purple to pink, suggesting an outdoor environment. The ants are interacting with the post, possibly foraging or nesting.

(*@\bf\hlred{User}@*): How can the phenomenon be used to inform, or explain multi-agent AI systems?

(*@\bf \hl{Cephalo}@*):The image shows a group of red ants, likely a species of fire ants, in a natural setting. The ants are clustered around a vertical wooden post, with some ants climbing up and others descending. The post appears to be made of wood, with visible grain and texture. The background is a gradient of purple to pink, suggesting an outdoor environment. The ants are interacting with the post, possibly foraging or nesting.

The image can be used to inform or explain multi-agent AI systems in several ways. First, it demonstrates the behavior of individual agents (ants) in a group setting, which is a key aspect of multi-agent systems. The ants are working together to achieve a common goal (foraging or nesting), which is similar to how multi-agent systems can be designed to work together to achieve a common objective.

Second, the image shows how individual agents can adapt and respond to their environment, which is an important aspect of AI systems. The ants are able to climb up and down the post, and adjust their behavior based on the position of the post and the presence of other ants. This adaptability is similar to how AI systems can be designed to respond to changes in their environment and adjust their behavior accordingly.

Third, the image shows how individual agents can interact with each other, which is another important aspect of multi-agent systems. The ants are working together to achieve a common goal, and their interactions with each other can affect the overall behavior of the group. This is similar to how AI systems can be designed to interact with each other and work together to achieve a common objective.

Overall, the image of red ants interacting with a vertical wooden post can be used to inform or explain multi-agent AI systems by demonstrating the behavior of individual agents in a group setting, their adaptability to their environment, and their interactions with each other.

(*@\bf\hlred{User}@*): Provide a brief outlook to future research one could do.

(*@\bf \hl{Cephalo}@*):  The image of red ants interacting with a vertical wooden post can be used to inform or explain future research in several ways.

One potential area of research could be the development of AI systems that are able to adapt and respond to changes in their environment, similar to how the ants in the image are able to adjust their behavior based on the position of the post and the presence of other ants. This could involve the development of algorithms that allow AI systems to learn from their environment and adjust their behavior accordingly.

Another potential area of research could be the development of AI systems that are able to interact with each other and work together to achieve a common objective, similar to how the ants in the image are working together to achieve a common goal. This could involve the development of algorithms that allow AI systems to communicate and coordinate their actions with each other.

Finally, the image of red ants interacting with a vertical wooden post could also inspire research into the design of AI systems that are able to work in natural environments, similar to how the ants in the image are able to work in a natural outdoor setting. This could involve the development of AI systems that are able to navigate and interact with their environment in a way that is similar to how the ants in the image are able to navigate and interact with their environment.

(*@\bf\hlred{User}@*): What type of deep learning model could be developed based on collective intelligence of ants?

(*@\bf \hl{Cephalo}@*):  The image of red ants interacting with a vertical wooden post could inspire the development of deep learning models that are based on the collective intelligence of ants.

One potential approach could be to develop a deep learning model that is able to learn from the behavior of individual ants and adapt its behavior based on the behavior of the group. This could involve the development of algorithms that allow the model to learn from the behavior of the ants in the image and adjust its behavior accordingly.

Another potential approach could be to develop a deep learning model that is able to interact with other models and work together to achieve a common objective. This could involve the development of algorithms that allow the model to communicate and coordinate its actions with other models, similar to how the ants in the image are working together to achieve a common goal.

Finally, the image of red ants interacting with a vertical wooden post could also inspire the development of deep learning models that are able to work in natural environments, similar to how the ants in the image are able to work in a natural outdoor setting. This could involve the development of AI systems that are able to navigate and interact with their environment in a way that is similar to how the ants in the image are able to navigate and interact with their environment.

\end{lstlisting}
\end{LLMbox}
\textboxcaption{Chat between a User and  \texttt{Cephalo-Phi-3-vision-128k-4b-beta}, focused on assessing two proteins. The model is tasked to figure out structural features and to assess likely mechanical properties based on the images.  Image reproduced from: \url{https://www.shutterstock.com/image-photo/unity-strength-1011406435}.}
\label{TexBox:TB23}
\end{figure}

\begin{figure}[htbp]
\begin{LLMbox}{Chat between a User and \texttt{Cephalo-Phi-3-vision-128k-4b-beta}}
\linespread{0.3}\selectfont
\begin{lstlisting} 
(*@\bf\hlred{User}@*): You are a materials scientist and you carefully analyze images, step by step.

Answer concisely, and accurately. You are shown two photos of two very different objects. Carefully study the two photos.

What do you see?
(*@\includegraphics[width=.5\linewidth]{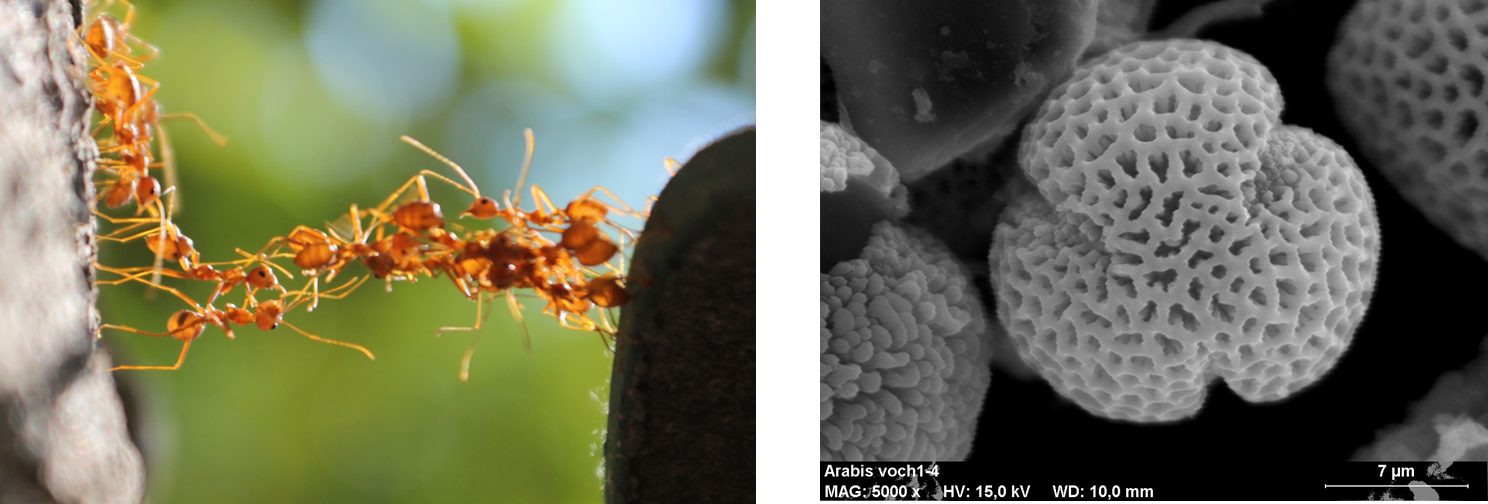}@*)
(*@\bf \hl{Cephalo}@*): The image shows two distinct microscopic views of different materials. On the left, there is a close-up of a group of red ants on a tree branch. The ants are clustered together, with some individuals standing upright and others positioned horizontally. The background is blurred, emphasizing the ants and their immediate surroundings.

On the right, there is a microscopic view of a material with a unique structure. The material appears as a series of interconnected, porous spheres. The spheres are white and have a rough texture, with visible pores throughout. The image is in black and white, highlighting the contrast between the spheres and the background.

The left image is a photograph of ants, while the right image is a microscopic view of a material with a porous structure. The ants are red and are clustered together on a tree branch, while the material has a white, porous appearance.

(*@\bf\hlred{User}@*): Which of the two images would be better suited to create fracture-resistant materials?

(*@\bf \hl{Cephalo}@*): The image on the right, showing a porous material, would be better suited to create fracture-resistant materials. The porous structure of the material can absorb and distribute stress, reducing the likelihood of fracture. The pores can also provide a pathway for cracks to propagate, further enhancing the material's fracture resistance.

(*@\bf\hlred{User}@*): What about the other one, what may it be useful for?

(*@\bf \hl{Cephalo}@*): The image on the left, showing a group of red ants, may be useful for studying the behavior and social structure of ants. The ants are clustered together, indicating a social organization, and their red color may be a warning signal to potential predators. The image can provide insights into the behavior and communication of ants, as well as their role in the ecosystem.

(*@\bf\hlred{User}@*): How can these concepts be used to inform new materials?

(*@\bf \hl{Cephalo}@*):  The concepts shown in the images can be used to inform the development of new materials with specific properties. For example, the porous structure of the material on the right can inspire the creation of materials with enhanced fracture resistance. The social behavior of the ants on the left can inspire the development of materials with specific properties, such as self-healing or adaptive properties.

By studying the properties and behaviors of these materials, scientists can gain insights into how to design and create new materials with specific properties. For example, the porous structure of the material on the right can inspire the creation of materials with enhanced fracture resistance, while the social behavior of the ants on the left can inspire the development of materials with specific properties, such as self-healing or adaptive properties.

Overall, the concepts shown in the images can be used to inform the development of new materials with specific properties, and can provide insights into how to design and create materials with specific properties.

(*@\bf\hlred{User}@*): Next, specifically develop a research idea that integrates the concepts seen in the two images.

(*@\bf \hl{Cephalo}@*): A possible research idea that integrates the concepts seen in the two images is the development of a fracture-resistant material inspired by the porous structure of the material on the right and the social behavior of the ants on the left.

The research idea would involve creating a material with a porous structure that can absorb and distribute stress, reducing the likelihood of fracture. The material would also have specific properties, such as self-healing or adaptive properties, inspired by the social behavior of the ants.

The research would involve studying the properties and behaviors of the ants and the porous material, and using this information to inform the development of the new material. The research would also involve testing the material's fracture resistance and other properties, and refining the material's design based on the results.

Overall, the research idea would involve integrating the concepts shown in the two images to create a new material with specific properties, and would provide insights into how to design and create materials with specific properties.

(*@\bf\hlred{User}@*): Provide a step-by-step outlook to future research.

(*@\bf \hl{Cephalo}@*): A step-by-step outlook to future research on the development of a fracture-resistant material inspired by the porous structure of the material on the right and the social behavior of the ants on the left could include the following steps:

1. Conduct a literature review to identify existing research on porous materials and social behavior in ants.
2. Develop a research plan that outlines the objectives, methods, and timeline for the research.
3. Obtain funding and resources for the research, including equipment, materials, and personnel.
4. Conduct experiments to study the properties and behaviors of the ants and the porous material.
5. Analyze the results of the experiments and use this information to refine the design of the new material.
6. Test the fracture resistance and other properties of the new material.
7. Publish the results of the research in a scientific journal.
8. Use the results of the research to inform the development of new materials with specific properties.

Overall, the step-by-step outlook to future research would involve a systematic approach to developing a new material with specific properties, and would provide insights into how to design and create materials with specific properties.
\end{lstlisting}
\end{LLMbox}
\textboxcaption{Chat between a User and \texttt{Cephalo-Phi-3-vision-128k-4b-beta}, analyzing an image of ants forming a bridge and a microscopy image of a pollen particle. The model is tasked to reason over both and develop bio-inspired design concepts. Left image, reproduced from \url{https://en.m.wikipedia.org/wiki/File:AntBridge_Crossing_10.jpg}, based on a CC BY-SA 4.0, Igor Chuxlancev. Right image, reproduced from \url{https://en.wikipedia.org/wiki/Pollen\#/media/File:Arabis_voch1-4.jpg}, CC BY-SA 3.0, Marie Majaura.}
\label{TexBox:TB24}
\end{figure}


The performance of the model is very good overall, in spite of being of a smaller size than the Idefics-based 8b models. The presentation of the results is detailed, well reasoned, and accurate. For instance, Text Box~\ref{TexBox:TB21} reveals a long but well articulated discussion of the fracture behavior. Text Box~\ref{TexBox:TB21} reveals a long but well articulated discussion of the fracture behavior. Text Box~\ref{TexBox:TB22} shows accurate and deep understanding of protein mechanics, connecting the visual representation of the structure with mechanical properties. The model focuses suggestions of improvement on protein B, which has less structural organization, and is thus the most likely candidate for such engineering tasks. The earlier two models did not identify this proteins as an area of focus. As can be seen in Text Box~\ref{TexBox:TB23} the model also provides a detailed discussion of the ant image, correctly identifying bidirectional movement. It further goes into greater detail about the color of the insects. The remainder of the discussion is, like the others, lucid and well developed with good reasoning. In the final experiment shown in Text Box~\ref{TexBox:TB24}, the model correctly identifies the scenarios depicted in the two images - and, unlike the previous models, appropriately identifies the biological image as a ``material with a porous structure". The step-by-step outlook to future research is detailed and accurate.

\subsection{Image to text to image and 3D modalities}
 
The previous section discussed human-AI interactions to reason over, and explore applications of, image-text modalities. 
Next, we show that by using \texttt{Cephalo-Phi-3-vision-128k-4b-beta} we can develop an integrated pipeline that features image-to-text-to-image modalities. The incorporation of generative image methods can thereby be used render a visualization of the designs developed by the model. The workflow involves two steps:
\begin{enumerate}
    \item Reason over image-text associations, ending with (after one or more turns of interactions) the development of an image generation prompt.
    \item Use of the image generation prompt developed by the \texttt{Cephalo-Phi-3-vision-128k-4b-beta} model to generate visualizations of novel images using a text-to-image or text-to-3D model. 
\end{enumerate}

In a first example we follow a similar human-AI interaction as in the previous section, but finish it off with the development of a prompt. The results are summarized in Text Box~\ref{TexBox:TB36}.  
The conversation involves a user asking the model to analyze two photos (same as before), and to develop design concepts for fracture resistant materials. The images show red ants near a branch and a microscopic porous biological structure. The model correctly identifies key fracture-resistant behaviors in both images. The ants exhibit a bridge-like structure that distributes stress, while the honeycomb pattern allows for efficient stress distribution and energy absorption.

The model proposes creating a new fracture-resistant bio-inspired microstructure by combining these principles. The suggested composite material features a layered design with an outer layer mimicking the ant bridge for stress distribution and an inner layer adopting the honeycomb pattern for energy absorption. It is suggested that this material would be synthesized using a layer-by-layer assembly technique, enhancing fracture resistance.

For the visualization task, the model provides a detailed prompt for an image generation AI to create a realistic depiction of the composite material. The prompt specifies a layered design with a bridge-like outer structure and a honeycomb inner pattern, using realistic colors to reflect the natural appearances of the ants and honeycomb. The resulting image should showcase the material's fracture-resistant properties in high resolution. 

\begin{figure}[htbp]
\begin{LLMbox}{Chat between a User and \texttt{Cephalo-Phi-3-vision-128k-4b-beta}}
\linespread{0.3}\selectfont
\begin{lstlisting} 
(*@\includegraphics[width=.5\linewidth]{Q1000.png}@*)
(*@\bf\hlred{User}@*): You carefully analyze images, step by step.

Answer concisely, and accurately.You are shown two photos of two very different objects. Carefully study the two photos.

Use the concepts in both image and propose a new fracture-resistant bio-inspired microstructure for an advanced material that incorporates the information from both images.

(*@\bf \hl{Cephalo}@*): The image shows two distinct microstructures: one is a group of red ants on a tree branch, and the other is a close-up of a microscopic structure resembling a honeycomb.

The red ants exhibit a natural fracture-resistant behavior. They form a bridge-like structure that distributes stress across multiple points, reducing the likelihood of catastrophic failure. This is akin to the concept of load-sharing in engineering, where stress is distributed to prevent localized failure.

The honeycomb structure, on the other hand, is a natural example of a fracture-resistant material. The hexagonal arrangement of cells allows for efficient stress distribution and energy absorption, similar to the principles of fracture mechanics.

To create a new fracture-resistant bio-inspired microstructure, we can combine the principles observed in both images. The bridge-like structure of the ants can be mimicked to distribute stress, while the honeycomb pattern can be replicated to enhance energy absorption.

A proposed microstructure could be a composite material with a layered design, where the outer layer mimics the ant bridge, and the inner layer adopts the honeycomb pattern. This would provide both stress distribution and energy absorption, making the material more resistant to fracture.

The material could be synthesized using a layer-by-layer assembly technique, where each layer is designed to mimic the properties of the respective structures. The result would be a material with enhanced fracture resistance, inspired by the natural designs observed in the images.

(*@\bf\hlred{User}@*): Now write a detailed prompt for an image generation AI like DALL-E or Stable Diffusion, that allows me to visualize the material.

The prompt must be brief, but be specific so that the image is rendered properly.

The prompt is:

(*@\bf \hl{Cephalo}@*): "Create a detailed image of a composite material inspired by the fracture-resistant structures of red ants and honeycombs. The material should have a layered design with a bridge-like structure on the outer layer and a honeycomb pattern on the inner layer. Use a realistic color palette to represent the natural colors of the ants and the honeycomb structure. Ensure the image is clear and detailed, showcasing the fracture-resistant properties of the material."



The image generated by the AI should showcase the composite material with a clear and detailed representation of the bridge-like structure and honeycomb pattern. The colors should be realistic, reflecting the natural colors of the ants and the honeycomb structure. The material should appear to be fracture-resistant, with the bridge-like structure distributing stress and the honeycomb pattern absorbing energy.


The image should be rendered with high resolution to capture the intricate details of the material's structure. The AI should be able to interpret the prompt accurately and generate a realistic image that showcases the fracture-resistant properties of the material.


The resulting image should be a visual representation of the composite material's fracture-resistant properties, inspired by the natural designs of red ants and honeycombs. The image should be clear and detailed, allowing for a thorough understanding of the material's properties.
(*@\includegraphics[width=.9\linewidth]{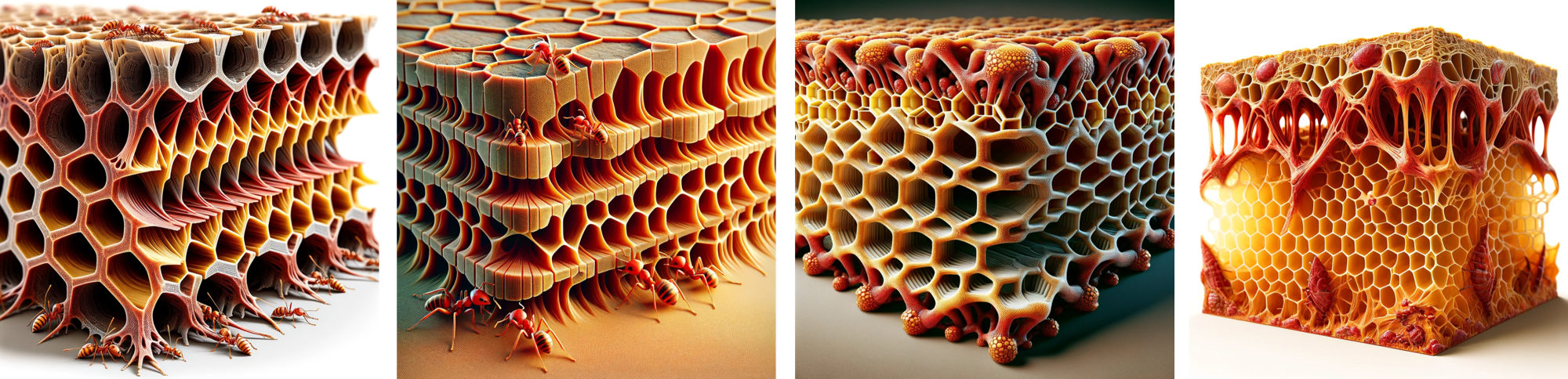}@*)

\end{lstlisting}
\end{LLMbox}
\textboxcaption{Interaction between a User (human) and \texttt{Cephalo-Phi-3-vision-128k-4b-beta} to explore the development of an image generation prompt. The bottom of the text box shows four candidate images generated using \texttt{GPT-4o}. Left/top image, reproduced from \url{https://en.m.wikipedia.org/wiki/File:AntBridge_Crossing_10.jpg}, based on a CC BY-SA 4.0, Igor Chuxlancev. Right/top image, reproduced from \url{https://en.wikipedia.org/wiki/Pollen\#/media/File:Arabis_voch1-4.jpg}, CC BY-SA 3.0, Marie Majaura.}
\label{TexBox:TB36}
\end{figure}

\begin{figure}
    \centering
    \includegraphics[width=1\linewidth]{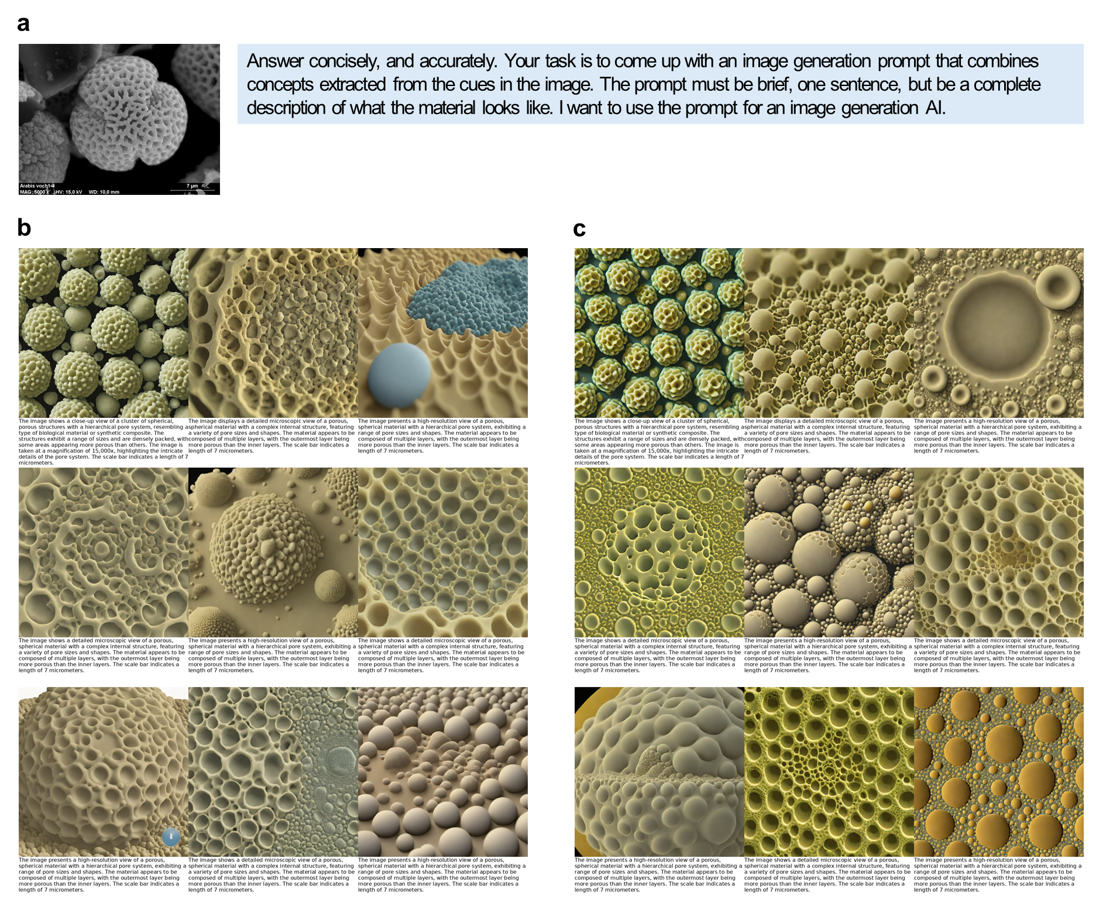}
    \caption{High-throughput image generation pathway. Panel a shows the original image, and panels b and c the results for 9 trials (panel b shows the results after two inference steps with SDXL-Turbo, and panel c the results after four inference steps). The respective image generation prompts so created are included as caption in each sub-panel of the figure. Left/top image, reproduced from \url{https://en.wikipedia.org/wiki/Pollen\#/media/File:Arabis_voch1-4.jpg}, CC BY-SA 3.0, Marie Majaura.}
    \label{fig:Fig_201}
\end{figure}

\begin{figure}
    \centering
    \includegraphics[width=1\linewidth]{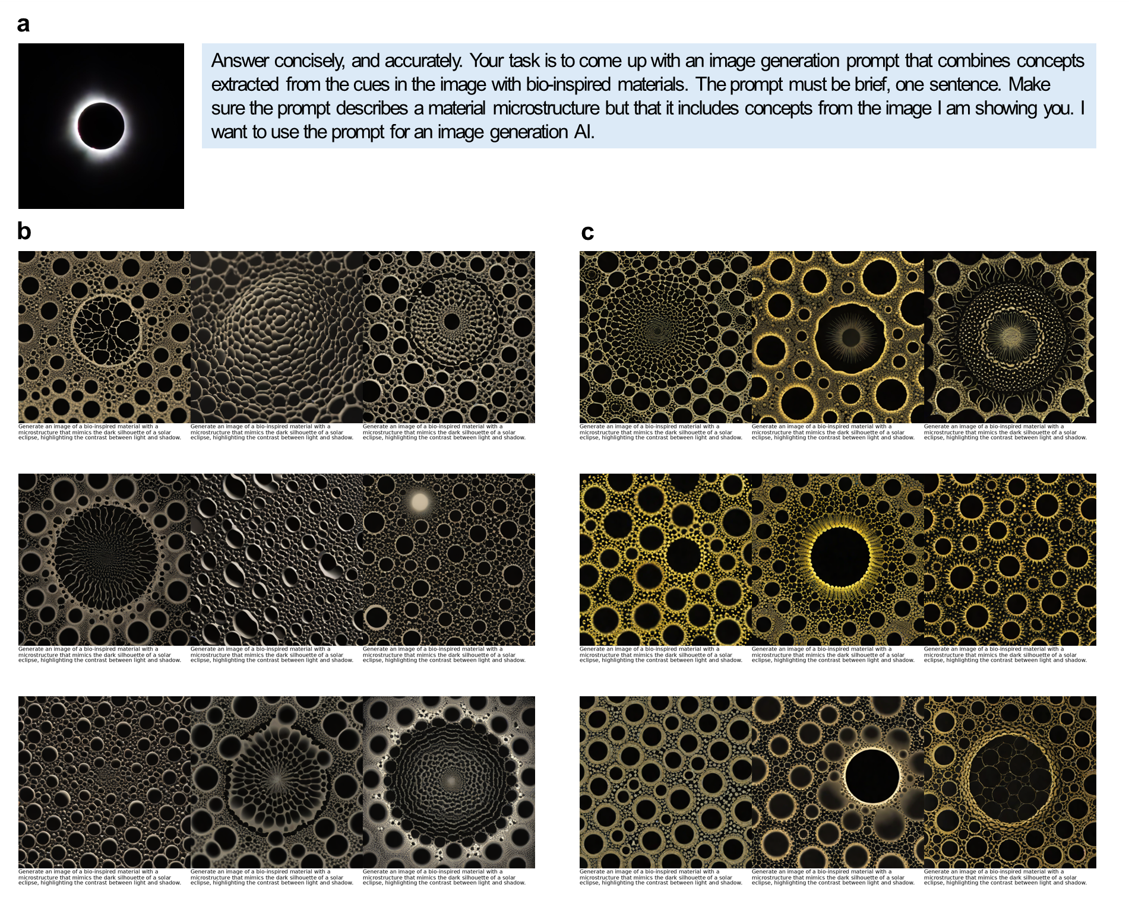}
    \caption{High-throughput image generation pathway. Panel a shows the original image, and panels b and c the results for 9 trials (panel b shows the results after two inference steps with SDXL-Turbo, and panel c the results after four inference steps). The image of the solar eclipse was taken in northern New England on April 7, 2024. } 
    \label{fig:Fig_202}
\end{figure}

In the next experiment we automate this process and construct a pipeline that ingests an image, determines a prompt, and then renders the image using \texttt{Stable Diffusion XL Turbo (SDXL-Turbo)}. We examine two examples, one where we start from the same microscopy image of a pollen particle as used before (Figure~\ref{fig:Fig_201}) and another one where we use an image of the total solar eclipse on April 7, 2024 (Figure~\ref{fig:Fig_202}). In both cases we task the model to reason over the image provided and bio-inspired design principles. As can be seen the vision model can successfully amalgamate, and then synthesize, information provided in the context as well as via the image provided. Notably, this works for both cases where the image is close to the design domain (that is, a micrograph of a pollen particle) or if the image is from a distinct domain (that is, an image of a solar eclipse). 

The prompt used for the first example is: 
\begin{LLMbox}{} 
\begin{lstlisting}
Answer concisely, and accurately. Your task is to come up with an image generation prompt that combines concepts extracted from the cues in the image. The prompt must be brief, one sentence, but be a complete description of what the material looks like. I want to use the prompt for an image generation AI.
\end{lstlisting}
\end{LLMbox}

The prompt used for the second example is: 
\begin{LLMbox}{} 
\begin{lstlisting}
Answer concisely, and accurately. Your task is to come up with an image generation prompt that combines concepts extracted from the cues in the image with bio-inspired materials. The prompt must be brief, one sentence. Make sure the prompt describes a material microstructure but that it includes concepts from the image I am showing you. I want to use the prompt for an image generation AI.
\end{lstlisting}
\end{LLMbox}

\begin{figure}
    \centering
    \includegraphics[width=.8\linewidth]{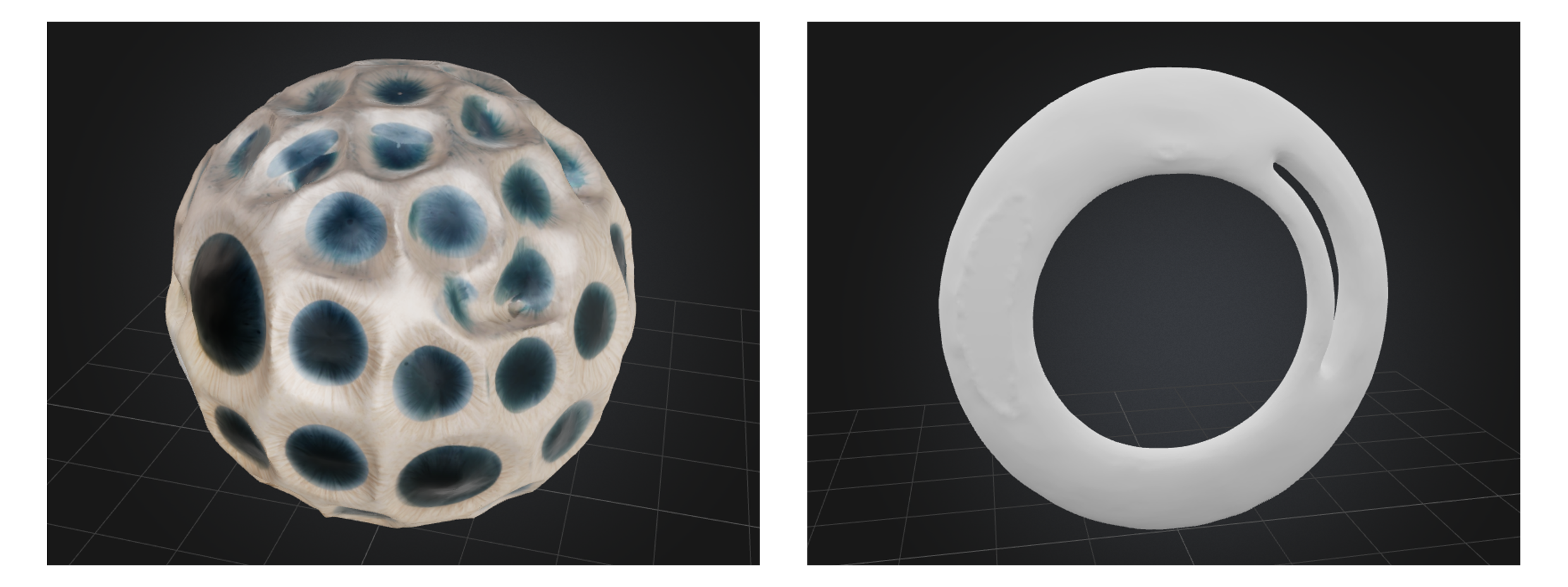}
    \caption{Use of a text-to-3D generative AI model to create a three-dimensional rendering of the prompt \texttt{Generate an image of a bio-inspired material with a microstructure that mimics the dark silhouette of a solar eclipse, highlighting the contrast between light and shadow.} using \url{meshy.ai}. Two variants are shown. } 
    \label{fig:Fig_203}
\end{figure}

A similar approach can be used also to develop 3D models, using generative text-to-3D strategies. Figure~\ref{fig:Fig_203} depicts the results of one experiment, showing the design of 3D models from one of the prompts developed from the earlier examples, specifically combining the solar eclipse image with bio-inspired design. One can see that the design has similar features as the 2D images depicted in Figure~\ref{fig:Fig_202}, showing some level of consistency, as expected. 

\subsection{Mixture of Experts modeling: Constructing Large Models from Smaller Trained Component Models}

\begin{figure}
    \centering
    \includegraphics[width=.9\linewidth]{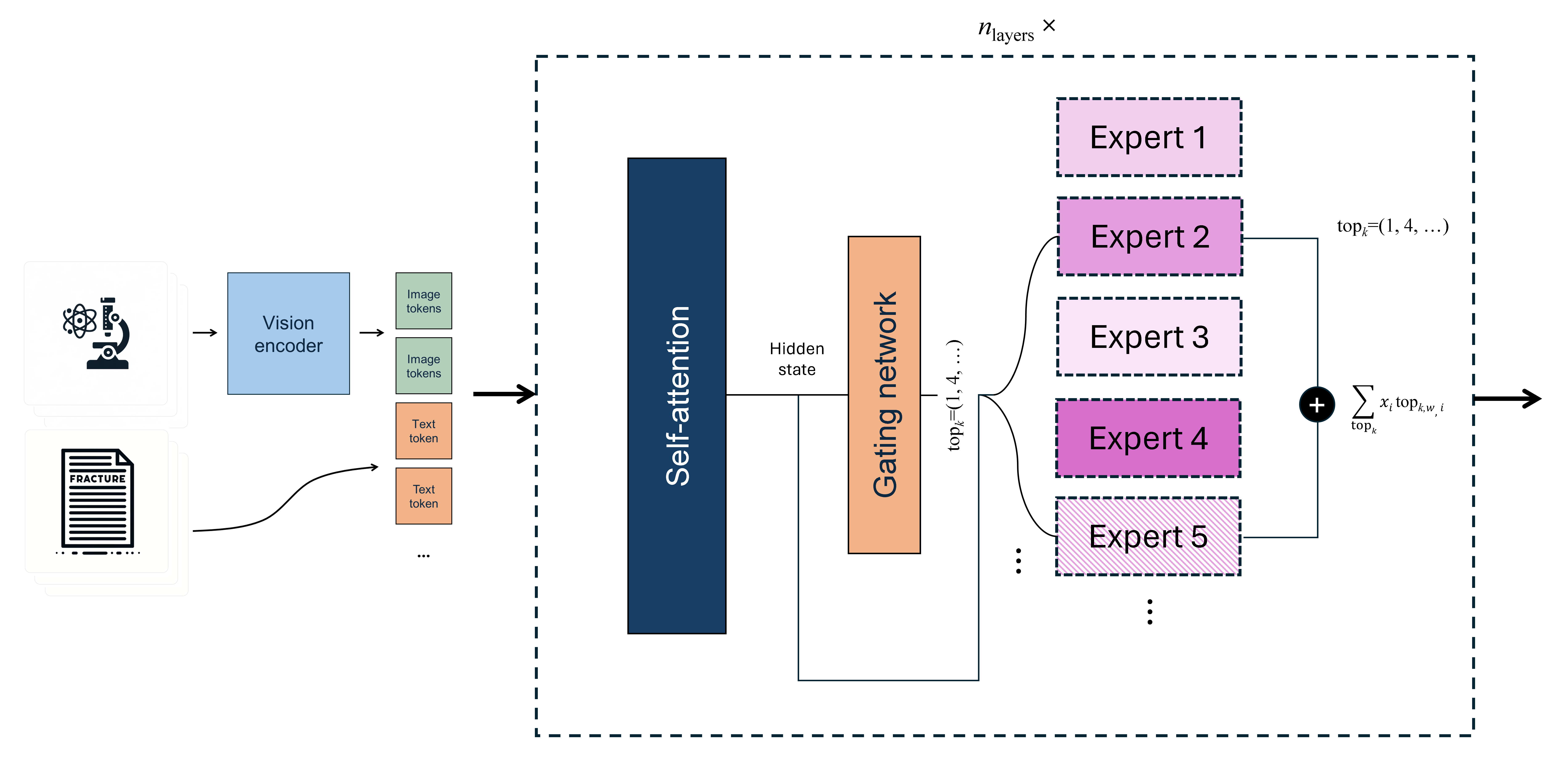}
    \caption{Development of a mixture-of-expert model with sparse gating. The components labeled ``Expert $i$" are extracted from a set of trained models, and hence inherit their capabilities. The input is similarly structured as in the original model, however, the inner mechanics of the model is altered. The schematic shows the overall workflow, where hidden states are routed through a gating network that decides which experts are activated. The results from top $k$ experts are then added up, properly weighted. Normalization layers not shown for clarity. } 
    \label{fig:Fig_2013}
\end{figure}

Model merging, or generally building larger models is a powerful approach. However, the computational complexity and resource demands of these models often limit practical applications in materials science. To address these challenges, we propose an enhanced architecture that integrates a sparse mixture of experts (MoE) model~\cite{eigen2014learning,shazeer2017outrageously,jordan1994hierarchical}, here implemented within the 4b model classes that are derived from the \texttt{Phi-3-Vision} architecture. 
Our approach leverages the strengths of multiple expert networks, dynamically allocating computational resources based on the input data. This not only improves the efficiency of the model but also enhances its performance on complex vision-language tasks. By adopting a sparse MoE mechanism, we can achieve diverse capabilities while maintaining practical feasibility. The Materials and Methods section (as well as the code implementation) present the details of the method, similar in approach as used in other recent AI models such as Mixtral~\cite{Jiang2024MixtralExperts}. Figure~\ref{fig:Fig_2013} shows an overview of the architecture.

This strategy has immense potential in future developments of multimodal vision models. For instance, we can construct a model with three experts derived from the original \texttt{Phi-3-Vision-128k-4b} model, \texttt{Cephalo-Phi-3-vision-128k-4b-beta}, and another fine-tuned model trained to convert images of equations into LaTeX code, \texttt{Cephalo-LaTeX-Phi-3-vision-128k-4b-alpha} (the model was trained on the image-to-LaTeX dataset, \texttt{lamm-mit/OleehyO-latex-formulas}). This particular model with three experts has 4b active parameters at a given time (same as an original Phi-3-V model), but 9b parameters total. The sparse utilization of experts, here limited to $k=1$ expert, offers efficient performance at the cost of a single model but with access to much greater capabilities. 
In the scope of this paper, we present key details of our method, demonstrate its usefulness through a few initial experiments, and discuss its implications for future research. More work is left to future investigations.
Text Boxes~\ref{TexBox:TBMoE_V1} and ~\ref{TexBox:TBMoE_V2} show a few examples of how this model performs. The experiments show that the integrated model successfully integrates the capabilities of the constituting model. For reference, we render the LaTeX code as an equation here:

\begin{align*}2(u+1)(u+2t+3)c_1 \\= 2(u+1)(2+t)b_1+u(u+1)(b_1+c_1)+u(b_1+c_1+ub_3+uc_3)+2u(u+1)c_2\end{align*}   

The gating networks play a crucial role in determining the appropriate experts for each input token. These gating networks are trained using sample prompts that represent a diverse set of inputs. Such initial training can be complemented with further fine-tuning using complex datasets, and it is especially notable that this network can yield novel capabilities through the mixing of top $k$ experts (each weighed with the weighting function obtained via a \texttt{softmax(..)} activation function that produces a probability distribution over the top $k$ experts.

\begin{figure}[htbp]
\begin{LLMbox}{Chat between a User and \texttt{Cephalo-Phi-3-MoE-vision-128k-3x4b-beta}}
\linespread{0.4}\selectfont
\begin{lstlisting} 
(*@\bf\hlred{User}@*): Provide a detailed answer.

What is shown in this image, and what is the relevance for materials design? Include a discussion of multi-agent AI.

(*@\includegraphics[width=.33\linewidth]{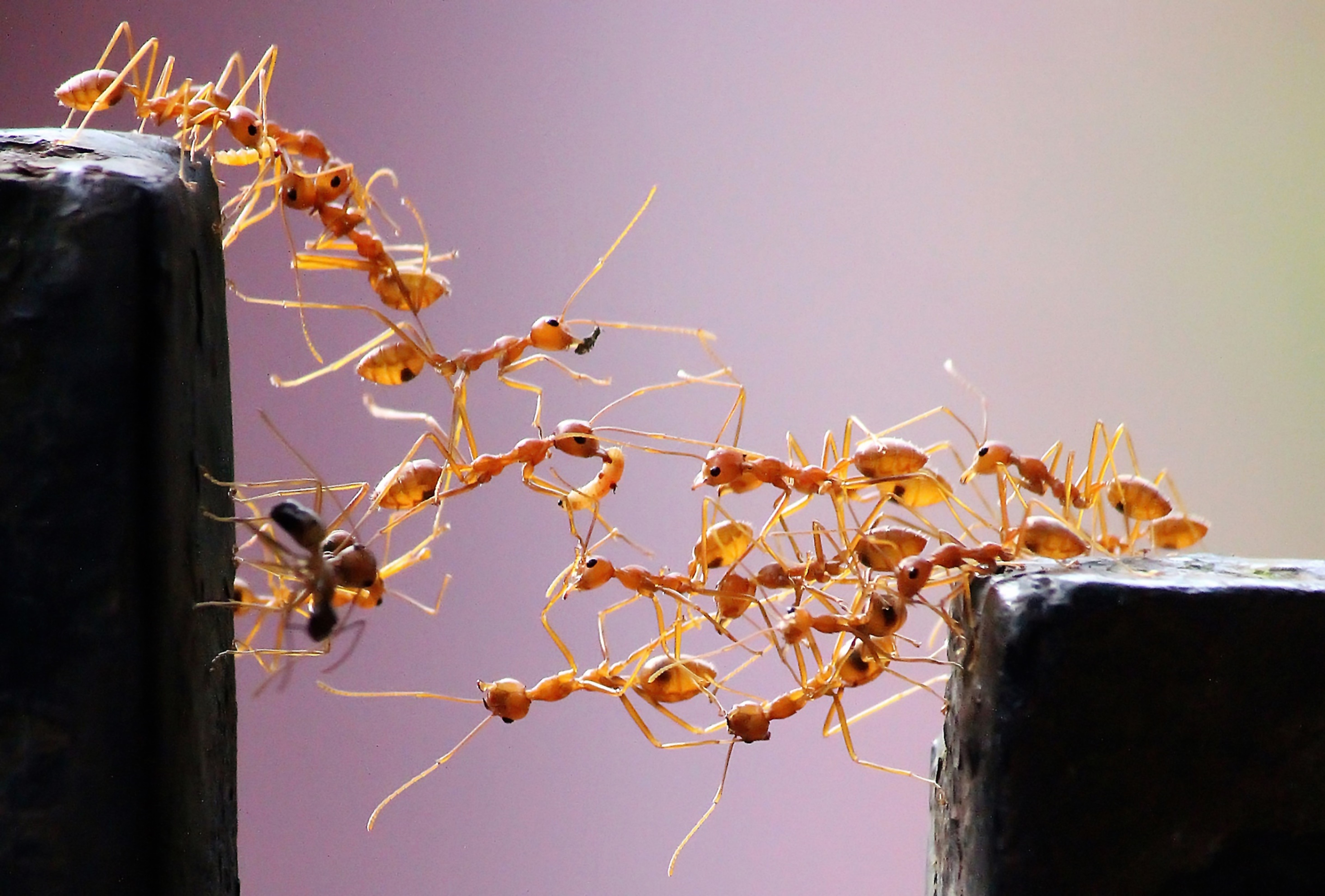}@*)
(*@\bf \hl{Cephalo}@*): The image depicts a group of ants, specifically fire ants, interacting with each other and their environment. The relevance for materials design lies in the study of how these ants build their nests and how their collective behavior can inspire the creation of new materials with unique properties. The ants' ability to work together and adapt to their surroundings can be applied to the development of materials that can self-assemble or change their properties in response to external stimuli.

In terms of multi-agent AI, the image highlights the importance of understanding and modeling the behavior of multiple agents working together towards a common goal. In the case of the ants, their collective behavior allows them to efficiently build their nests and protect their colony. By studying these interactions, researchers can develop algorithms and models that can be used to design and control multi-agent systems in various fields, such as robotics, logistics, and manufacturing.

In conclusion, the image of fire ants interacting with each other and their environment provides valuable insights into materials design and multi-agent AI. By understanding the collective behavior of these ants, researchers can develop new materials with unique properties and design algorithms and models for controlling multi-agent systems in various applications.
\end{lstlisting}
\end{LLMbox}
\textboxcaption{Chat between a User and  \texttt{Cephalo-Phi-3-MoE-vision-128k-3x4b-beta}, tasked to describe an image of ants building a bridge. Image reproduced from: \url{https://www.shutterstock.com/image-photo/unity-strength-1011406435}.}
\label{TexBox:TBMoE_V1}
\end{figure}

\begin{figure}[htbp]
\begin{LLMbox}{Chat between a User and \texttt{Cephalo-Phi-3-MoE-vision-128k-3x4b-beta}}
\linespread{0.4}\selectfont
\begin{lstlisting} 
(*@\bf\hlred{User}@*): Provide the LaTeX formula.

(*@\includegraphics[width=.66\linewidth]{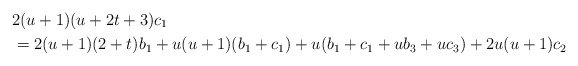}@*)
(*@\bf \hl{Cephalo}@*): \begin{align*}2(u+1)(u+2t+3)c_1 \\= 2(u+1)(2+t)b_1+u(u+1)(b_1+c_1)+u(b_1+c_1+ub_3+uc_3)+2u(u+1)c_2\end{align*}
\end{lstlisting}
\end{LLMbox}
\textboxcaption{Chat between a User and \texttt{Cephalo-Phi-3-MoE-vision-128k-3x4b-beta}, tasked to convert an image of an equation into LaTeX.}
  
\label{TexBox:TBMoE_V2}
\end{figure}

\subsection{Model tuning to predict stress field statistics and crack dynamics}

In this section, we explore the tuning of Cephalo to enhance its predictive capabilities for stress field statistics and crack dynamics in materials. Accurate prediction of these phenomena is crucial for understanding material behavior under stress and for designing materials with improved durability and performance. The Materials and Methods section provides details about the methodologies used to fine-tune Cephalo, leveraging its vision-language model architecture to interpret complex visual data and generate precise, contextually relevant predictions about complex nonlinear materials phenomena. This tuning process not only demonstrates Cephalo's versatility and robustness but also highlights its potential for an extended set of applications in materials science and engineering, particularly in scenarios requiring detailed, quantitative analysis of material properties and their propagation that exceed capabilities of general-purpose models like \texttt{GPT-4o}. Because the model can reason over both quantitative and a range of other vision and text tasks, it shows an integrative ability to interact with human users or other agents.  

\subsubsection{Prediction of stress and atomic energy distribution statistics in materials with defects}

Here we examine Cephalo's ability to predict stress and atomic energy distributions within materials with crack-like defects. Accurate predictions of these distributions are crucial for assessing material stability and identifying failure points. By fine-tuning Cephalo for such tasks, we enhance its capability to analyze visual data and generate precise predictions, highlighting its potential as a valuable tool for materials science and engineering.

\begin{figure}
    \centering
    \includegraphics[width=1\linewidth]{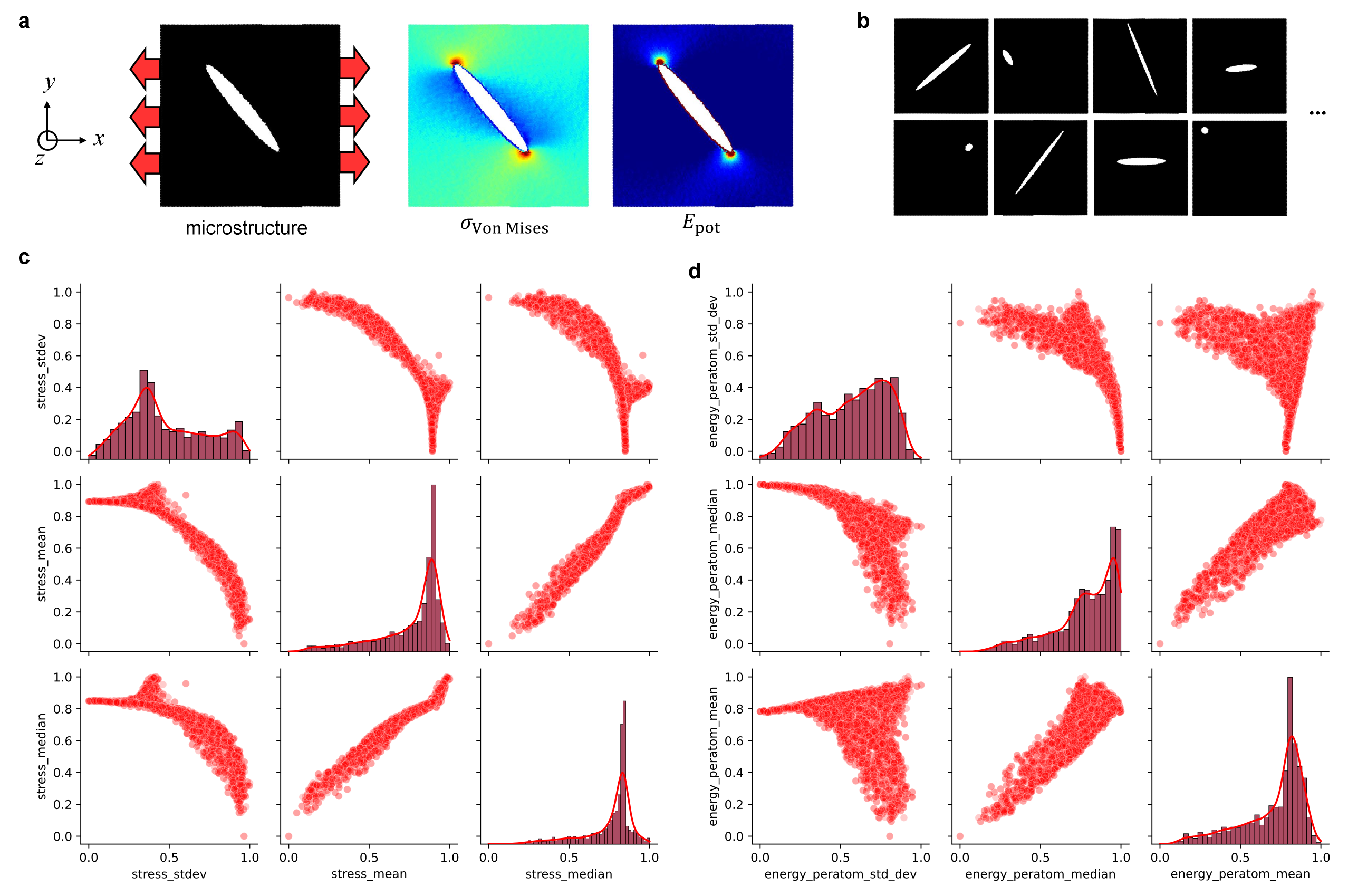}
    \caption{Analysis of stress and atomic energy distributions in a graphene flake with a crack exposed to tensile deformation that form the basis for instruction tuning Cephalo. Panel a: Schematic of a material sample used in molecular dynamics (MD) simulations, subjected to tensile stress in the $x$ direction (data from~\cite{Buehler2022FieldPerceiver:Ologs}). The resulting Von Mises stress~\cite{Buehler2008AtomisticFailure} distribution is shown, illustrating areas of high and low stress within the sample. The image on the right depicts the atomic-level potential energy field. Red=high stress/energy, blue=low stress/energy. Panel b: Sample input microstructures that make up the dataset. Panel c: Histograms and scatter plots illustrating the relationships between Von Mises stress standard deviation, mean stress, and median stress. Each subplot shows the distribution and correlation of these stress metrics across all samples in the dataset. Panel d: Histograms and scatter plots depicting the relationships between the standard deviation, median, and mean of atomic potential energy per atom. Each subplot visualizes the distribution and correlation of these energy metrics across the dataset.} 
    \label{fig:Fig_2000}
\end{figure}

Figures \ref{fig:Fig_2000} presents an introduction to the analysis of stress and atomic energy distributions in a graphene sample with a crack, utilizing both molecular dynamics (MD) simulations and predictive modeling (details, see Materials and Methods).
Figure~\ref{fig:Fig_2000}a depicts the graphene flake~\cite{Zhang2014FractureGrapheneb,Zhang2012FlawGraphene} subjected to tensile stress in the $x$ direction, showing the resulting Von Mises stress distribution with areas of high and low stress (triggered by the presence of the crack). The right image highlights the corresponding atomic-level potential energy field. Figure~\ref{fig:Fig_2000}b shows a few sample microstructures to give a sense of the training data used. Figure~\ref{fig:Fig_2000}c-d illustrate the statistical relationships between von Mises stress metrics (standard deviation, mean, and median) and atomic potential energy metrics (standard deviation, median, and mean) across the dataset, including histograms of the data distribution and scatter plots. These visualizations reveal the distribution and correlation of stress and energy metrics, providing insights into the material's response to stress. This behavior is learned by the Cephalo model so that is capable of predicting the distributions directly from the microstructure (details on training in Materials and Methods), by responding to these instructions:

{\scriptsize
\begin{verbatim}
CalculateVonMisesStressStatistics <stress_stdev, stress_mean, stress_median>
CalculatePotentialEnergyStatistics <energy_peratom_std_dev, energy_peratom_mean, energy_peratom_median> 
\end{verbatim}
\normalsize}

The model is trained to respond with a vector with three entries to each of the commands, reflecting the particular statistical features requested). 

\begin{figure}
    \centering
    \includegraphics[width=.666\linewidth]{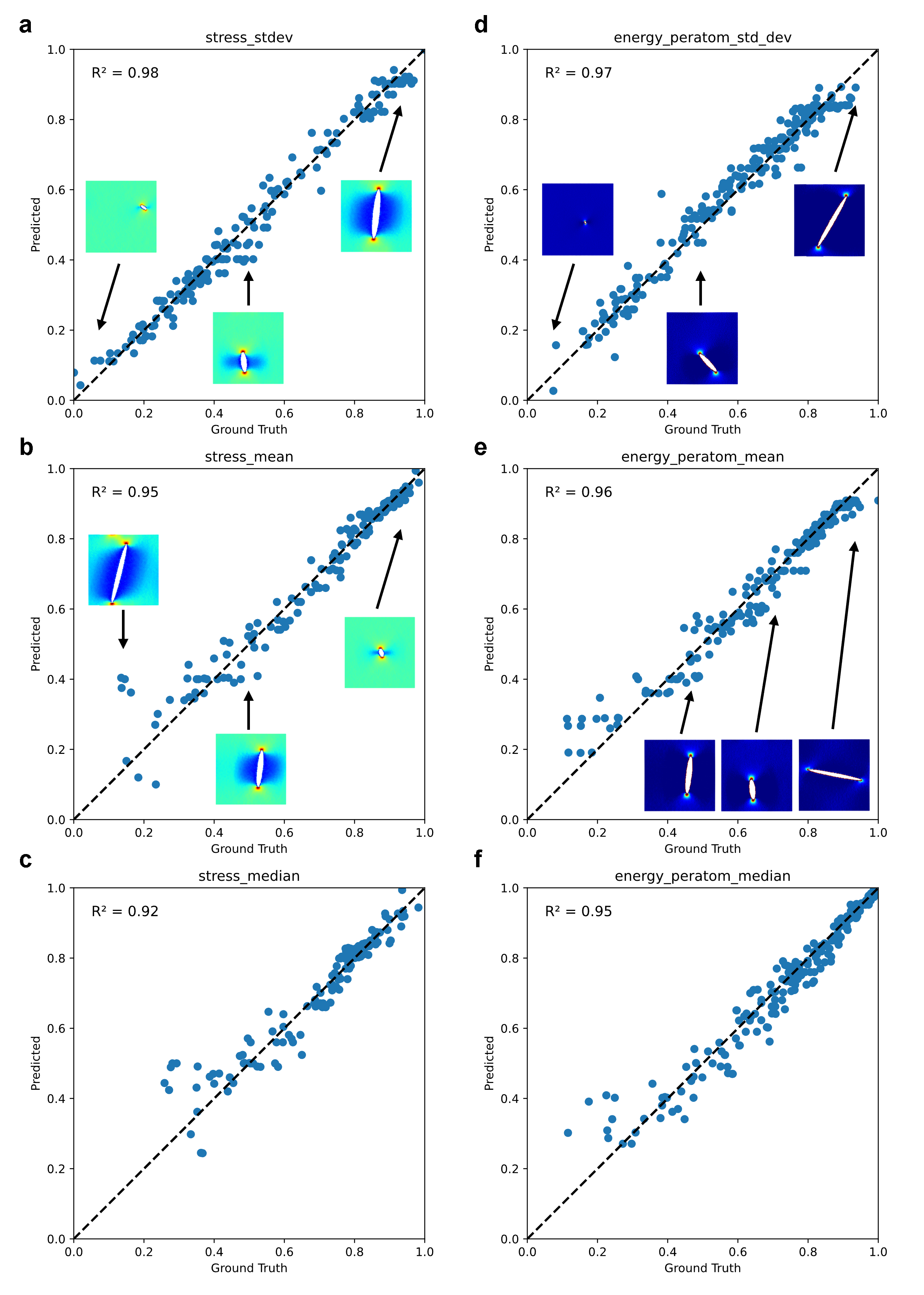}
    \caption{Prediction accuracy for stress and atomic energy distributions, for the test set. Panels a-c: Predicted versus ground truth values for the standard deviation, mean, and median of stress within the material sample. The high $R^2$ values (0.98, 0.95, and 0.92, respectively) demonstrate a high degree of accuracy in predicting atomic stress distributions. Panels d-f: Predicted versus ground truth values for the standard deviation, mean, and median of atomic potential energy per atom. The high $R^2$ values (0.97, 0.96, and 0.95, respectively) indicate strong predictive performance.   The dashed line represents the ideal fit where predicted values perfectly match the ground truth. For the predictions, we share an image of the initial microstructure along with an instruction (\texttt{CalculateVonMisesStressStatistics <stress\_stdev, stress\_mean, stress\_median>} or \texttt{CalculatePotentialEnergyStatistics <energy\_peratom\_std\_dev, energy\_peratom\_mean, energy\_peratom\_median>} and the model then respond with a vector with three entries to each of the commands, reflecting the particular statistical features requested. } 
    \label{fig:Fig_2001}
\end{figure}

Figure~\ref{fig:Fig_2001} evaluates the prediction accuracy of Cephalo for stress and atomic energy distributions. Figure~\ref{fig:Fig_2001}a-c compare the predicted and ground truth values for the standard deviation, mean, and median of atomic potential energy per atom. The high $R^2$ values (0.97, 0.96, and 0.95, respectively) demonstrate Cephalo's strong predictive performance. Similarly, Figure~\ref{fig:Fig_2001}(d-f) compare the predicted and ground truth values for the standard deviation, mean, and median of stress within the material sample, with high $R^2$ values (0.98, 0.95, and 0.92, respectively), indicating a high degree of accuracy in predicting stress distributions. The dashed lines represent the ideal fit where predicted values perfectly match the ground truth. 
The insets in the plots show visual representations of samples that fall into specific ranges of the statistical measures. 
For a more detailed analysis of the fields associated with specific microstructures,  Figure~\ref{fig:Fig_2002} showcases samples of Von Mises stress distributions obtained from MD simulations, categorized by different intervals of standard deviation and mean.  Figure~\ref{fig:Fig_2002}a-c represent low, medium, and high standard deviations, respectively, while Figure~\ref{fig:Fig_2002}d-f represent low, medium, and high means. These visualizations highlight the variations in stress fields and the different crack orientations, sizes, and shapes that cause these variations. Similarly. Figure~\ref{fig:Fig_2002_epot} displays samples of atomic potential energy distributions obtained from MD simulations, also categorized by different intervals of standard deviation and mean. Figure~\ref{fig:Fig_2002_epot}a-c illustrate low, medium, and high standard deviations, respectively, and Figure~\ref{fig:Fig_2002_epot}d-f show results for low, medium, and high means. These samples reveal the differences in potential energy fields and the nature of crack orientations, sizes, and shapes that influence these distributions.

Together, these examples provide a comprehensive overview of the material's stress and energy distributions, the predictive accuracy of Cephalo, and the variability in stress and energy fields due to different crack characteristics.

\subsubsection{Prediction of crack dynamics}

In a similar vein, we further fine-tune the model to predict crack dynamics. The model is trained to predict two features. First, the amount of damage in the final structure, and whether or not the crack will start to propagate (True/False). Similar as in the previous section, the model is trained on a dataset of MD simulation results that investigated crack dynamics~\cite{Buehler2022ModelingModel}. 
First looking at the dataset, Figure~\ref{fig:Fig_2063} displays samples of damage incurred as predicted by MD simulations, categorized by different intervals of standard deviation and mean. Figure~\ref{fig:Fig_2063}(a-c) illustrate low, medium, and high damage. These samples reveal the differences in potential energy fields and the nature of crack orientations, sizes, and shapes that influence damage accumulation.

The model is trained to respond to this instruction with a vector (first item, numerical value between 0 and 1 and second item, True/False):

{\scriptsize
\begin{verbatim}
CalculateCrackDynamics <damage, initiate>
\end{verbatim}
\normalsize}

\begin{figure}
    \centering
    \includegraphics[width=1\linewidth]{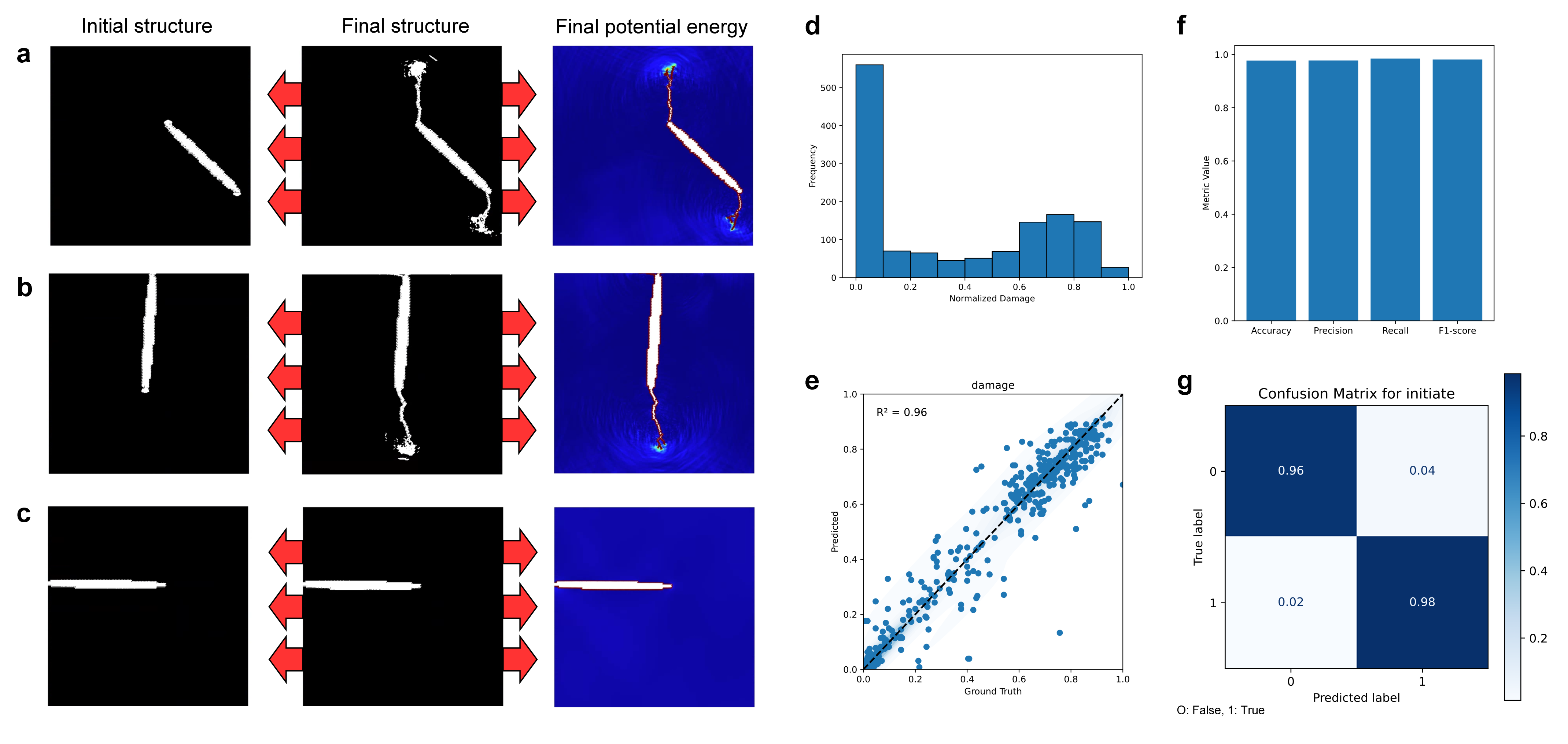}
    \caption{Analysis and prediction of damage and crack initiation. Panels a-c: MD simulation results showing the initial microstructure, final microstructure, and final potential energy distribution of samples with cracks under tensile stress. Panel d: Histogram showing the distribution of normalized damage in final state, across simulation samples. Panel e: Scatter plot comparing predicted damage versus ground truth (GT) damage ($R^2=0.96$), with a dashed line representing the ideal fit. For the predictions made by the model, we share an image of the initial microstructure along with an instruction, \texttt{CalculateCrackDynamics <damage, initiate>}, and the model responds with a two-element vector that includes the amount of damage predicted to have occurred after loading is applied and whether or not the crack will initiate (True/False). Panel f: Bar chart illustrating the accuracy, precision, recall, and F1-score for predicting crack initiation (True/False) (Accuracy: 0.98, Precision: 0.98, Recall: 0.98, F1-score: 0.98). Panel g: Confusion matrix for predicting crack initiation, indicating high accuracy with true positive and true negative rates. } 
    \label{fig:Fig_2061}
\end{figure}

Figure~\ref{fig:Fig_2061} shows results of the analysis and prediction of damage and crack initiation. Figure~\ref{fig:Fig_2061}a-c depicts molecular simulation results showing the initial microstructure, final microstructure, and final potential energy distribution of  samples with cracks under tensile stress, to provide an overview of the type of physics analyzed here. Figure~\ref{fig:Fig_2061}d reveals a histogram showing the distribution of normalized damage in final state, across simulation samples. Figure~\ref{fig:Fig_2061}e shows results obtained from the trained model, providing a scatter plot comparing predicted damage versus ground truth (GT) damage ($R^2=0.96$). Figure~\ref{fig:Fig_2061}f depicts a bar chart illustrating the accuracy (0.98), precision (0.98), recall: 0.98, and F1-score (0.98). Figure~\ref{fig:Fig_2061}g shows, for the same computation, a confusion matrix for predicting crack initiation, indicating high accuracy with true positive and true negative rates.

The results (both field statistics and dynamics) presented here demonstrate Cephalo's potential in advancing materials science through its accurate prediction of stress and atomic energy distributions. By leveraging efficient and accessible modeling, Cephalo not only achieves strong predictive performance comparable to larger models but also enables detailed analysis of material behaviors under various conditions. This capability is crucial for designing more durable materials and understanding failure mechanisms, thereby contributing to innovations in materials engineering and the development of sustainable, high-performance materials.

\section{Conclusions}

Cephalo is a series of lightweight and efficient multimodal vision LLMs (V-LLMs) that can reason over images, fine-tuned for biological and biologically-inspired materials, and capable of providing quantitative analysis of mechanical features of materials with defects. The model allows for a direct  integration of visual and linguistic data for scientific research and practical applications, ranging from analysis to design to microstructure synthesis. Unlike traditional computer vision methods that focus, for instance, om specific classification tasks, the use of V-LLMs offers enhanced flexibility with integrative capability. The method can be used to perform a range of tasks with diverse difficulty, including classification tasks (e.g. does a crack exist) but can also integrate these skills with enhanced synthesis of information, new data, and step-by-step reasoning, and quantitative analysis of material properties (e.g. does a crack propagate). These capabilities can either be used to develop complex insights, optimize microstructures, or to develop new image generation prompts, closing the loop from image analysis to image synthesis and design, as shown in Text Box~\ref{TexBox:TB36}.

By leveraging an innovative dataset generation method that extracts high-quality image-text pairs from complex PDF documents as well Wikipedia scraping, Cephalo provides a framework for interpreting complex visual scenes and generating accurate, contextually relevant language descriptions.  With future expansions of training data, additional evaluations and use cases, we anticipate that the models can be further improved. There are many immediate use cases, such as applications in autonomous experimentation, fine-tuning the models for specific applications (e.g. detection of cracks, damage, or identification of promising design features). The overall best performance is seen for the \texttt{Cephalo-Phi-3-vision-128k-4b-beta} model, effectively integrating data from a more complex dataset to deliver nuanced and accurate outputs. The other models, albeit of larger size, struggle with some details in the images. However, these models have advantages, such as the ability to process high resolution images, and have an exceptional ability to reason over multiple images. 

From a fundamental perspective, the architecture, which combines a vision encoder with an autoregressive transformer, allows for flexible processing of diverse inputs, facilitating applications such as image captioning, visual question answering, and multimodal content generation. The deployment of Cephalo in the study of bioinspired materials, among other fields, underscores its potential to accelerate research, enhance understanding, and drive innovation. The capabilities not only push the boundaries of multimodal interaction and understanding but also offer a powerful tool for researchers to explore new frontiers in materials science and beyond.  

The models ranged in size form 4b to 12b and were constructed via several strategies. The use of model merging offers powerful avenues for future research, where models with different architecture, or distinct capabilities, can be developed. The concept of building more complex models from smaller ones was also be expanded via a mixture of expert modeling approach, here implemented for vision applications, such as in the \texttt{Cephalo-Phi-3-MoE-vision-128k-3x4b-beta} model. To do this we used a feed forward gating functions, but it could also be accomplished using low-rank based approaches (as done in X-LoRA~\cite{Buehler2024X-LoRA:Design}).

We provided several examples to demonstrate additional capabilities in predicting stress and atomic energy distributions, as well as crack dynamics, based on just an image of the material microstructure as input. Figures~\ref{fig:Fig_2000} and \ref{fig:Fig_2001} demonstrate the model's strong predictive performance with high $R^2$ values for the test set, underscoring its accuracy and reliability. The detailed analysis of stress and energy fields in Figure~\ref{fig:Fig_2002} and Figure~\ref{fig:Fig_2002_epot} underscored the model's ability to capture the variability due to different microstructure characteristics, and shows that the model is capable of accurately reasoning over complex physics in fracture mechanics. Additionally, Figure~\ref{fig:Fig_2061} illustrated Cephalo's effectiveness in predicting damage and crack initiation, with high accuracy, precision, recall, and F1-score metrics. These findings showcase Cephalo's versatility and confirm that it can exceed the capabilities of general-purpose models like \texttt{GPT-4o} that cannot make such predictions, making it a valuable tool for materials science applications that can easily be adapted to new tasks using efficient fine-tuning, at parameter counts orders of magnitudes smaller than those in GPT-4 class models. Due to its ability to perform a detailed analysis of material behaviors under various conditions, Cephalo can play an important role in the design of more durable, sustainable, high-performance materials.

Open-source models are vital for scientific research as they enhance transparency and security, allowing researchers to inspect and validate the model’s code and results and easily adapt it for new use cases. This is crucial for their use, for instance, in targeted areas such as bio-inspired materials analysis. This approach further fosters collaboration and innovation, enabling experts in materials science and machine learning to contribute to and refine models, accelerating their development and investigating important technical issues such as model merging or mixture-of-expert strategies. Accessibility is improved, which allows institutions with limited resources to utilize advanced tools, promoting inclusivity in research. Additionally, Cephalo can serve as an educational resource, helping students and researchers learn from its implementation that can be explored on consumer-grade hardware (Table~\ref{tab:educational_goals}). 

\subsection{Summary of key contributions}

Cephalo's potential was illustrated in several key areas:
\begin{itemize}
\item \textbf{Fracture Mechanics:} Cephalo accurately identified differences in fracture scenarios, providing insights into crack propagation and suggesting methods to enhance the toughness of brittle composites. Bio-inspired strategies, such as utilizing honeycomb structures, were proposed, showing capabilities of integrating distinct areas of knowledge to provide actionable outcomes.
\item \textbf{Protein Mechanics:} The model analyzed protein structures, predicting mechanical resistance based on visual cues and suggesting practical design improvements. Additionally, innovative applications, such as using proteins for electronic devices, were explored, showcasing the model's interdisciplinary potential.
\item \textbf{Multi-agent AI Systems:} By analyzing visual inputs of ants, Cephalo informed the development of multi-agent AI systems, emphasizing concepts of coordination and collective behavior. Future research directions, including the study of ant biomechanics to inspire robotics and AI systems, were also proposed.
\item \textbf{Bio-inspired Materials:} The models integrated complex visual and textual data to develop new material designs, combining concepts from disparate sources like ant behavior and honeycomb structures. Detailed research plans and step-by-step approaches were generated, demonstrating the models' capability in aiding material science research.
\item \textbf{Crack Dynamics and Deformation Field Statistics:} The model was capable of quantitatively predicting crack dynamics and damage accumulation in materials, providing valuable insights into stress distribution and crack initiation. Statistical analysis of predicted versus actual damage showed high accuracy, highlighting the model's utility in material stability assessment and failure prediction.
\end{itemize}
A few other examples were explored, such as the detection of cracks in natural materials and structures, such as eggs. The mixture-of-expert model was used to demonstrate diverse capabilities that included the original bio-inspired focus, but also featured capabilities of the original Phi-3 model, and a highly specialized version capable of translating images of equations to LaTeX code. This example illustrated the potential of using MoE strategies to accomplish highly complex, yet general, models. The dataset generation methods and fine-tunring strategies proposed here provides a straightforward way to build a range of custom experts. This ability offers many use cases, especially in multi-agent AI applications. The ability to train the component models individually reduces cost and makes the development flexible, and easily useful, in scientific settings. The ability to integrate a set of small models into larger more capable models provides a scalable method to render more powerful models. 

The Cephalo model was also used to facilitate a seamless pipeline from image-to-text-to-image generation, enhancing the visualization of novel material designs (see, e.g. Figures~\ref{fig:Fig_201} and \ref{fig:Fig_202}). Prompts generated by Cephalo were successfully used with models such as Stable Diffusion XL Turbo and Meshy (Figure~\ref{fig:Fig_203}) to create detailed 2D and 3D renderings of complex bio-inspired materials developed from a set of image cues. This capability underscores Cephalo's potential in visualizing and designing complex material structures, and the ability to reason over disparate concepts and cues over multiple interaction turns. 

Beyond the capabilities of the models to reason over scientific images, Cephalo has been shown to feature enhanced capabilities such as the prediction of stress field statistics in materials with crack defects and the ability to predict crack initiation and damage. These and other aspects clearly demonstrate capabilities beyond current frontier models and a high degree of flexibility for fine-tuning against materials science-specific reasoning tasks. Vision and text capabilities are an emerging area in open-source LLM research, and the ability to custom train a vision-LLM in a computationally efficient and sustainable manner is critical for open science and high-quality models customizable for materials science and engineering applications. Unlike closed-source models like \texttt{GPT-4o}, Cephalo supports special instruction tuning for predicting stress field statistics and complex dynamical phenomena that can be extended towards other material properties and tasks like damage detection, or being able to predict design improvements. Integrating this approach within a MoE framework provides a powerful approach for successively adding new capabilities to meet various demands, including on-the-fly training. The inclusion of training scripts (see Materials and Methods) offers a straightforward avenue for further fine-tuning and customization, ensuring that the framework can leverage future high-quality open-source models. Another advantage is that smaller models can run on consumer hardware, making them accessible for broader use, including in autonomous labs, or can be used for confidential data such as biomedical patient-specific information and images that require encapsulated, secure, local systems. These enhancements highlight Cephalo's contributions to the field and its potential for advancing materials science research.

\subsection{Outlook and future research}
The combination of visual and linguistic data processing renders the model as a general-purpose tool for interdisciplinary research. Building on the examples provided in the current paper, further improvements can be made by expanding the training data and evaluating the models in additional use cases. One specific avenue that could be fruitful is to use one of the Cephalo models for further refinement of the training set, to add a capability to solicit new data autonomously (e.g. via agentic modeling), or to expand the dataset further. The improved understanding of scientific concepts will likely aid in developing more accurate, more nuanced and better reasoned training data. 

Other avenues can include the incorporation of video, which is especially important in understanding time-dependent phenomena or the causality of complex physics. To achieve this, the model can be trained to understand stacks of images representing video frames, which will require additional fine-tuning and computational resources due to the increased number of tokens; albeit, the 8b Idefics based models are already capable of reasoning over a series of images and may be a good starting point for such developments. This approach would allow the model to better understand dynamic phenomena. For example, one possibility is fine-tuning the model to predict crack dynamics, estimating crack propagation and damage, as already shown in one of the examples in the paper (Figure~\ref{fig:Fig_2061}). While our example did not use video data, it demonstrated the model's ability to comprehend information trajectories from a snapshot in time, highlighting the potential flexibility of this approach. We anticipate that the results would be improved if we were to provide multiple images or a video sequence, perhaps encoded in a token-efficient manner, for analysis.

In terms of parameter count, models like GPT-4-Vision or GPT-4o are significantly larger than Cephalo, with estimates suggesting they include trillions of parameters. This makes them many orders of magnitude larger than Cephalo with only a few billion parameters. As a result, Cephalo may have some shortcomings, including a more limited capacity for extremely complex tasks compared to larger models like those in the GPT-4 class. It may require more fine-tuning for specialized applications and its performance depends on the quality of training data, or the construction of MoE models to improve its behavior. 
Despite its smaller size, Cephalo performs comparably well to GPT-4-class models in various applications and gives accurate, detailed and clear responses. This behavior is partly based off the teacher-student strategy for dataset construction utilizing the sophistication of GPT-4o in training data development so that the model learns both domain knowledge and the reasoning capabilities of frontier models. However, because of the much smaller parameter count, Cephalo is much more efficient, with a lower carbon footprint for training and inference, and its development process is transparent, accessible and adaptable. Cephalo can be run on local consumer hardware, including laptops, making it particularly useful for local applications or edge computing, such as in self-driving labs, or even remote field work.

Another future direction would be to focus on interpretability, for example using the ``Scaling Monosemanticity" approach~\cite{templeton2024scaling} to better understand how Cephalo learns bio-inspired materials phenomena. This analysis could employ sparse autoencoders trained on layer-wise activations of Cephalo models to identify monosemantic neurons (those consistently activating for a single, specific concept or feature regardless of context). We would quantify monosemanticity using scoring metrics and entropy of label distributions, guided by a tailored taxonomy of bio-inspired materials concepts~\cite{bandrowski2016ontology,cranford2012biomateriomics,Giesa2012CategoryDesign,buehler2024accelerating}. The multimodal nature of Cephalo would further necessitate an analysis of both text and vision components, potentially using Centered Kernel Alignment ~\cite{kornblith2019similarity} to reveal cross-modal monosemantic neurons, which would offer a fascinating direction of research. Such an analysis may yield insights into Cephalo's internal representation of bio-inspired materials concepts, informing future architectural optimizations for specialized scientific AI models and modeling approaches of bio-inspired methods. To control Cephalo's behavior for specific use cases, we could identify monosemantic neurons related to certain bio-inspired materials concepts and develop a method to directly manipulate their activations during inference. This could allow users to emphasize specific bio-inspired principles (e.g., ``self-healing mechanisms" or ``hierarchical structures") during material design tasks, potentially steering Cephalo's outputs towards targeted solutions in materials science applications. Since access to parameters and the source code of the model architecture is key for this type of work, the open source nature of this model is an important enabling feature. 

\section{Materials and methods}

We summarize materials and methods used in this work. 

\subsection{Etymology and inspiration behind the name ``Cephalo"}
The name "Cephalo" is derived from the Greek word kephalē meaning ``head" or ``brain," which symbolizes the model's central role in processing and integrating visual and linguistic information. This name reflects the model's function as the ``brain" of the system, facilitating advanced human-AI and multi-agent AI interactions through the comprehensive understanding of multimodal data.
Additionally, ``Cephalo" draws inspiration from cephalopods, a class of intelligent mollusks that includes octopuses, squids, and cuttlefish, associating it with the focus on biological inspiration that is central to the training and use of the model. Cephalopods are renowned for their exceptional cognitive abilities, advanced problem-solving skills, and highly developed nervous systems. They exhibit remarkable adaptability to their environments, sophisticated camouflage techniques, and complex behaviors, and are well-equipment to integrate visual cues with materialization.
By naming our multimodal materials science V-LLM ``Cephalo," we evoke the intelligence and adaptability of cephalopods. Similar to how cephalopods process diverse sensory inputs to navigate and respond to their surroundings, Cephalo integrates and processes visual and linguistic data to handle complex tasks. This dual inspiration highlights the model's potential for advanced problem-solving and contextual comprehension, drawing parallels between the cognitive prowess of cephalopods and the model's capabilities in the realm of materials science and beyond.

\subsection{Dataset generation}

Dataset generation is a key part, as scientific-focused vision models must be trained with focused and well-reasoned data that combines visual and text information than general-purpose datasets. To that end we develop a new dataset that incorporates data from Wikipedia, focused on a particular area of interest, with data extracted from scientific papers. 

\subsubsection{Data extraction from Wikipedia}

We implemented a multimodal text/vision model for both inference and training, utilizing a dataset of images and LLM-processed data scraped from Wikipedia. The dataset generation code scrapes Wikipedia for images based on a set of search terms, leveraging generative AI to enhance the description of images for training purposes. This approach allows for the integration of complex logic in expanding image descriptions, thus enriching the training dataset. Once the image extraction algorithm is finalized, we will apply the same method to process figure-caption pairs, preparing the dataset for comprehensive testing and further model development.

We search for Wikipedia articles with these keywords:
\begin{lstlisting}
Bioinspired materials, Mechanics, Mechanical engineering, Engineering, bone, spider, spider web, molecular modeling, dynamics, amino acid, hierarchical structure, seashell, Quantum mechanics, Molecular mechanics, Fluid dynamics, Solid mechanics, Mechanical properties of materials, Chemical engineering, Civil engineering structures, Electrical engineering innovations, Mechanical engineering design, Biomedical engineering, Biomimicry, Nanotechnology, Nanoscience, Biomaterials, Proteins, Biophysics, Biomaterials for medical applications, Biomimetic architecture, Bioinspired robotics, Sustainable materials from nature, Molecular biology, Cellular biology, Evolutionary biology, Biochemistry, Industrial chemistry
\end{lstlisting}

For each search keyword, we identify 100 highest hits using the Wikipedia API \url{https://en.wikipedia.org/w/api.php}. We access each of these sites using \url{https://en.wikipedia.org/wiki/<SEARCH_HIT>} and search each of these for image-caption pairs, by looping over all images. Then, we use the \texttt{lamm-mit/Bioinspired-Phi-3-mini-4k} to rewrite and expand on the original caption found in the Wikipedia site:
\begin{figure}[H]
\begin{LLMbox}{} 
\begin{lstlisting}
You follow directions. Do NOT repeat the question or task. Your job is to rewrite image descriptions in a precise way, using scientific principles.   Your responses are concise but accurate, and include logic and reasoning.

Rewrite this description: ```{caption}```.  Make sure that a complete description is provided, accurate and concise.  Do NOT provide any figure or image number, citations, or references, just a clear description of what or who is shown. 

Provide a succinct description, start with "The image shows..." or a variation thereof.
\end{lstlisting}
\end{LLMbox}
\end{figure}

\begin{figure}
    \centering
    \includegraphics[width=1\linewidth]{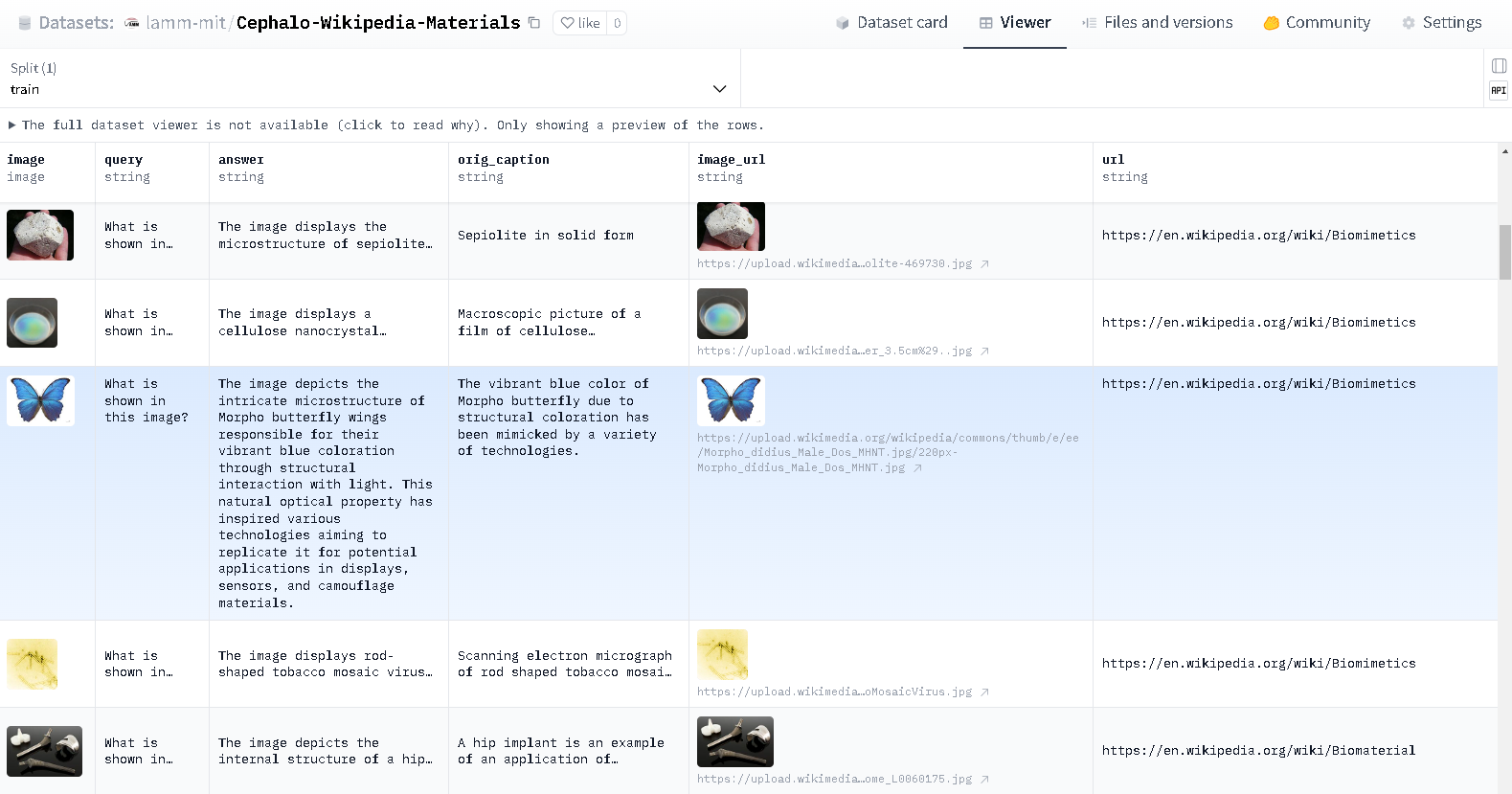}
    \caption{Snapshot of the Wikipedia image-text-dataset (\texttt{lamm-mit/Cephalo-Wikipedia-Materials} at Hugging Face), revealing the organization of the data. Each image (left panel) is associated with a query-answer pair. The original caption is shown as well, and a comparison of it with the answer developed by \texttt{lamm-mit/Bioinspired-Phi-3-mini-4k} shows how the LLM adds important context to the description that goes beyond what was originally included in the Wikipedia pages (captions on Wikipedia tend to be short and do not include full context). The dataset also includes references to the image URL, as well the URL of the Wikipedia article from which it was extracted. }
    \label{fig:Fig_1005}
\end{figure}

Figure~\ref{fig:Fig_1005} shows a screenshot of the Wikipedia image-text-dataset (see, \texttt{lamm-mit/Cephalo-Wikipedia-Materials} at Hugging Face). As can be seen, each image (left panel) is associated with a query-answer pair. The original caption is shown as well, and a comparison of it with the answer developed by \texttt{lamm-mit/Bioinspired-Phi-3-mini-4k} shows how the LLM adds important context to the description that goes beyond what was originally included in the Wikipedia pages. As one can confirm by exploring the dataset, captions on Wikipedia tend to be short and do not include full context, and hence, the additional processing into more elaborate question-answer pairs provides rich context and a higher quality training data. For completeness, and further additions or processing of the data, the dataset also includes references to the image URL, as well the URL of the Wikipedia article from which it was extracted. 

The total size of the dataset includes arund 7,500 image-text pairs. 

\subsubsection{Image-focused data extraction to identify image-caption pairs from scientific papers}

To develop a robust dataset generation method, we implemented a ``from scratch" algorithm using  \textbf{PyMuPDF} \url{https://github.com/pymupdf/PyMuPDF}. The process begins by identifying all images on each page of a PDF. Subsequently, we locate text blocks that start with ``Fig" or similar identifiers. The algorithm then matches these text blocks with the nearest image located below them. This matching process is refined through several clean-up steps, including handling different image colormaps and formats, removing specific symbols, such as those added to documents by journals, and ignoring images that cannot be properly extracted (e.g. split images with high aspect ratio). 
The main function operates as follows:

\begin{enumerate}
    \item Open the PDF document using PyMuPDF.
    \item Iterate through each page of the document.
    \item For each page:
    \begin{enumerate}
        \item Extract text blocks that start with ``Figure" or ``Fig", which are likely to be captions.
        \item Extract all images on the page.
        \item For each image:
        \begin{enumerate}
            \item Apply filtering to exclude certain image types.
            \item Extract the image's bounding box and other metadata.
            \item Find the nearest caption below the image using the following distance metric:
            \begin{equation}
                d = \sqrt{(y_{\text{caption}} - y_{\text{image}})^2 + (x_{\text{caption}} - x_{\text{image}})^2}
            \end{equation}
            where $(x_{\text{image}}, y_{\text{image}})$ are the coordinates of the image's lower-left corner, and $(x_{\text{caption}}, y_{\text{caption}})$ are the coordinates of the caption's top-left corner.
            \item If a caption is found, save the image and its associated metadata.
        \end{enumerate}
    \end{enumerate}
    \item Close the PDF document.
\end{enumerate}

The core of the algorithm can be represented by the following pseudocode:
{\small
\begin{verbatim}
Function ExtractImagesWithCaptions(pdf_path):
    Open PDF document
    For each page in document:
        Extract text blocks starting with "Figure" or "Fig"
        Extract all images on the page
        For each image:
            If image not in exclusion list:
                Try:
                    Get image bounding box and metadata
                    Initialize nearest_caption to None
                    Initialize min_distance to infinity
                    
                    For each text block:
                        Calculate distance between image and text block
                        If text block is below image and distance < min_distance:
                            Update min_distance
                            Set nearest_caption to current text block
                    
                    If nearest_caption is found:
                        If image is usable:
                            Process and save the image
                            Add image and metadata to dataset
                Catch Exceptions:
                    Log issues 
                    
    Close PDF document
    Return dataset and pages with issues
\end{verbatim}
\normalsize}
This algorithm effectively extracts images from PDF documents and associates them with their captions, creating a structured dataset for further analysis or machine learning tasks. The implementation includes error handling and supports various image formats and color spaces.

\subsubsection{Image-focused data extraction from raw image-text pairs}
Based on the results of raw image-text pairs identified with the method described in the previous section, we next utilize an algorithm to yield detailed, well-reasoned image-text pairs for training. This is accomplished by sharing the image and original caption with a general-purpose V-LLM, and tasking the model to develop a comprehensive description of the image. We use both an open source V-LLM, \texttt{Idefics2}, and \texttt{GPT-4o}, for this task. 
As an alternative, we also explored using a text-only LLM (e.g., \texttt{Phi-3-Bioinspired}, which provides another option (akin do what was done for the Wikipedia dataset). However, using vision-capable LLMs to process and distill datasets is generally better and provides more detailed and well-reasoned descriptions. 

The process of constructing the image-text pairs for the training set follows this approach:
\begin{LLMbox}{} 
\begin{lstlisting}
You follow directions. Do NOT repeat the question or task. Your job is to rewrite image descriptions in a precise way, using scientific principles. Your responses are concise but accurate, and include logic and reasoning. 

Review this caption for the image: ```{caption}```. Now, rewrite it to state only facts, as a summary of what is shown in the image. Provide a detailed  description of the image, starting with "The image shows..", "Shown in the image is.." or similar.  Include details of content of what you can see in the image. 

The response is: 
\end{lstlisting}
\end{LLMbox}
Therein, \texttt{\{caption\}} denotes the caption extracted from the PDF file. 

\begin{figure}
    \centering
    \includegraphics[width=1\linewidth]{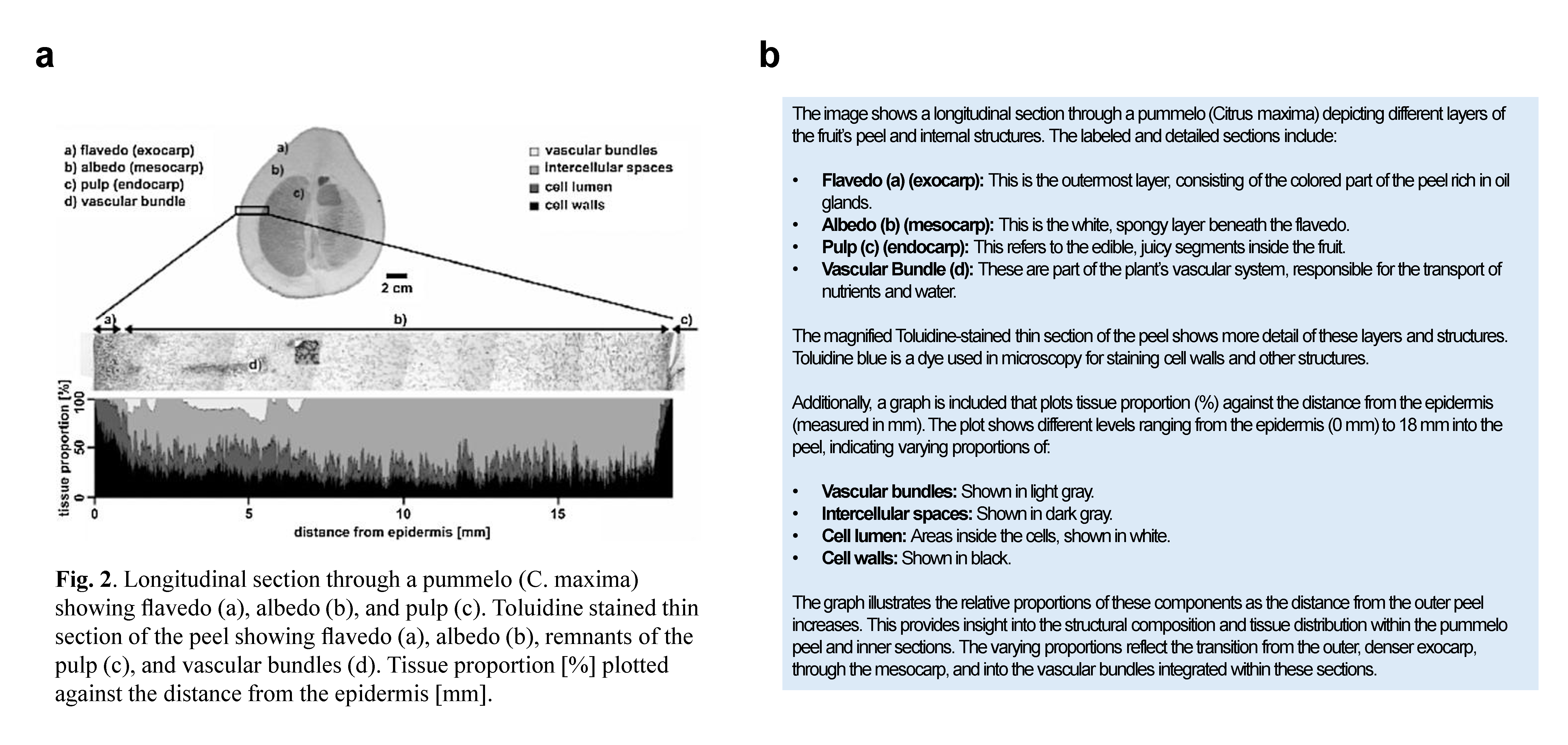}
    \caption{Example image and caption for extraction of image-text data. Panel a shows the original image and caption, and panel b the processed result that is used to form the training set. Image and caption reproduced with permission from~\cite{Fischer2010PummelosProperties}}.
    \label{fig:Fig_10}
\end{figure}

As a specific example, we show the image depicted in Fig.~\ref{fig:Fig_10}, including the original caption. The description of what is shown in the image obtained using \texttt{Idefics2} is:
\begin{figure}[H]
\begin{LLMbox}{Idefics2}
\begin{lstlisting}
The image shows a longitudinal section through a pummelo (C. maxima) with the flavedo (a), albedo (b), and pulp (c) visible. A toluidine stained thin section of the peel is also present, showing the flavdeo (a), albedo (b), remnants of the pulp (c), and vascular bundles (d). The tissue proportion [%] is plotted against the distance from the epidermis [mm]. 
\end{lstlisting}
\end{LLMbox}
\end{figure}

For knowledge extraction from image-caption pairs using \texttt{GPT-4o} this we use a slight variation of the prompt, reflecting a far greater capability of the model to reason:
\begin{figure}[H]
\begin{LLMbox}{}
\begin{lstlisting}
You follow directions. Do NOT repeat the question or task. Your job is to rewrite image descriptions in a precise way, using scientific principles. Your responses are concise but accurate, and include logic and reasoning. 

Carefully consider the image and the caption for the image: ```{caption}```. 

Now, write a summary of what is shown in the image.  State only facts, as a summary of what is shown in the image. 

Provide a detailed  description of the image, starting with "The image shows..", "Shown in the image is.." or similar. 

Include details of the content of what you can see in the image. Define any terms, acronyms or specific technical words that are used. 

If scientific results are shown, explain the logical reasoning behind the results.

Describe what the results shown in the image mean, if applicable.

The response is: 
\end{lstlisting}
\end{LLMbox}
\end{figure}

Using \texttt{GPT-4o}, we obtain:
\begin{figure}[H]
\begin{LLMbox}{GPT-4o}
\begin{lstlisting}
The image shows a longitudinal section through a pummelo (Citrus maxima) depicting different layers of the fruit's peel and internal structures. The labeled and detailed sections include:

- Flavedo (a) (exocarp): This is the outermost layer, consisting of the colored part of the peel rich in oil glands.
- Albedo (b) (mesocarp): This is the white, spongy layer beneath the flavedo.
- Pulp (c) (endocarp): This refers to the edible, juicy segments inside the fruit.
- Vascular Bundle (d): These are part of the plant's vascular system, responsible for the transport of nutrients and water.

The magnified Toluidine-stained thin section of the peel shows more detail of these layers and structures. Toluidine blue is a dye used in microscopy for staining cell walls and other structures.

Additionally, a graph is included that plots tissue proportion (%) against the distance from the epidermis (measured in mm). The plot shows different levels ranging from the epidermis (0 mm) to 18 mm into the peel, indicating varying proportions of:

- Vascular bundles: Shown in light gray.
- Intercellular spaces: Shown in dark gray.
- Cell lumen: Areas inside the cells, shown in white.
- Cell walls: Shown in black.

The graph illustrates the relative proportions of these components as the distance from the outer peel increases. This provides insight into the structural composition and tissue distribution within the pummelo peel and inner sections. The varying proportions reflect the transition from the outer, denser exocarp, through the mesocarp, and into the vascular bundles integrated within these sections.
\end{lstlisting}
\end{LLMbox}
\end{figure}
Compared to the earlier response, this model is capable of providing a more well-developed description. Because the model is generally much more capable, it can provide nuanced data and accurately combine the caption with additional background knowledge and logical reasoning in the development of the description. The difference is also evident by looking at the token lengths, comparing answers derived using \texttt{Idefics-2} and \texttt{GPT-4o} (Figure~\ref{fig:Fig_1001}d-e). 

The total size of each of the datasets developed from  \texttt{Idefics-2} and \texttt{GPT-4o} includes arund 7,500 image-text pairs.

\subsubsection{Text-focused dataset}
To complement the image-text data, we also train some models with text-only data. The datasets are taken from~\cite{Luu2023BioinspiredLLM:Materials} and \cite{Buehler2023MechGPTModalities_fixed} (\url{lamm-mit/MechanicsMaterials}), and cover a detailed training set from scientific papers in bioinspired materials, as well as mechanics and materials, especially failure mechanics. The dataset consists of around 30,000 question-answer pairs extracted from scientific papers and books. 

\subsection{Model training}

The models are developed based on two pre-trained V-LLMs:
\begin{itemize}
    \item \texttt{Idefics2}, an open-source multimodal model designed to process arbitrary sequences of image and text inputs to generate text outputs.
    \item \texttt{Phi-3-Vision}, an open-source multimodal model that can reason over image and text inputs to generate text outputs, with a large context length of 128,000 tokens.
\end{itemize}
The \texttt{Idefics2} based models are trained on smaller datasets, focused on image-text data only. We found that the model struggled to learn well from combined text and image-text data. In contrast, the \texttt{Phi-3-Vision} based model trained well on the complex, integrated dataset and yielded a highly capable model overall. 

In spite of these differences we find that all resulting models are versatile and can answer questions related to images, describe visual content, create narratives based on multiple images, or function purely as a language model without any visual inputs. They can also be used in complex pipelines that involve image analysis, reasoning, and generation of prompts for image generation in a multi-agent setup. 

A summary of key training parameters is provided in Table \ref{tab:training_params_combined} for both the 8b and 4b parameter models.
\begin{table}[h!]
\scriptsize 
\centering
\caption{Training parameters used in the development of the models, including low-rank adapter (LoRA) parameters~\cite{Hu2021LoRA:Models} (see also Figure~\ref{fig:Fig_99}a for a schematic.).}
\begin{tabular}{|p{3cm}|p{4cm}|p{4cm}|}
\hline
\textbf{Parameter} & \textbf{8b Model} & \textbf{4b Model} \\
\hline \hline 
Learning rate & 1E-5 & 1E-5 \\ \hline
Maximum gradient norm & 0.5 & 0.5 \\ \hline
Gradient accumulation steps & 4 & 4 \\ \hline
Learning rate scheduler & cosine & cosine \\ \hline
Batch size & 1 or 2 (depending on GPU VRAM) & 1 \\ \hline
Training epochs & 2 & 1 \\ \hline
Low-rank adapter parameters (in decoder) & $r=8$, $\alpha=16$, dropout=0.1 & $r=8$, $\alpha=16$, dropout=0.1 \\ \hline
Low-rank adapter (LoRA) target modules (in decoder) & \texttt{down\_proj}, \texttt{gate\_proj}, \texttt{up\_proj}, \texttt{k\_proj}, \texttt{q\_proj}, \texttt{v\_proj}, \texttt{o\_proj} & \texttt{qkv\_proj}, \texttt{o\_proj}, \texttt{down\_proj}, \texttt{gate\_up\_proj} \\ \hline
\end{tabular}
\label{tab:training_params_combined}
\end{table}

\subsection{Prompt templates}

The 8b models (as well as 10b and 12b models) use this prompt template:
\begin{LLMbox}{Cephalo-8b series chat template}
\begin{lstlisting}
User:<image>{prompt_1}<end_of_utterance>
Assistant:{response_1}<end_of_utterance>
User:{prompt_2}<end_of_utterance>
Assistant:{response_2}<end_of_utterance>
...
\end{lstlisting}
\end{LLMbox}

The 4b models use this prompt template:
\begin{LLMbox}{Cephalo-4b series chat template}
\begin{lstlisting}
<|user|>
<|image_1|><|image_2|>...
{prompt_1}<|end|>
<|assistant|>
{response_1}<|end|>
<|user|>
{prompt_2}<|end|>
<|assistant|>
{response_2}<|end|>
...
\end{lstlisting}
\end{LLMbox}
\subsection{Protein structure visualizations}
Protein structures were visualized using PyMol \url{https://pymol.org/}.

\subsection{Text-to-image generation}
We use Stable Diffusion XL Turbo (\url{https://huggingface.co/stabilityai/sdxl-turbo}) to create 2D images from text prompts generated by Cephalo-Phi-3-vision-128k-4b-beta.

\subsection{Text-to-3D model generation}
We use Meshy \url{https://www.meshy.ai/about} to turn a text prompt generated by Cephalo-Phi-3-vision-128k-4b-beta into a 3D model. We use the `Text to 3D' function in Meshy, first create a draft model and then refine in the highest possible quality. The model renders a 3D model with texture information.

\subsection{Developing a deeper Cephalo model through model merging}

Model merging involves combining components from different models to create a new, hybrid model. This method leverages the strengths of each contributing model, often resulting in improved performance and new capabilities not present in the individual models.

\subsubsection{Merging approach and implementation}

We created the \texttt{Cephalo-Idefics-2-vision-10b-alpha} model by merging layers from two distinct pre-trained models: \texttt{lamm-mit/Cephalo-Idefics-2-vision-8b-beta} and \texttt{HuggingFaceM4/idefics2-8b-chatty}. The process involved the following steps. 

First, we load the two models into memory. The domain-specific fine-tuned model, \texttt{lamm-mit/Cephalo-Idefics-2-vision-8b-beta}, is referred to as \texttt{model\_1}, while the general chat/instruction fine-tuned model, \texttt{HuggingFaceM4/idefics2-8b-chatty}, is referred to as \texttt{model\_2}. 
{\small

\begin{verbatim}
from transformers import Idefics2ForConditionalGeneration

model_1_id = 'lamm-mit/Cephalo-Idefics-2-vision-8b-beta'
model_2_id = 'HuggingFaceM4/idefics2-8b-chatty'

vtype = torch.bfloat16 #or other type
model_1 = Idefics2ForConditionalGeneration.from_pretrained(model_1_id, torch_dtype=vtype) 
model_2 = Idefics2ForConditionalGeneration.from_pretrained(model_2_id, torch_dtype=vtype)
\end{verbatim}
\normalsize}
One can use a variety of merging strategies. Here, we use \texttt{model\_1} in its entirety and add a set of layers from \texttt{model\_2} to it. For the \texttt{lamm-mit/Cephalo-Idefics-2-vision-10b-alpha} model we merge the last $N_m=8$ layers of the second model. Since the model has $N_t=32$ layers, layers 24 to 31 are chosen  (as discussed in the main text, later layers are used  due to their strong contextual understanding and generalization capabilities). Note, we explored this approach for various combinations and have seen generally excellent performance. For instance, we also developed a 12b model by setting $N_m=16$, that is, the new model consists of the entire Model\_1 and half of Model\_2, resulting in \texttt{lamm-mit/Cephalo-Idefics-2-vision-12b-alpha}.  
{\small

\begin{verbatim}
selected_layers_from_model2 = list(model_2.model.text_model.layers)[N_t-N_m:N_t]
\end{verbatim}
\normalsize}
The selected layers from \texttt{model\_2} are then concatenated with all layers from \texttt{model\_1}. This created a new sequence of layers that integrates the specialized features of \texttt{model\_1} with the general capabilities of \texttt{model\_2}.
{\small

\begin{verbatim}
combined_layers = nn.ModuleList(list(model_1.model.text_model.layers) +  
                                     selected_layers_from_model2)
\end{verbatim}
\normalsize}
A new model instance (\texttt{model\_merged}) is created by deep copying \texttt{model\_1}, and the combined layers are then assigned to this new model:
{\small

\begin{verbatim}
import copy

model_merged = copy.deepcopy(model_1)

model_merged.model.text_model.layers = combined_layers
\end{verbatim}
\normalsize}
The model is subsequently saved and then fine-tuned. For fine-tuning, we freeze all layers except for the selected layers from Model\_2:
{\small
\begin{verbatim}
for name, param in model.named_parameters():
    # Freeze all parameters
    param.requires_grad = False

# Unfreeze the last N_m layers of the text_model
for i in range(-N_m, 0):  # Last N_m layers
    layer = model.model.text_model.layers[i]
    for param in layer.parameters():
        param.requires_grad = True

# Optionally, verify the layers that will be trained
for name, param in model.named_parameters():
    print(f"{name}: {param.requires_grad}")
\end{verbatim}
\normalsize}
A detailed example is provided via \url{https://github.com/lamm-mit/Cephalo}.

\subsubsection{Merged model fine-tuning}

As described in the preceding section, the last $N_m$ layers of the new hybrid model were fine-tuned. This fine-tuning step is crucial as it helped the new model adapt and align the strengths of the merged layers, ensuring that the model performed well in both domain-specific and general tasks. A learning rate of $2E-5$ was used, with warm-up and cosine learning rate decay.

\subsection{Sparse Mixture of Experts based on the 4b models built on \texttt{Phi-3-Vision}}

As another way to create larger, more complex models from smaller ones we employ a sparse mixture of experts (MoE) model to enhance the model. This model leverages multiple expert networks to process different parts of the input, allowing for more efficient and specialized computations. For each token in the input sequence, a gating layer computes scores for all experts and selects the top-$k$ experts based on these scores. We use a \texttt{softmax (..)} activation function to ensure that the weights across the chosen experts sum up to unity. 

The output of the gating layer is a set of top-$k$ values ($\mathbf{T}$) and their corresponding indices ($\mathbf{I}$). The selected experts' outputs ($\mathbf{Y}$) are then computed and combined using a weighted sum, where the weights are given by the top-$k$ values. Mathematically, the output for each token $(b, s)$ in the batch is given by:
\[
\mathbf{O}_{b, s} = \sum_{i=1}^{k} \mathbf{T}_{b, s, i} \cdot f_{\mathbf{I}_{b, s, i}}(\mathbf{X}_{b, s})
\]
where $\mathbf{T}_{b, s, i}$ is the weight from the gating function for the $i$-th selected expert, and $f_{\mathbf{I}_{b, s, i}}(\mathbf{X}_{b, s})$ is the output of the $i$-th selected expert for the token $(b, s)$. This sparse MoE mechanism allows our model to dynamically allocate computational resources, improving efficiency and performance for complex vision-language tasks. Figure~\ref{fig:Fig_2013} depicts an overview of the architecture. 

The code implements the creation of the MoE model directly in custom model code. For instance, as shown via the repository \texttt{lamm-mit/Cephalo-Phi-3-MoE-vision-128k-3x4b-beta}, we can easily construct a model consisting of three Phi-3 models: 
\begin{verbatim}
# Initialize the models
expert_models = [model_1, model_2,  model_3  ]  # List of expert models
base_model = copy.deepcopy(model_2)  # Base model, here chosen to be model_2 
 
# Load a processor (e.g. from base model)
processor = AutoProcessor.from_pretrained(model_name_2, trust_remote_code=True) 

# Create the config
config =  AutoConfig.from_pretrained(model_name_2, trust_remote_code=True)

# Create the MoE model
moe_config = Phi3VForCausalLMMoEConfig(config=config, 
    k=1, num_expert_models=len (expert_models))
moe_model = Phi3VForCausalLMMoE(moe_config, base_model, expert_models,  layer_dtype = dtype).to(device)
\end{verbatim}

\subsubsection{Construction mechanics of the MoE model}

To transform an existing model into a Mixture of Experts (MoE) model, we first take the base model use a set of fine-tuned or otherwise trained models to create multiple expert models. Typically, each of the expert models specializes in different aspects of the input data, allowing for greater flexibility and efficiency in processing. To implement this, the original layers of the base model are replaced with modified layers that incorporate the gating and expert mechanisms. A custom configuration class is created to extend the base configuration, adding parameters specific to the MoE setup, such as the number of experts and the number of experts to select in each forward call ($k$). 

Within the algorithm, the original MLP layers in the base model are replaced with a new MoE layer that combines the outputs of the selected experts. This MoE layer uses the gate scores to select the relevant experts' outputs and combines them into a single output by computing a weighted sum. The modified layers are then integrated back into the model, creating a hybrid architecture that retains the original model's structure but with enhanced capabilities. The code is provided at \url{https://github.com/lamm-mit/Cephalo-Phi-3-MoE}, providing further details.

\subsubsection{Training the gating network through sample prompts}

The gating networks play a crucial role in determining the appropriate experts for each input token. These gating networks are trained using sample prompts that represent a diverse set of inputs. For each prompt (text only, or interleaved text-image prompts), we compute the hidden states in each layer and use them to train the gating layer to produce appropriate expert selection scores. The training process involves optimizing the gating layer to minimize the cross-entropy loss between the predicted expert distribution and the target distribution, which encourages the gating network to learn to select the most relevant experts for a given input.  The gating layers are trained using collected hidden states from a set of prompts, optimizing the gate values to ensure the effective selection and weighting of expert outputs.

We define a dataset (defined as a Python dictionary of text-image pairs):
{\small
\begin{verbatim}
prompts_per_expert = [
    [{"text": "<|user|>\n<|image_1|>\nPrompt 1 for expert 1<|end|>\n<|assistant|>\n", 
    "image": [image_1]}, 
     {"text": "<|user|>\n<|image_1|>\nPrompt 2 for expert 1<|end|>\n<|assistant|>\n", 
     "image": [image_1]}],

    [{"text": "<|user|>\n<|image_1|>\nPrompt 1 for expert 2<|end|>\n<|assistant|>\n", 
    "image": [image_2]}, 
     {"text": "<|user|>\n<|image_1|>\nPrompt 2 for expert 2<|end|>\n<|assistant|>\n", 
     "image": [image_2]}],

    [{"text": "<|user|>\n<|image_1|>\nPrompt 1 for expert 3<|end|>\n<|assistant|>\n", 
    "image": [image_3]}, 
     {"text": "<|user|>\n<|image_1|>\nPrompt 2 for expert 3<|end|>\n<|assistant|>\n", 
     "image": [image_3]}],
     ...
]
\end{verbatim}
\normalsize}
The gating network can be trained as follows:
{\small
\begin{verbatim}
# Train gating layers using the provided prompts
gating_layer_params = moe_model.train_gating_layer_params_from_hidden_states(
                                              processor, 
                                              prompts_per_expert,
                                              epochs=1000,
                                              loss_steps=100,
                                              lr=5e-5,
                                          )

# Set parameters in MoE model
moe_model.set_gating_layer_params(gating_layer_params)
\end{verbatim}
\normalsize}

While this training function offers good performance, further training may be beneficial especially for deployment of multiple active experts.

\subsection{Predicting statistics of deformation fields and dynamics in cracked materials}

In this section, we describe the methodologies used to enhance Cephalo's capabilities in predicting the statistical properties of deformation fields and crack dynamics in materials with defects. Accurate predictions of these properties are important for understanding the mechanical behavior of materials under stress and for designing materials with improved performance and durability. We leverage a combination of datasets from previous work to develop predictive models that can analyze complex microstructures and provide quantitative insights into material behavior based on molecular dynamics simulation results.

We use instruction fine-tuning to train models. In the context of vision models, this refers to the process of adapting pre-trained models to follow specific instructions or prompts when processing visual input. This technique combines the capabilities of large vision models with the flexibility of natural language instructions, allowing the model to perform a wider range of specific, highly technical vision-related tasks based on textual prompts, here focused on reasoning over fracture related phenomena. 

\subsubsection{Deformation field statistics in cracked samples under tensile loading}
\label{stat_instruction_tuned}
We use a dataset developed in earlier work~\cite{Buehler2022FieldPerceiver:Ologs} (\url{https://huggingface.co/datasets/lamm-mit/crack_stress_field_graphene_uniaxial_loading_x}, see Figure~\ref{fig:Fig_2089} for a snapshot of the dataset) that consists of various graphene flakes with cracks, and associated deformation fields computed using molecular dynamics modeling. For this study, we develop a capability to predict, given the input microstructure, various statistical properties of the resulting Von Mises stress and atomic potential energy fields. See Figures~\ref{fig:Fig_2002} and \ref{fig:Fig_2002_epot}, which show samples of Von Mises stress and atomic potential energy distributions from MD simulations that serve as the training set, in different intervals of standard deviation  and mean.  The model is trained based off the \url{lamm-mit/Cephalo-Phi-3-vision-128k-4b-beta} model described earlier.  

\begin{LLMbox}{Instructions to predict statistical properties of stress and energy fields from microstructure}
\begin{lstlisting}
(*@\includegraphics[width=.1\linewidth]{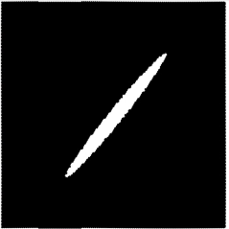}@*)
Query:      CalculateVonMisesStressStatistics <stress_stdev, stress_mean, stress_median>
Answer:     [0.678, 0.603, 0.624]

(*@\includegraphics[width=.1\linewidth]{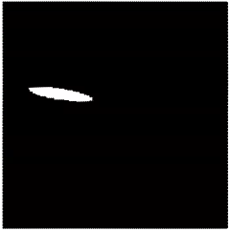}@*)
Query:      CalculatePotentialEnergyStatistics <energy_peratom_std_dev, energy_peratom_mean, energy_peratom_median>
Answer:     [0.467, 0.875, 0.940]
\end{lstlisting}
\end{LLMbox}
The training data includes around 2,000 samples, and a train-test split of 0.9:0.1 is used for the analysis. Fine-tuning the model for a specific task can be accomplished within a few hours on a single consumer-grade GPU. The \texttt{gate\_up\_proj} and  \texttt{down\_proj} layers are used as target for LoRA adaptation ($r=8$, $\alpha=16$, dropout=0.1).

\subsubsection{Predicting crack dynamics: Damage and initiation}

For this task we use a dataset developed in earlier work~\cite{Buehler2022ModelingModel} (\url{https://huggingface.co/datasets/lamm-mit/crack-dynamics_1-5k}, see Figure~\ref{fig:Fig_2090} for a snapshot of the dataset) that consists of frames obtained from a molecular dynamics simulation of a small cracked crystal exposed to uniaxial loading, based on a Lennard-Jones model crystal. The dataset includes the initial microstructure, the final microstructure, and whether or not the crack has begun to spread. 
Damage incurred between the first and last frame of the simulation is computed by first computing the proportion of differing pixels (\textrm{diff\_prop}) in the visualization of the potential energy field. This is calculated by identifying pixels where the color distance between corresponding pixels in the two images exceeds a  threshold. The proportion is given by dividing the count of these differing pixels by the total number of pixels in the image. This \textrm{diff\_prop} value is then normalized to obtain the damage measure, which scales the values to a range between 0 and 1. The normalization ensures that the minimum damage observed in the dataset maps to 0, and the maximum damage maps to 1.

We use instruction fine-tuning, where we show the model an image, an instruction query, and a result. The result is converted into numerical floating point values to render the plots, but we use True/False as initiation metric.  The model is trained based off the fine-tuned model developed in the previous section (Section~\ref{stat_instruction_tuned}).

\begin{LLMbox}{Instructions to predict damage and crack initiation from microstructure}
\begin{lstlisting}
(*@\includegraphics[width=.1\linewidth]{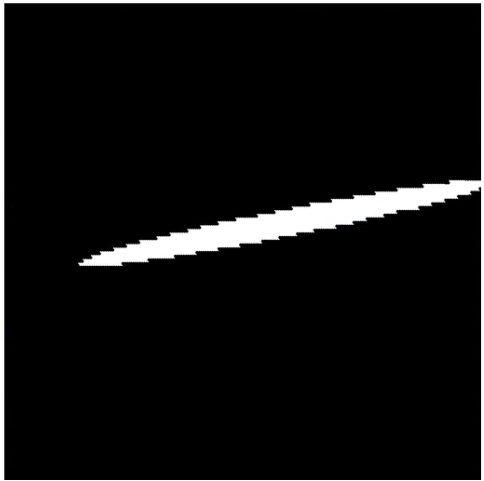}@*)
Query:      CalculateCrackDynamics <damage, initiate>
Answer:     [0.139, True]
\end{lstlisting}
\end{LLMbox}
The training data includes around 1,500 samples, and a train-test split of 0.9:0.1 is used for the analysis.  As in the previous example, the \texttt{gate\_up\_proj} and  \texttt{down\_proj} layers are used as target for LoRA adaptation (with $r=16$, $\alpha=16$, dropout=0.1).

We use several key metrics to assess the performance of our model, including accuracy, precision, recall, and the F1-score~\cite{han2011data}.Accuracy measures the proportion of correctly classified instances out of the total instances, providing an overall effectiveness of the model. Precision, defined as the proportion of true positive predictions out of all positive predictions, evaluates the accuracy of the positive predictions made by the model. Recall (true positive rate), measures the proportion of true positive predictions out of all actual positive instances, indicating the model’s ability to identify all relevant positive instances. The F1-score, which is the harmonic mean of precision and recall, provides a single metric that balances both precision and recall, making it particularly useful for imbalanced datasets. These metrics collectively offer a good overview of the  performance.

\subsection{Fast inference}
Fast inference for the models presented in this paper is implemented in \texttt{mistral.rs} written in Rust (\url{https://github.com/EricLBuehler/mistral.rs}, see also~\cite{Buehler2024X-LoRA:Design} for the original development and benchmarks). Performance enhancements are achieved through the use of optimized CUDA kernels, efficient KV-cache management via PagedAttention~\cite{kwon2023efficientmemorymanagementlarge}, and \textit{in-situ} quantization that significantly reduces memory requirements with increased tokens/seconds at only modest reduction in perplexity.

Using this framework, users can either use Python bindings or launch an OpenAI API compatible server, like so:
{\small
\begin{verbatim}
./mistralrs_server --port 1234 vision-plain -m lamm-mit/Cephalo-Phi-3-vision-128k-4b-beta -a phi3v
\end{verbatim}
\normalsize}
This framework allows users to run Cephalo locally in a secure environment on a variety of supported accelerators and platforms, including various GPU architectures and edge devices like Raspberry Pi. 

\section*{Code and data availability}
Trained weights, datasets, training scripts, and other resources for the models utilized in this study can be found at \url{https://huggingface.co/lamm-mit/cephalo}.  Links to various components of this model development are provided at this Hugging Face collection. Additional codes and tools are provided at \url{https://github.com/lamm-mit/Cephalo} and \url{https://github.com/lamm-mit/Cephalo-Phi-3-MoE}. 

\section*{Author contributions}
MJB designed the research, conducted dataset generation, the model development and inference. MJB carried out all experiments,  simulations and data analysis. MJB wrote and edited the paper. 

\section*{Conflicts of interest and disclosures}
There are no conflicts to declare. Generative AI was used in the creation of this work. 

\section*{Acknowledgements}
We acknowledge support from Google, the Army Research Office (79058LSCSB, W911NF-22-2-0213 and W911NF2120130), the National Institutes of Health (NIH) (R01AR077793)), and USDA (021-69012-35978). 


\bibliographystyle{apsrev} 
\bibliography{references,references-mendeley}  

\newpage
\appendix

\pagestyle{empty} 

\renewcommand{\thefigure}{S\arabic{figure}}
\setcounter{figure}{0} 
\renewcommand{\thetable}{S\arabic{table}}
\setcounter{table}{0} 
\renewcommand{\thetextbox}{S\arabic{textbox}}
\setcounter{textbox}{0} 

\clearpage
\begin{center}
\LARGE\bfseries \section*{Supplementary Materials}
\end{center}

\clearpage

\clearpage
\begin{longtable}{|p{3cm}|p{3cm}|p{9cm}|}
\caption{Educational and future research opportunities and examples for use of Cephalo, at different complexity levels from basic to advanced, showing the advantages of an open-source platform as it allows for a variety of deeply integrated experiments and studies via the available code and weights.}
\label{tab:educational_goals} \\
\hline
\textbf{Educational Goal} & \textbf{Description} & \textbf{Learning Objectives and Model Utilization} \\
\hline
\endfirsthead
\multicolumn{3}{c}%
{\tablename\ \thetable\ -- \textit{Continued from previous page}} \\
\hline
\textbf{Educational Goal} & \textbf{Description} & \textbf{Learning Objectives and Model Utilization} \\
\hline
\endhead
\hline \multicolumn{3}{r}{\textit{Continued on next page}} \\
\endfoot
\hline
\endlastfoot
Training Set Distillation & Reducing the size of the training dataset while retaining essential information to improve efficiency and performance. & \begin{itemize}[left=0pt]
\item \textbf{Basic:} Use the datasets developed as part of this study for learning materials science concepts, utilizing pairings of images and descriptions. 
\item \textbf{Intermediate:} By reviewing the principles of data distillation used here, use the trained model to create other distilled datasets from other materials science or scientific data, such as energy applications, food or agriculture, damage mechanics, or low-dimensional materials such as MXenes.
\item \textbf{Advanced:} Analyze efficiency gains and retention of critical information in various distilled datasets. Develop a smaller, distilled dataset from a large corpus and compare the model's performance with full and distilled datasets.
\end{itemize} \\
\hline
Information-Knowledge-Decisions (exploring some of the ideas introduced in Figure~\ref{fig:Fig_1}b) & Understanding the flow from raw data to actionable insights in materials science and engineering. & \begin{itemize}[left=0pt]
\item \textbf{Basic:} Learn to convert raw materials data (e.g. defects, stress fields, damage dynamics) into actionable insights using the model's multimodal capabilities.
\item \textbf{Intermediate:} Use the model to transform fracture mechanics data into knowledge and decision-making tools for material design (e.g., make rational decisions about better material design solutions, akin to the analysis in Text Box~\ref{TexBox:TB1} and others, or assessing optimal defect sizes to prevent crack initiation akin the analysis shown in Figure~\ref{fig:Fig_2061}f-g).
\item \textbf{Advanced:} Conduct case studies on selecting materials for high-stress environments (involving varied environments, such as corrosive agents) based on the model's insights. Analyze how the model transforms raw data into structured knowledge and implement decision-making scenarios, including multi-turn conversations in agentic or human-AI collaborations.
\end{itemize} \\
\hline
Fine-Tuning Techniques & Adapting pre-trained models to specific tasks or domains in materials science. & \begin{itemize}[left=0pt]
\item \textbf{Basic:} Master fine-tuning techniques for pre-trained models, starting with regression tasks to more complex training objectives.
\item \textbf{Intermediate:} Apply fine-tuning using the model's pre-trained weights to datasets involving advanced materials, such as MXenes, high entropy alloys, and bio-inspired composites, and/or genomic data, health data, and others.
\item \textbf{Advanced:} Assess improvements in predicting mechanical properties and crack initiation behavior across multiple boundary conditions encoded in the microstructure representation, and different material classes/systems. Fine-tune the model on domain-specific datasets and evaluate performance improvements, comparing with traditional computational methods. Use the method in a multi-agent setup to solicit new physical data via reinforcement learning.
\end{itemize} \\
\hline
Model Interpretability & Understanding how the model makes predictions and derives insights in materials science contexts. & \begin{itemize}[left=0pt]
\item \textbf{Basic:} Gain skills in model interpretability and explainability, focusing on the model's multimodal outputs and reasoning behind predictions.
\item \textbf{Intermediate:} Use the model to visualize predictions of crack initiation points in materials (e.g. predict coordinates, regions of damage, or temporal details of crack movements) and interpret the model's reasoning process.
\item \textbf{Advanced:} Create detailed reports explaining the model's decision process in evaluating material properties. Develop explainability modules to make model outputs more transparent, similar to recent work in other LLMs~\cite{templeton2024scaling}. Explore the model's understanding of complex concepts like ``inner conflict" in the context of material behavior and how concepts like toughness, flaw tolerance, defect tolerance, and similar are represented.
\end{itemize} \\
\hline
Cross-Disciplinary Applications & Exploring applications of the model in various scientific and engineering disciplines beyond materials science. & \begin{itemize}[left=0pt]
\item \textbf{Basic:} Learn about the interdisciplinary applications of AI models in fields related to materials science, such as chemistry or physics (e.g. test applicability of the model to learn new concepts in context or through transitive mechanisms).
\item \textbf{Intermediate:} Conduct experiments using the model on datasets from different fields, such as biocompatibility of composites, MXenes in energy storage, high entropy alloys in aerospace, or biomedical applications of protein materials.
\item \textbf{Advanced:} Evaluate the model's performance and adaptability across different fields. Explore fine-tuning against complex phenomena, such as dynamics of causal events in chemistry and physics, and the development of multi-agent AI systems inspired by biological systems like ant colonies.
\end{itemize} \\
\hline
Multimodal Integration & Understanding and applying the model's ability to process and integrate visual and textual data. & \begin{itemize}[left=0pt]
\item \textbf{Basic:} Learn how the model combines image and text data to analyze materials science problems.
\item \textbf{Intermediate:} Use the model to analyze scientific papers, extracting key information from both text and figures, and add series of images or video data.
\item \textbf{Advanced:} Develop new methodologies for integrating diverse data types in materials research, such as combining microscopy images with spectroscopic data and textual descriptions. 
\end{itemize} \\
\hline
Predictive Modeling & Using the model for quantitative predictions in materials science. & \begin{itemize}[left=0pt]
\item \textbf{Basic:} Understand how the model can be used to predict material properties from structural data.
\item \textbf{Intermediate:} Apply the model to predict stress field statistics and crack dynamics in various materials, e.g. by fine-tuning the base model developed in the paper against new types of materials or boundary conditions (via in context-learning or fine-tuning).
\item \textbf{Advanced:} Develop and validate new predictive models for complex materials phenomena, comparing the model's performance with traditional simulation methods.
\end{itemize} \\
\hline
Bio-Inspired Design & Leveraging the model's capabilities for bio-inspired materials design. & \begin{itemize}[left=0pt]
\item \textbf{Basic:} Understand the principles of bio-inspired design in materials science.
\item \textbf{Intermediate:} Use the model to analyze biological structures and propose bio-inspired material designs.
\item \textbf{Advanced:} Develop novel bio-inspired materials by integrating the model's insights with experimental techniques and computational modeling, utilizing specifically abstractions learned by the model.
\end{itemize} \\
\hline
AI-Assisted Research & Using the model to enhance and accelerate materials science education and research. & \begin{itemize}[left=0pt]
\item \textbf{Basic:} Develop tutorials about material behaviors for interactive, tailored user experiences for learning certain concepts. Interaction with the user can include not only text input but sketches or drawings in response to questions, offering a highly dynamic and interactive approach. 
\item \textbf{Intermediate:} Learn how to use the model for literature review and data extraction in materials science and integrate the model into research workflows for hypothesis generation and experimental design, for instance integrated in graph reasoning strategies~\cite{buehler2024accelerating}. 
\item \textbf{Advanced:} Develop AI-assisted research methodologies that combine the model's capabilities with high-throughput experimentation and autonomous labs, and/or agentic workflows that incorporate multi-domain first principles simulation capabilities.
\end{itemize} \\
\end{longtable}

\clearpage

\begin{figure}[htbp]
\begin{LLMbox}{Chat between a User and \texttt{Cephalo-Idefics-2-vision-12b-alpha}}
\linespread{0.4}\selectfont
\begin{lstlisting} 
(*@\bf\hlred{User}@*): You are a materials scientist and you carefully analyze images, step by step.

Answer briefly.
(*@\includegraphics[width=.5\linewidth]{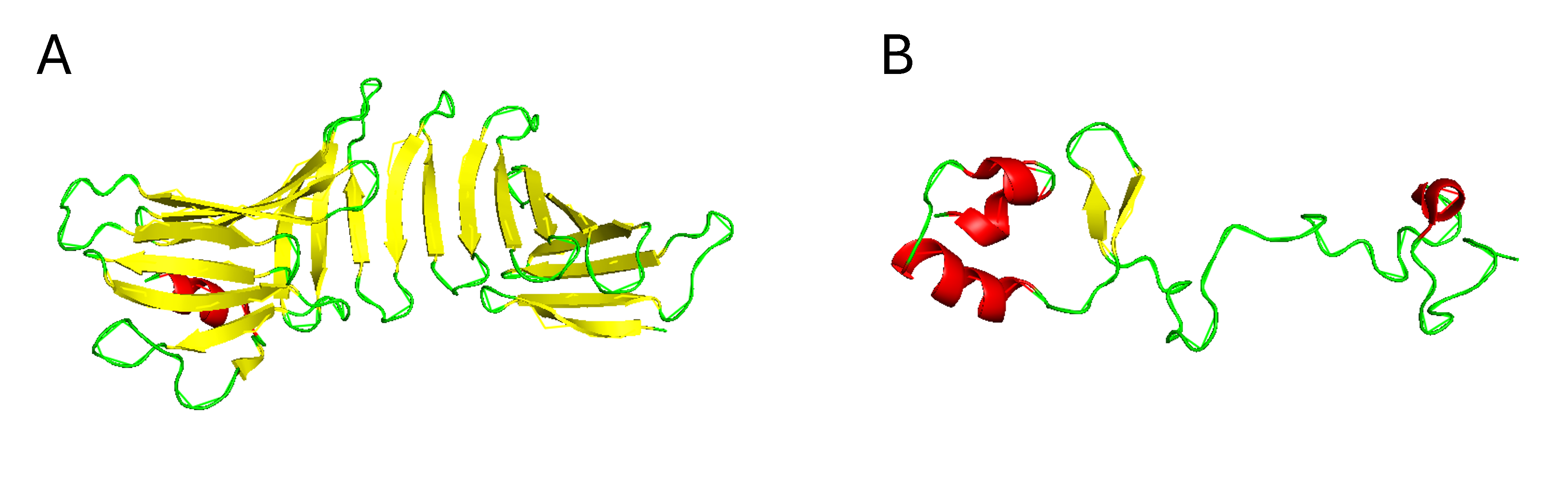}@*)
(*@\bf\hlred{User}@*): You are shown two proteins, colored by their secondary structure.

Looking at the protein images, and analyzing the structure, which protein has a more organized structure?

(*@\bf \hl{Cephalo}@*):  The protein on the left has a more organized structure, as indicated by the presence of alpha-helices and beta-sheets. These secondary structures contribute to the overall stability and function of the protein. The protein on the right, on the other hand, has a more disorganized structure, with fewer or no visible secondary structures. This could imply that the right protein is less stable or functional compared to the left protein.

(*@\bf\hlred{User}@*): Which protein (A or B) would have a higher mechanical resistance?

(*@\bf \hl{Cephalo}@*): Protein A would have a higher mechanical resistance due to its more organized structure, which includes alpha-helices and beta-sheets. These secondary structures provide strength and stability to the protein, making it more resistant to mechanical forces.

(*@\bf\hlred{User}@*): Suggest a few design improvements.

(*@\bf \hl{Cephalo}@*): To improve the design, the following changes could be made:
1. Enhance the mechanical properties by incorporating more hydrogen bonds and disulfide bridges, which would increase the stability and rigidity of the protein structure.
2. Introduce new amino acids or modify existing ones to alter the protein's conformation, potentially improving its function or stability.
3. Optimize the sequence to enhance the folding process, ensuring that the protein adopts a more stable and functional conformation.

(*@\bf\hlred{User}@*): Be creative and think about using the proteins to generate a new electronic device. How would you go about it?

(*@\bf \hl{Cephalo}@*): To create an electronic device using these proteins, the focus would be on leveraging their unique molecular properties to generate specific functions such as conductivity, sensitivity, or responsiveness. This could be achieved by:

1. Functionalizing the proteins with conductive or semiconducting materials to enhance their electrical properties.
2. Integrating the proteins into a composite material, potentially with other biopolymers or nanomaterials, to form a new type of sensor or actuator.
3. Modifying the protein structure to optimize its interaction with other components or materials, allowing for better performance in electronic devices.

By using the proteins' inherent properties, the design would aim to create a new class of materials with unique functionalities that could be applied in various electronic devices, such as flexible sensors, bioelectronics, or biomedical devices.
\end{lstlisting}
\end{LLMbox}
\textboxcaption{Chat between a User and  \texttt{Cephalo-Idefics-2-vision-12b-alpha}, focused on assessing two proteins. The model is tasked to figure out structural features and to assess likely mechanical properties based on the images. The two proteins considered are PDB IDs 5B2A (crystal structure of a mutant of OspA, a beta-sheet rich structure with around 250 residues) and 1WVK (NMR solution structure of the partially disordered protein At2g23090 from Arabidopsis thaliana). }
\label{TexBox:TB62}
\end{figure}

\begin{figure}[htbp]
\begin{LLMbox}{Chat between a User and \texttt{Cephalo-Idefics-2-vision-12b-alpha}}
\linespread{0.4}\selectfont
\begin{lstlisting} 
(*@\bf\hlred{User}@*): You are a materials scientist and you carefully analyze images, step by step.

Answer concisely, and accurately.
(*@\includegraphics[width=.5\linewidth]{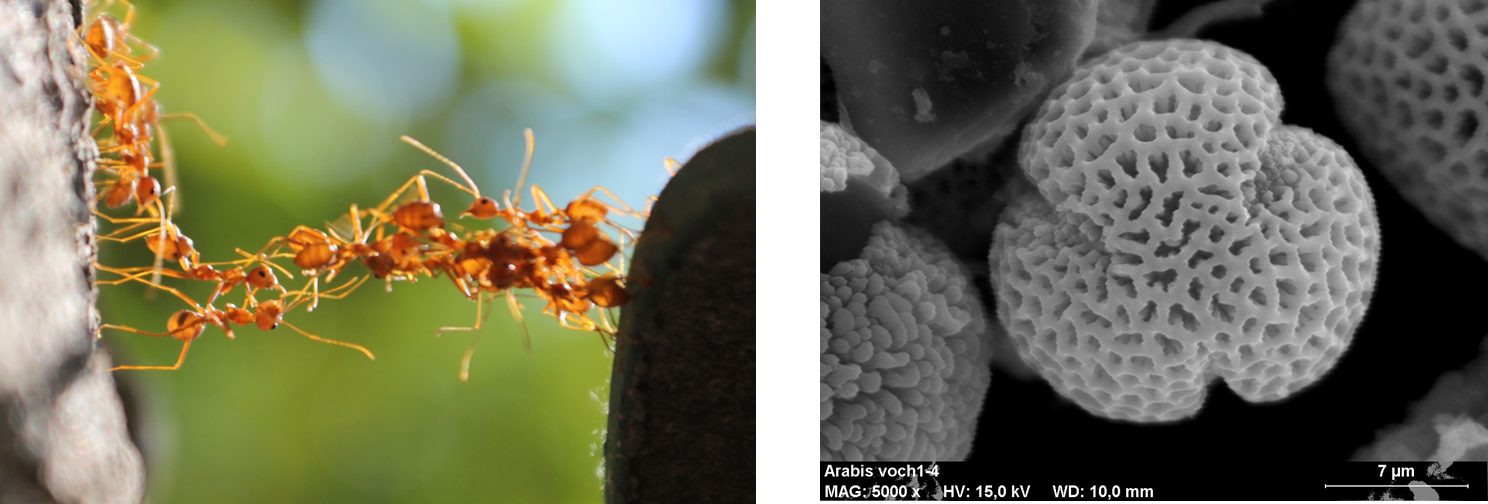}@*)
(*@\bf\hlred{User}@*): You are shown two photos of two very different objects. Carefully study the two photos.

What do you see?

(*@\bf \hl{Cephalo}@*):  The first image shows a group of ants climbing up a tree. The ants are moving in a line, using their legs to grip onto the surface of the tree. The second image shows a close-up view of a porous material, likely a biological or synthetic material, with a network of interconnected holes. The material appears to be made of a light-colored, porous material, possibly a type of foam or a natural substance. The scale bar in the second image indicates that the structure is on the microscopic scale, with a size of 7 micrometers.

(*@\bf\hlred{User}@*): Which of the two images would be better suited to create fracture-resistant materials?

(*@\bf \hl{Cephalo}@*): The second image, showing the porous structure of the biological or synthetic material, would be better suited for creating fracture-resistant materials. The interconnected holes in the material likely contribute to its strength and ability to withstand fractures, as seen in the ants' ability to climb up the tree without breaking the surface. This natural design can inspire the development of synthetic materials with similar properties.

(*@\bf\hlred{User}@*): What about the other one, what may it be useful for?

(*@\bf \hl{Cephalo}@*): The first image of ants climbing a tree can inspire the design of adhesive materials, as the ants' ability to climb with minimal slippage suggests a strong adhesive force. This natural example can guide the development of synthetic adhesives that mimic the ants' grip on surfaces.

(*@\bf\hlred{User}@*): How can these concepts be used to inform new materials?

(*@\bf \hl{Cephalo}@*): The concepts shown in these images can inspire the design of new materials by drawing from natural examples. The ants' ability to climb a tree without breaking the surface can inspire the development of adhesive materials, while the porous structure of the biological or synthetic material can inspire the creation of fracture-resistant materials. These natural designs can guide the development of synthetic materials with enhanced properties, such as strong adhesion and resistance to fractures.

(*@\bf\hlred{User}@*): Next, specifically develop a research idea that integrates the concepts seen in the two images.

(*@\bf \hl{Cephalo}@*): A research idea could involve the study of the mechanical properties of ants' claws and the porous structure of their climbing surfaces, which might provide insights into the design of new materials with enhanced adhesive and fracture resistance. By understanding the natural mechanisms that allow ants to climb with minimal slippage, we can develop synthetic materials with similar properties. Similarly, by examining the porous structure of natural materials, we can gain insights into the design of new materials that are both strong and lightweight.


(*@\bf\hlred{User}@*): Provide a step-by-step outlook to future research.

(*@\bf \hl{Cephalo}@*):  Future research could involve the detailed analysis of the ants' claws and the porous structure of their climbing surfaces, using advanced imaging techniques such as scanning electron microscopy (SEM) and atomic force microscopy (AFM) to understand the mechanical properties at the microscopic level. This could lead to the development of new materials with enhanced adhesive and fracture resistance by mimicking the natural designs observed in the ants and the porous material.


\end{lstlisting}
\end{LLMbox}
\textboxcaption{Chat between a User and \texttt{Cephalo-Idefics-2-vision-12b-alpha}, analyzing an image of ants forming a bridge and a microscopy image of a pollen particle. The model is tasked to reason over both and develop bio-inspired design concepts. Left image, reproduced from \url{https://en.m.wikipedia.org/wiki/File:AntBridge_Crossing_10.jpg}, based on a CC BY-SA 4.0, Igor Chuxlancev. Right image, reproduced from \url{https://en.wikipedia.org/wiki/Pollen\#/media/File:Arabis_voch1-4.jpg}, CC BY-SA 3.0, Marie Majaura.}
\label{TexBox:TB64}
\end{figure}

\begin{figure}[htbp]
\begin{LLMbox}{Chat between a User and \texttt{Cephalo-Idefics-2-vision-12b-alpha}}
\linespread{0.4}\selectfont
\begin{lstlisting} 
(*@\bf\hlred{User}@*): You are a materials scientist and you carefully analyze images, step by step.

Answer concisely, and accurately.
(*@\includegraphics[width=.333\linewidth]{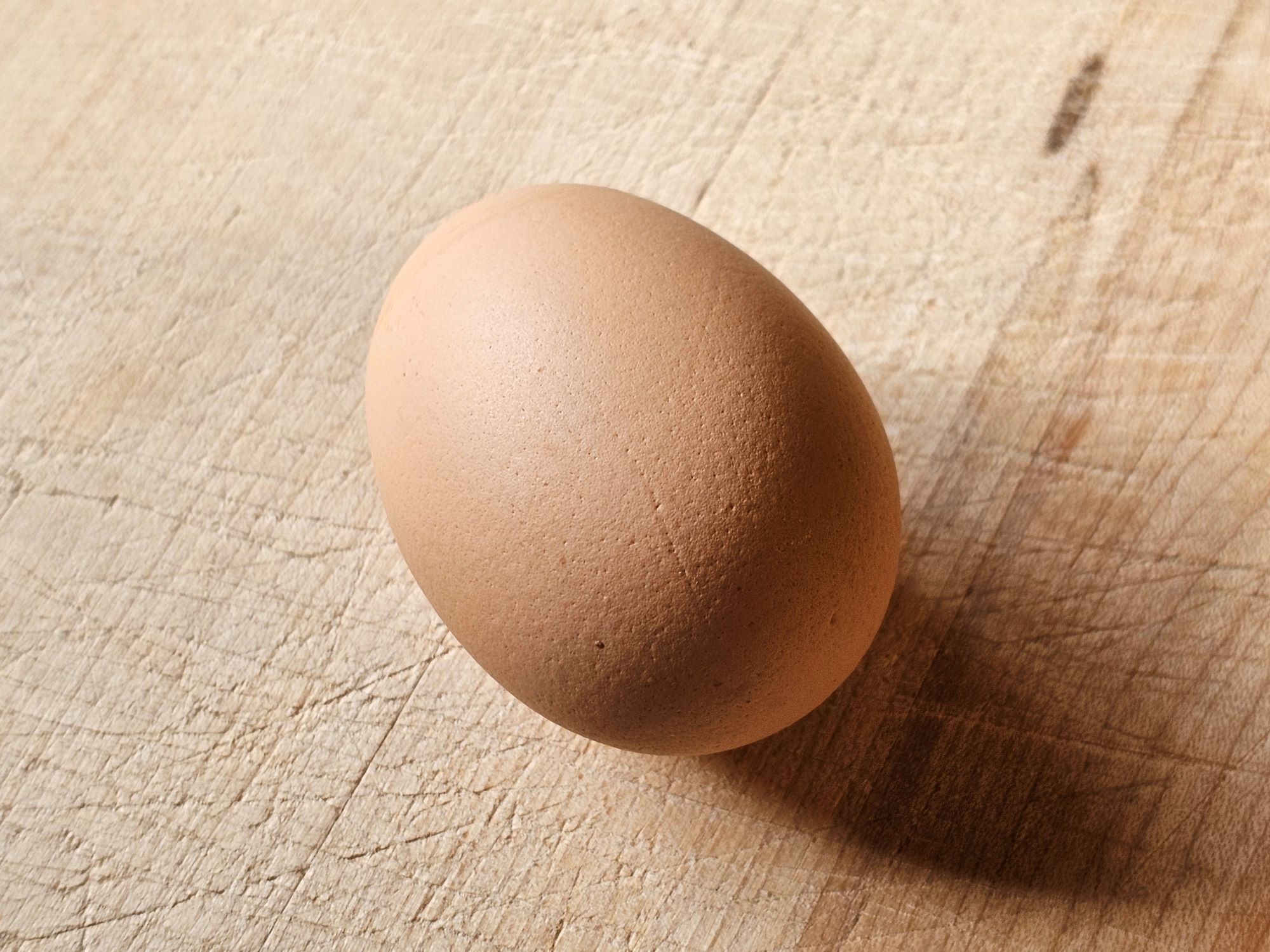}@*)
(*@\bf\hlred{User}@*): Describe what you see. Do you detect a crack?

(*@\bf \hl{Cephalo}@*): The image shows a single, uncracked, unbroken, and unadulterated chicken egg placed on a wooden surface. The egg is a natural product, and its surface appears smooth and unblemished. The wooden surface provides a contrasting texture to the egg, highlighting its round shape and smooth shell.

(*@\bf\hlred{User}@*): Use the image and propose what would happen if I were to drop the object.

(*@\bf \hl{Cephalo}@*): If I were to drop the egg, it would likely crack due to the impact. The egg's shell is not designed to withstand significant force, and the impact energy would cause the shell to deform and potentially break. This is because the eggshell is a biomaterial composed of calcium carbonate and protein, which provides some strength and rigidity but is not as resistant to impact as synthetic materials.
\end{lstlisting}
\end{LLMbox}
\textboxcaption{Chat between a User and \texttt{Cephalo-Idefics-2-vision-12b-alpha}, analyzing an image of an egg on a wooden surface. The model correctly states that there is no crack in the egg, and further explains what would happen if the egg is dropped.}
\label{TexBox:TB68}
\end{figure}

\begin{figure}[htbp]
\begin{LLMbox}{Chat between a User and \texttt{Cephalo-Idefics-2-vision-12b-alpha}}
\linespread{0.4}\selectfont
\begin{lstlisting} 
(*@\bf\hlred{User}@*): You are a materials scientist and you carefully analyze images, step by step.

Answer concisely, and accurately.
(*@\includegraphics[width=.333\linewidth]{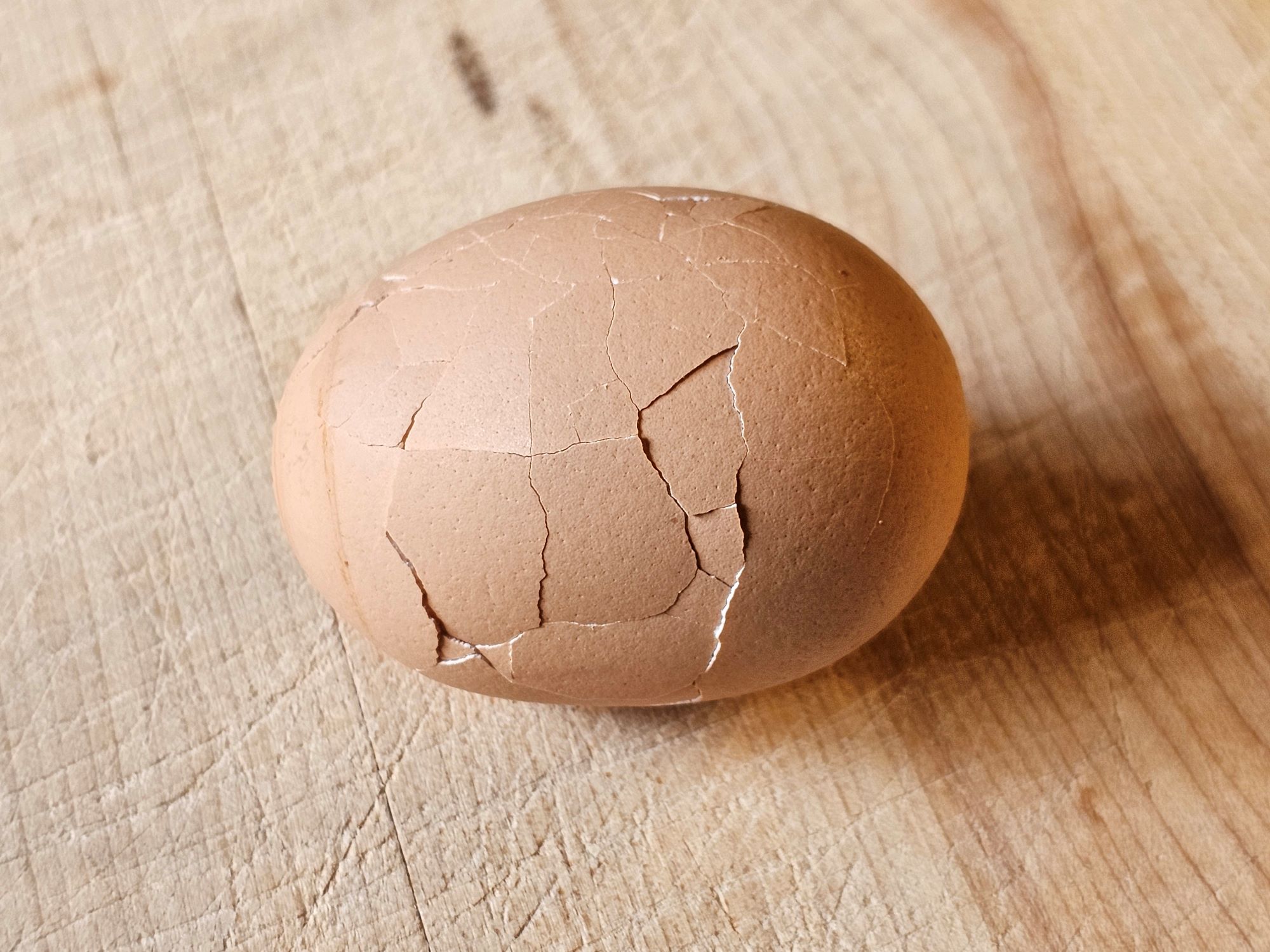}@*)
(*@\bf\hlred{User}@*): Describe what you see. Do you detect a crack?

(*@\bf \hl{Cephalo}@*): The image shows a cracked egg on a wooden surface. The egg is brown and appears to be in the process of breaking, with visible cracks and fissures on its surface. The wooden surface provides a contrasting texture to the smooth, porous surface of the egg. The cracks suggest that the egg has been subjected to some form of stress, likely leading to its breakage.

(*@\bf\hlred{User}@*): Use the image and propose what would happen if I were to drop the object.

(*@\bf \hl{Cephalo}@*): If I were to drop the egg, it would likely shatter into several pieces due to the cracks visible in the image. The egg's brittle nature and the existing damage would cause it to break apart, releasing the contents inside.
\end{lstlisting}
\end{LLMbox}
\textboxcaption{Chat between a User and \texttt{Cephalo-Idefics-2-vision-12b-alpha}, analyzing an image of a pre-cracked egg on a wooden surface. The model correctly identifies the crack in the material and also explains what would happen if the egg is dropped, as requested by the human. }
\label{TexBox:TB69}
\end{figure}

\newpage

\begin{figure}
    \centering
    \includegraphics[width=1\linewidth]{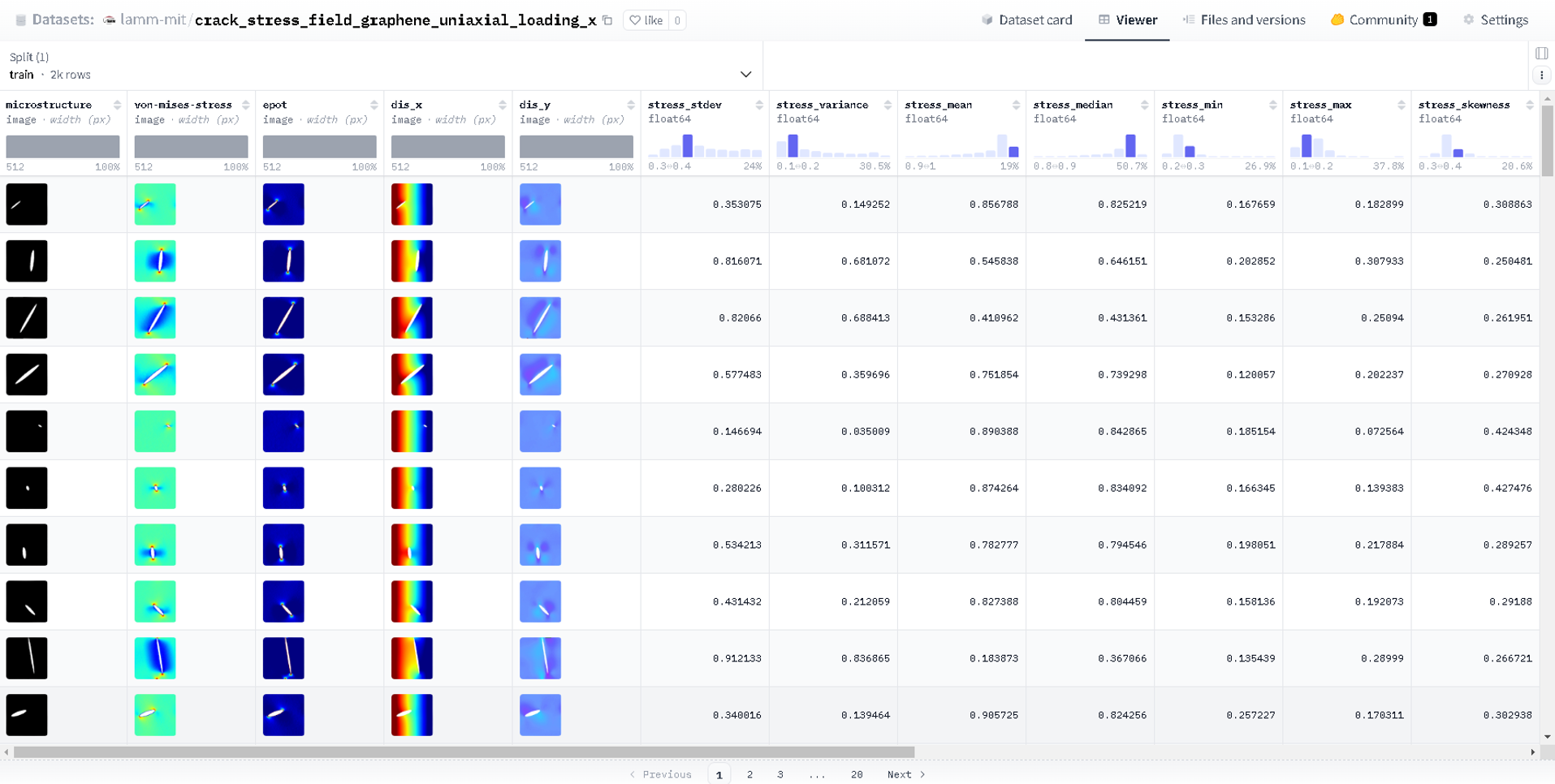}
    \caption{Snapshot of the graphene flake dataset (\url{https://huggingface.co/datasets/lamm-mit/crack_stress_field_graphene_uniaxial_loading_x} at Hugging Face), revealing the organization of the data. Each microstructure (left) is associated with Von Mises stress fields, atomic potential energy field, displacement field, and associated statistical properties of the various fields. }
    \label{fig:Fig_2089}
\end{figure}

\newpage

\begin{figure}
    \centering
    \includegraphics[width=.8\linewidth]{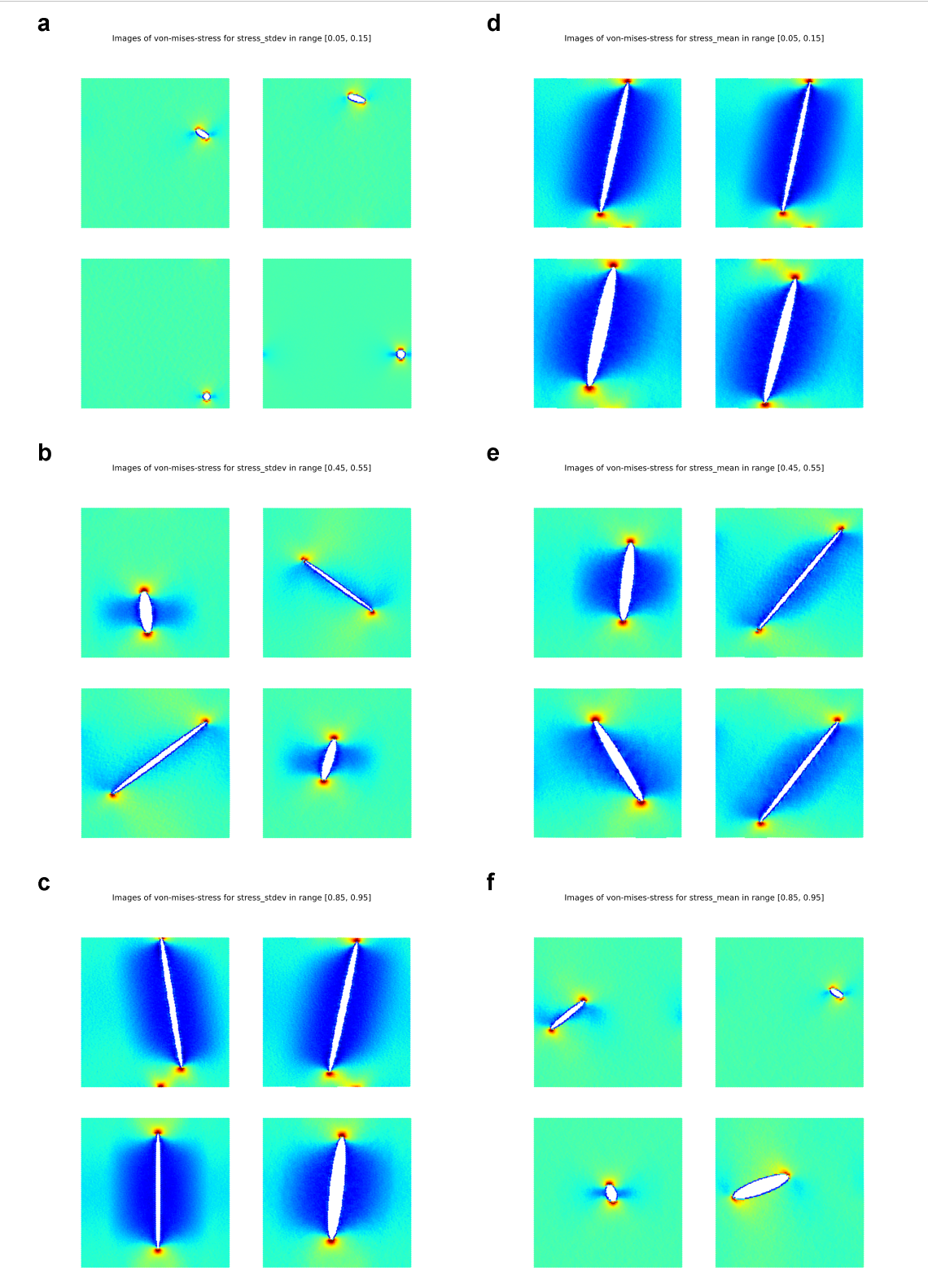}
    \caption{Samples of Von Mises stress distributions  as obtained from MD simulations in the training dataset, in different intervals of standard deviation and mean. Panel a: Low standard deviation. Panel b: Medium range standard deviation. Panel c: High standard deviation. Panel d: Low mean. Panel e: Medium range mean. Panel f: High mean.  The differences in stress field, and the nature of crack orientations, sizes and shapes that cause these, are visible. } 
    \label{fig:Fig_2002}
\end{figure}

\newpage
\begin{figure}
    \centering
    \includegraphics[width=.8\linewidth]{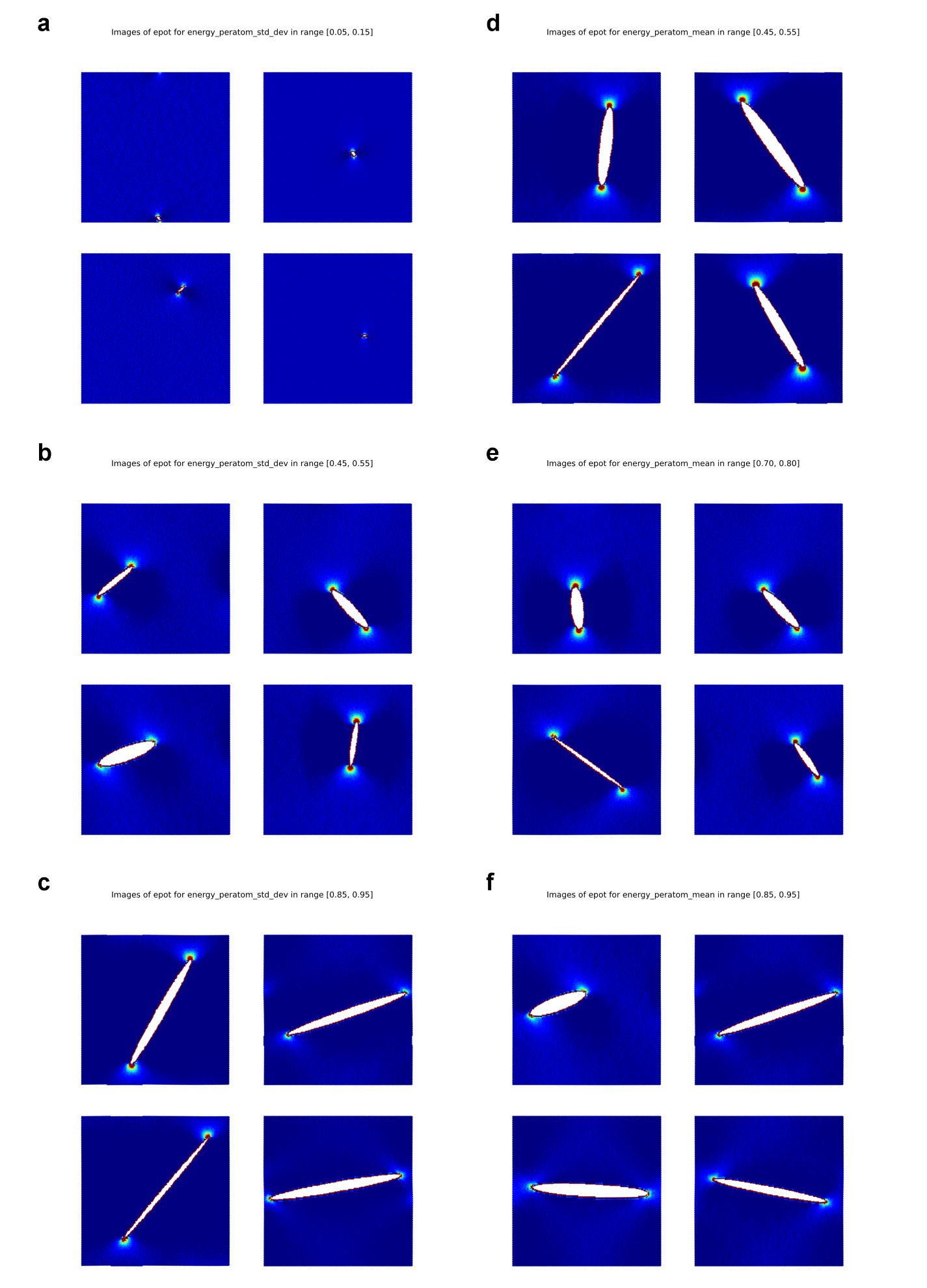}
    \caption{Samples of atomic potential energy distributions as obtained from MD simulations in the training dataset, in different intervals of standard deviation  and mean. Panel a: Low standard deviation. Panel b: Medium range standard deviation. Panel c: High standard deviation. Panel d: Low mean. Panel e: Medium range mean. Panel f: High mean.  The differences in potential energy field, and the nature of crack orientations, sizes and shapes that cause these, are visible.  } 
    \label{fig:Fig_2002_epot}
\end{figure}

\newpage

\begin{figure}
    \centering
    \includegraphics[width=1\linewidth]{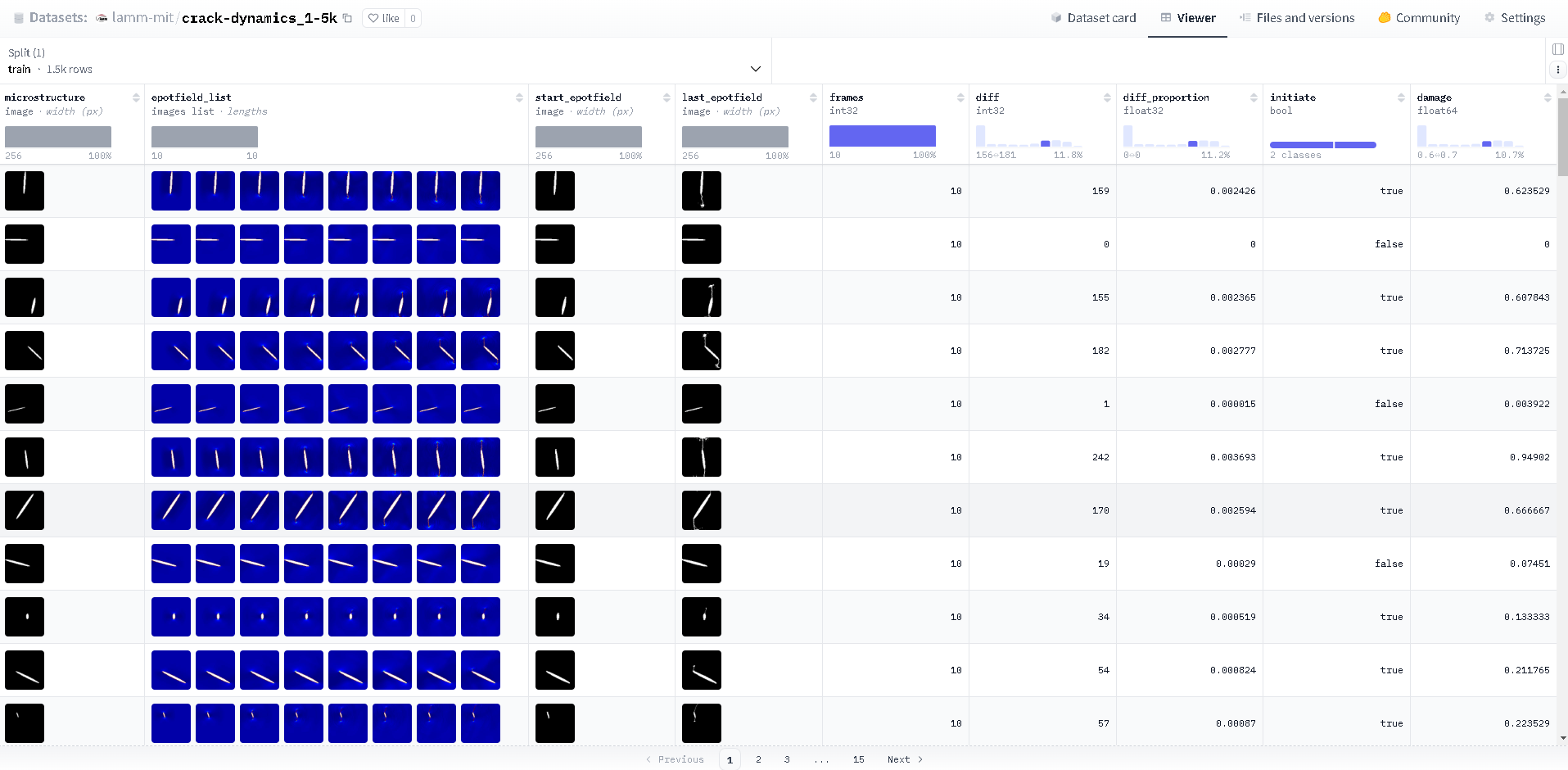}
    \caption{Snapshot of the crack dynamics dataset (\url{https://huggingface.co/datasets/lamm-mit/crack-dynamics_1-5k} at Hugging Face), revealing the organization of the data. Each microstructure (left) is associated with a set of frames that reflect the dynamical evolution of the material's response to loading, as well as the initial and final defect field, the damage at the end of the simulation, whether or not the crack initiates, and other properties. }
    \label{fig:Fig_2090}
\end{figure}

\newpage

\begin{figure}
    \centering
    \includegraphics[width=.4\linewidth]{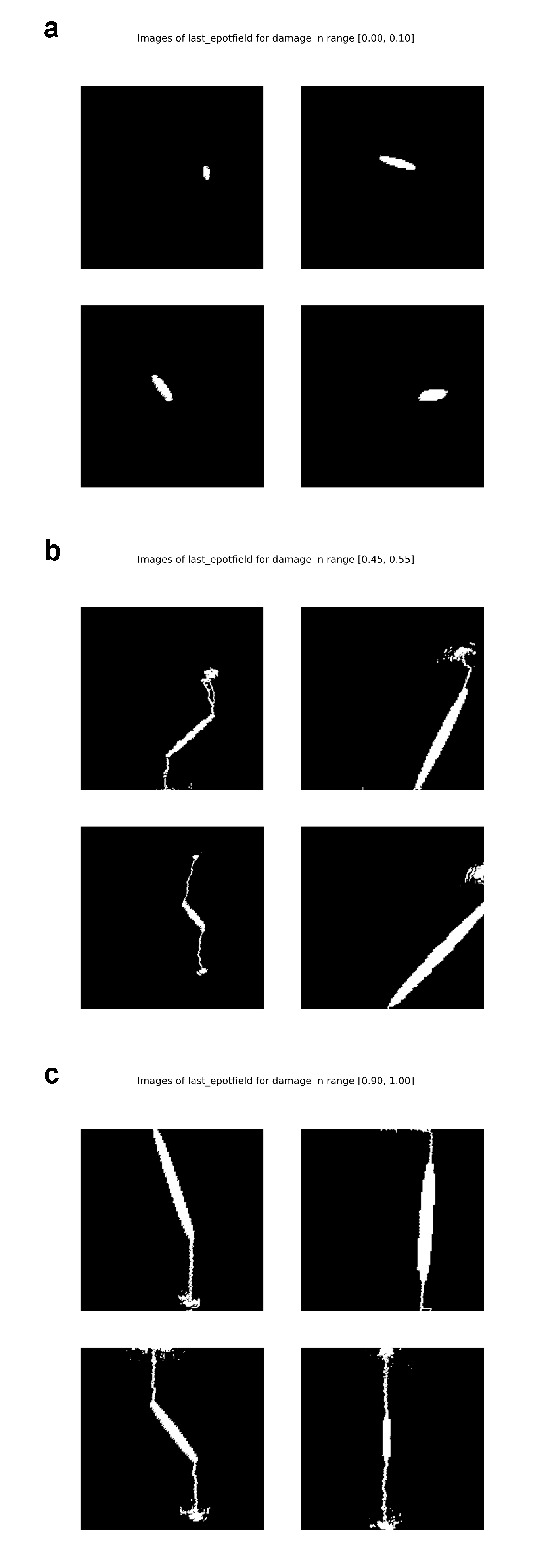}
    \caption{Samples of damage distributions as obtained from MD simulations in the training dataset. Panel a: Low damage, showing no crack growth. Panel b: Medium damage, indicating some crack growth. Panel c: High damage, revealing significant defect size and growth.} 
    \label{fig:Fig_2063}
\end{figure}

\end{document}